\documentclass[lettersize,journal]{IEEEtran}
\usepackage{amssymb}
\usepackage{bm}
\usepackage{amsmath,amsfonts}
\usepackage{algorithmic}
\usepackage{nicematrix,tikz}
\usepackage[most]{tcolorbox}
\usepackage{hf-tikz}

\usetikzlibrary{decorations.pathreplacing,calc}

\usepackage[linesnumbered,ruled]{algorithm2e}

\usepackage{array}
\usepackage{float}
\usepackage[caption=false,font=normalsize,labelfont=sf,textfont=sf]{subfig}
\usepackage{textcomp}
\usepackage{stfloats}
\usepackage{url}
\usepackage{verbatim}
\usepackage{nicefrac} 
\usepackage{microtype} 
\usepackage{booktabs}
\usepackage{colortbl}
\usepackage{hhline}
\usepackage{multicol}
\usepackage{multirow}
\usepackage[colorlinks=true]{hyperref}
\usepackage{pifont}

\newsavebox{\measurebox}
\usepackage{graphicx}
\usepackage{cite}
\usepackage{enumitem}
\setlist{leftmargin=5.5mm}
\usepackage{xcolor} 
\DeclareMathAlphabet\mathbfcal{OMS}{cmsy}{b}{n}
\newcommand{\x}{\mathbf{x}}
\newcommand{\y}{\mathbf{y}}
\newcommand{\bPhi}{\mathbf{\Phi}}
\newcommand{\n}{\mathbf{n}}
\newcommand{\R}{\mathbb{R}}
\newcommand{\D}{\mathbf{D}}
\newcommand{\bd}{\pmb{d}}

\newcommand{\z}{\mathbf{z}}
\newcommand{\balpha}{\pmb{\alpha}}
\newcommand{\DC}{\text{D}^{\text{3}} \text{C}^{\text{2}}}
\newcommand\shline{\specialrule{0.8pt}{0pt}{0pt}}
\newcommand{\mrev}[1]{{#1}}
\newcommand{\rev}[1]{{#1}}

\newcommand{\msecondbest}[1]{\underline{\textcolor{blue}{#1}}}

\begin{document}

\title{$\DC$-Net: Dual-Domain Deep Convolutional Coding Network for Compressive Sensing}
\author{Weiqi~Li, Bin~Chen, Shuai~Liu, Shijie~Zhao, Bowen~Du,~\IEEEmembership{Member,~IEEE}, \\~~~~~~~Yongbing~Zhang,~\IEEEmembership{Member,~IEEE}, and Jian~Zhang,~\IEEEmembership{Member,~IEEE}
\thanks{Manuscript received February 20, 2024; revised April 13, 2024; accepted April 30, 2024. This work was supported in part by the National Natural Science Foundation of China (No. 62331011) and the Shenzhen Research Project (No. JCYJ20220531093215035).}
\thanks{Weiqi Li, Bin Chen, and Jian Zhang are with the School of Electronic and Computer Engineering, Peking University, Shenzhen 518055, China. (e-mail: liweiqi@stu.pku.edu.cn; chenbin@stu.pku.edu.cn; zhangjian.sz@pku.edu.cn). (\textit{Corresponding author: Jian~Zhang.})}
\thanks{Shuai Liu is with the Graduate School at Shenzhen, Tsinghua University, Shenzhen 518055, China. (e-mail: liu-s20@mails.tsinghua.edu.cn). }
\thanks{Shijie Zhao is with ByteDance Inc, Shenzhen 518055, China. (e-mail: zhaoshijie.0526@bytedance.com).}
\thanks{Bowen Du is with the School of Software Engineering, Tongji University, Shanghai 200092, China. (e-mail: bowendu@tongji.edu.cn).}
\thanks{Yongbing Zhang is with the School of Computer Science and Technology, Harbin Institute of Technology (Shenzhen), Shenzhen 518055, China. (e-mail: ybzhang08@hit.edu.cn).}
}

\markboth{Journal of \LaTeX\ Class Files, 2023}%
{Shell \MakeLowercase{\textit{et al.}}: A Sample Article Using IEEEtran.cls for IEEE Journals}


\maketitle

\IEEEpubid{\begin{minipage}{\textwidth}\ \centering
		Copyright \copyright 2024 IEEE. Personal use of this material is permitted. \\
		However, permission to use this material for any other purposes must be obtained 
		from the IEEE by sending an email to pubs-permissions@ieee.org.
\end{minipage}}
\IEEEpubidadjcol

\begin{abstract}
By mapping iterative optimization algorithms into neural networks (NNs), deep unfolding networks (DUNs) exhibit well-defined and interpretable structures and achieve remarkable success in the field of compressive sensing (CS). However, most existing DUNs solely rely on the image-domain unfolding, which restricts the information transmission capacity and reconstruction flexibility, leading to their loss of image details and unsatisfactory performance. To overcome these limitations, this paper develops a dual-domain optimization framework that combines the priors of (1) image- and (2) convolutional-coding-domains and offers generality to CS and other inverse imaging tasks. By converting this optimization framework into deep NN structures, we present a Dual-Domain Deep Convolutional Coding Network ($\DC$-Net), which enjoys the ability to efficiently transmit high-capacity self-adaptive convolutional features across all its unfolded stages. Our theoretical analyses and experiments on simulated and real captured data, covering 2D and 3D natural, medical, and scientific signals, demonstrate the effectiveness, practicality, superior performance, and generalization ability of our method over other competing approaches and its significant potential in achieving a balance among accuracy, complexity, and interpretability. \mrev{Code is available at \url{https://github.com/lwq20020127/D3C2-Net}}.
\end{abstract}

\begin{IEEEkeywords}
Compressive sensing, inverse imaging problem, image restoration, convolutional coding, deep unfolding network.
\end{IEEEkeywords}

\section{Introduction}
\IEEEPARstart{A}{s} a novel paradigm in signal acquisition, compressive sensing (CS) aims to reconstruct the original signal from a few of linear measurements \cite{candes2006robust, baraniuk2007compressive, zhang2023physics, gan2023learned,zhuang2023ucsnet,zhou2023iterative,shi2020video,zhao2016video,yu2022perceptual}. It has found successful applications in various areas, including but not limited to single-pixel imaging (SPI) \cite{duarte2008single, rousset2016adaptive}, magnetic resonance imaging (MRI) \cite{lustig2007sparse}, and snapshot compressive imaging (SCI) \cite{wu2021dense}.

Mathematically, given the original vectorized image $\x\in\R^{N}$ and a sampling matrix $\bPhi\in\R^{M\times N}$, the observation process is formulated as $\y=\bPhi\x+\n \in\R^{M}$, where $\n$ is the additive white Gaussian noise (AWGN) with standard deviation (or noise level) $\sigma$. Recovering $\x$ from the obtained $\y$ is a typical ill-posed inverse problem due to the common setup of $M\ll N$, and the CS ratio (or sampling rate) is defined as $\gamma = {M}/{N}$.
Generally, traditional model-based approaches reconstruct the latent clean $\x$ by solving the following optimization problem:
\begin{equation} \label{eq:1}
    \hat{\x}=\underset{\x}{\arg\min}~\frac{1}{2}\|\bPhi\x-\y\|_2^2+\lambda\phi(\x).
\end{equation}
Here, \mrev{$\hat{\x} \in \R^{N}$ denotes the reconstructed result}, while $\lambda\phi(\x)$ represents an image prior term with a regularization parameter $\lambda \in \R^{+}$. In the case of sparsity-based traditional methods \cite{zhang2014group, zhang2014image, chen2022image}, $\lambda\phi(\mathbf{x})$ is often designed manually using some pre-defined sparsifying bases, such as the classic wavelet basis and discrete cosine transform (DCT) \cite{zhao2014image} basis. While these model-based techniques offer interpretability and strong convergence guarantees, they suffer from high computational complexity and the challenge of selecting optimal hyper-parameters \cite{zhao2016nonconvex}.

\begin{figure}
  \centering
  \includegraphics[width=\linewidth]{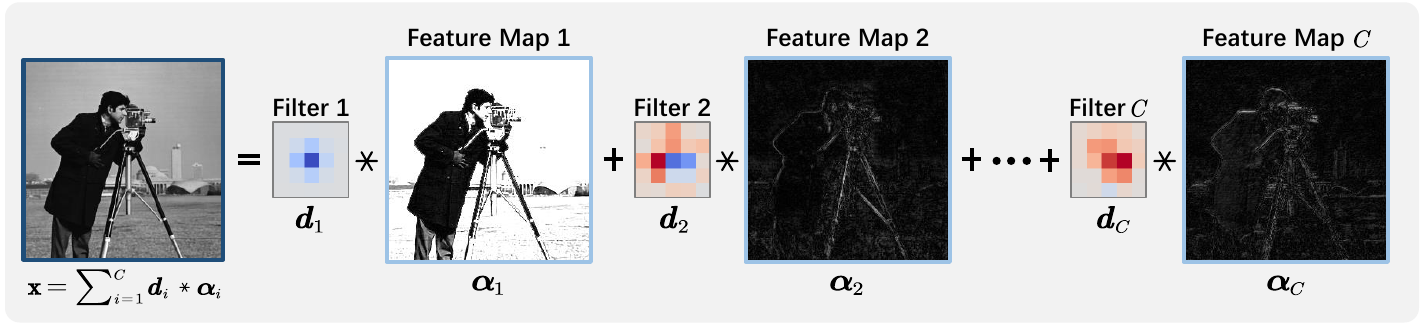}
  \vspace{-20pt}
  \caption{\mrev{Illustration of the classic concept of convolutional coding.} An image $\x$ can be expressed by the summation of multiple image-level convolution results, \textit{i.e.}, there is $\x=\sum_{i=1}^{C}\left(\bd_i \ast \balpha_i\right) \in \R^{H\times W}$, where $\bd_i\in\R^{k\times k}$ represents the $i$-th dictionary filter, $\balpha_i\in\R^{H\times W}$ is the $i$-th coefficient map, $\ast$ denotes the convolution operator, and $C$ is the number of feature channels. The darker \textcolor{red}{red} (or \textcolor{blue}{blue}) colors in visualized filter $\bd_i$ correspond to the \textcolor{red}{positive} (or \textcolor{blue}{negative}) filter elements with larger absolute values. Compared to single-channel images, this feature-level representation naturally enjoys higher capacity.}
  \vspace{-10pt}
  \label{fig1:ccmodel}
\end{figure}

\begin{figure*}[t]
  \centering
  \includegraphics[width=\textwidth]{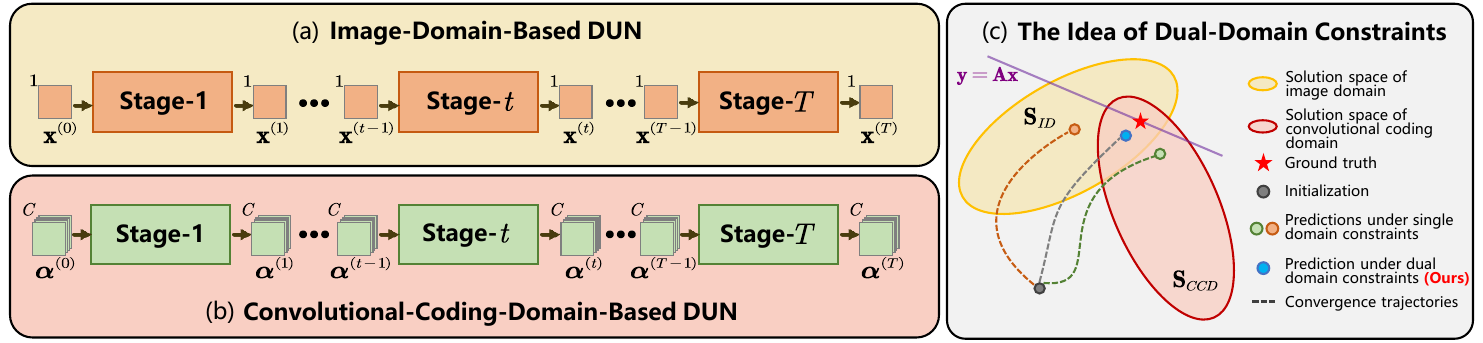}
  \vspace{-23pt}
  \caption{Illustration of the core concepts underlying the design of our introduced dual-domain deep unfolding networks (DUNs). \textcolor{blue}{(a)} showcases the architecture of image-domain (ID)-based DUN, while \textcolor{blue}{(b)} depicts the architecture of convolutional-coding-domain (CCD)-based DUN. In comparison to (a), which sequentially reconstructs the target image ($\{\x^{(t)}\}$), CCD-based DUN in (b) transmits high-dimensional features ($\{\balpha^{(t)}\}$) across all stages. Furthermore, \textcolor{blue}{(c)} provides a conceptual visualization of the assumed convergence trajectories under single-domain and dual-domain constraints, demonstrating that the latter has the potential of achieving more accurate recoveries by leveraging the combined knowledge acquired in both of the two domains. 
  }
  \vspace{-13pt}
  \label{fig:dualdomaincomparison}
\end{figure*}

\rev{With the rapid advancements in deep learning~\cite{yan2022video, wang2023visual, yan2022solve, liu2023tripartite, qin2024unified}}, a multitude of NN-based approaches have emerged, generally categorized as deep non-unfolding networks (DNUNs) and deep unfolding networks (DUNs). DNUNs treat the CS reconstruction as a de-aliasing problem, aiming to learn the inverse mapping $\y \mapsto \x$ through end-to-end ``\textit{black-box}'' NNs \cite{hyun2018deep, kulkarni2016reconnet, sun2020dual, shi2019image, shi2019scalable}. However, these methods heavily rely on careful tuning, introducing challenges to analysis and real deployments. On the other hand, DUNs leverage the benefits of networks and optimization frameworks by training truncated unfolding inference models \cite{zhang2018ista, zhang2020optimization, you2021coast, you2021ista, song2021memory, chen2022cas}. Typically composed of a fixed number of iterative unfolded NN stages, DUNs have gained prominence in the CS field due to their well-defined structure and superior performance.
\IEEEpubidadjcol

However, the majority of existing DUNs are built based on straightforward image-domain unfoldings, where the input and output of each stage are single-channel images with limited information capacity. There are often some NN operations (\textit{e.g.} convolutions) of channel number reduction from many to one at the end of each deep unfolded stage, leading to constrained transmission and the loss of image details \cite{zhang2018ista, zhang2020optimization, you2021ista, you2021coast, chen2022cas}.

Recently, convolutional coding techniques have been successfully integrated into DUNs, as demonstrated in Fig.~\ref{fig1:ccmodel} \cite{fu2019jpeg, wang2020model, zheng2021deep}. The fundamental concept behind the classic convolutional coding model is to represent target image $\x\in\R^{H\times W}$ using $\x = \D \circledast \balpha \mrev{:=}\sum_{i=1}^{C}\left(\bd_i \ast \balpha_i\right)$. Here, $\ast$ is the 2D convolution operator, while $C$ denotes the number of feature channels. $\D \in\R^{C \times k\times k}$ is the convolutional dictionary of size $k$, and $\bd_i$ is the $i$-th dictionary filter. Moreover, $\balpha\in\R^{C \times H \times W}$ represents the convolutional coefficients of $\x$, where $\balpha_i$ signifies the $i$-th channel of $\balpha$. Leveraging the inherent advantage of $\balpha$ being a multi-channel feature, these DUNs enable the high-capacity feature transmission across stages. However, their focus remains primarily on specific tasks such as rain removal \cite{wang2020model} and image denoising \cite{zheng2021deep}, thereby overlooking more general scenarios.

This paper proposes a \textbf{D}ual-\textbf{D}omain \textbf{D}eep \textbf{C}onvolutional \textbf{C}oding \textbf{Net}work, referred to as the \textbf{$\DC$-Net}. \mrev{Our focus lies in tackling general CS and other inverse imaging problems based on a dual-domain unconstrained Lagrangian formulation of both the image-domain (ID) prior and the convolutional-coding-domain (CCD) prior. Mathematically, it overrides Eq.~(\ref{eq:1}) as $\{\z, \balpha\}= {\arg\min}_{\{\z, \balpha\}}~(1/2)\|\bPhi\z-\y\|_2^2 + (\mu_\z/2)\|\z-\D\circledast\balpha\|_2^2 + \lambda\phi(\z) + \tau\psi(\balpha)$, which closely relates to the constrained optimization problem \cite{boyd2004convex,fowler2012block} of minimizing $\lVert \bPhi \z - \y \rVert_2^2$ with respect to $\z$ and $\balpha$, subject to $\phi(\z)\le\epsilon_1$, $\psi(\balpha)\le\epsilon_2$, and $\z=\D \circledast \balpha$. Here, $\D$ represents a designed or learned convolutional dictionary, $\phi$ and $\psi$ are generally assumed to be non-negative and convex mappings, and $\epsilon_1,\epsilon_2\in[0,+\infty)$ denote the tolerance. As illustrated in Fig.~\ref{fig:dualdomaincomparison} (c), our method aims to guide the DUN-empowered recovery process towards the intersection of the image-domain feasible set $\mathbf{S}_{ID}=\{\z|\phi(\z)\le\epsilon_1\}$ and the CCD feasible set $\mathbf{S}_{CCD}=\{\z|\psi(\balpha)\le\epsilon_2, \z=\D \circledast \balpha\}$, \textit{i.e.}, $\mathbf{S}=\mathbf{S}_{ID}\cap \mathbf{S}_{CCD}$.} By introducing a new dual-domain deep unfolding framework, we overcome the limitations of existing approaches, enabling $\DC$-Net to efficiently transmit high-capacity information while benefiting from the explicit constraints imposed by both ID and CCD. \mrev{Our proposed $\DC$-Net can be regarded as a significant generalization of DCDicL~\cite{zheng2021deep} to the CS tasks by considering non-identity $\bPhi$s, incorporating ID prior, and employing a universal dictionary.}

The contributions of this work are summarized as follows:

\vspace{3pt}
\noindent \ding{113} (1) We introduce a new dual-domain optimization framework, which leverages the strengths of both image-domain (ID) and convolutional-coding-domain (CCD) priors. This framework tries to constrain the feasible solution space and can be effective for CS and readily extended to other inverse imaging problems.

\vspace{3pt}
\noindent \ding{113} (2) Building upon the framework, we design $\DC$-Net, a dual-domain deep convolutional coding network tailored for CS. It propagates high-capacity feature-level image representations across all unfolded stages, adaptively capturing crucial features and enhancing the recovery of intricate details and textures.

\vspace{3pt}
\noindent \ding{113} (3) Through theoretical analyses and extensive experiments conducted on simulated and real data of 2D and 3D natural, medical, and scientific signals, we demonstrate the significant superiority and effectiveness of $\DC$-Net in improving both the reconstruction quality and generalization ability across four representative tasks: image CS, accelerated MRI, inpainting, and sparse-view computed tomography (CT). A real SPI optics system is established for performance validation on the CS-based fluorescence microscopy \cite{lichtman2005fluorescence} task. Corresponding analyses and discussions are also provided to try to offer an understanding and insights into our method's working principle.

\textbf{Organization of This Paper:} The remainder of this paper is organized as follows. In Sec.~\ref{sec:related_work}, we present a comprehensive review of deep unfolding networks (DUNs) and deep convolutional coding models that are most relevant to our work in the context of inverse imaging problems. Sec.~\ref{sec:proposed_method} introduces our proposed approach, the $\DC$-Net, which draws inspiration from our developed dual-domain (ID-/CCD-based) optimization algorithm framework. This section also includes theoretical analyses on the basic convergence of the \mrev{objective} function value. In Sec.~\ref{sec:experiments}, we validate the effectiveness of high-throughput transmission and our dual-domain unfolding design. We report our experimental results on five representative CS-related tasks, including natural image CS, compressive sensing MRI, inpainting, sparse-view CT, and single-pixel imaging for fluorescence microscopy. These tasks encompass both simulated and real-world data, covering a range of natural, medical, and scientific signals. Finally, Sec.~\ref{sec:conclusion} summarizes the conclusions of this work. Further details regarding our real CS-based SPI experiments and theoretical analyses are provided in the \textbf{\textit{\textcolor{blue}{supplemental material}}}.

\section{Related Work}
\label{sec:related_work}
\textbf{Deep Unfolding Network.} DUNs have emerged as effective solutions for a wide range of inverse imaging problems~\cite{chen2016trainable, zheng2021deep}. In the context of image CS and CS-MRI, DUNs typically incorporate some NN denoisers and iterative optimization algorithms, such as the alternating minimization (AM) \cite{schlemper2017deep}, half quadratic splitting (HQS) \cite{zhang2017learning}, iterative shrinkage-thresholding algorithm (ISTA) \cite{zhang2018ista, zhang2020optimization, song2021memory}, alternating direction method of multipliers (ADMM) \cite{yang2018admm}, and inertial proximal algorithm for nonconvex optimization (iPiano) \cite{su2020ipiano}.
Despite the well-defined frameworks of existing DUNs, their inherent reliance on ID unfolding hampers the feature transmission capabilities of NN processing. Some DUNs incorporate the intermediate features from previous deep unfolded stages as auxiliary information in the current stage, but remain rooted in the strictly inflexible deep ID unfolding, thereby limiting their further enhancement \cite{chen2020learning, song2021memory} for real CS applications.

\textbf{Deep Convolutional Coding.} Convolutional coding has been extensively investigated in the context of image restoration \cite{gu2015convolutional, deng2020deep}. In contrast to other general sparse coding techniques, convolutional coding introduces the inductive bias of locality and shift-invariance. Nevertheless, most existing convolutional coding methods rely on hand-crafted sparsity priors \cite{fu2019jpeg, xu2020limited, sreter2018learned, gao2022multi}, such as $\ell_0$- and $\ell_1$-regularization, rather than learning rich knowledge from data. Recently, there has been an exploration of integrating convolutional coding models into DUNs. Wang \textit{et al.}\cite{wang2020model} propose an interpretable deep network for rain removal, while Zheng \textit{et al.}\cite{zheng2021deep} introduce a deep convolutional dictionary learning framework known as DCDicL for denoising. However, these methods are specifically designed for scenarios where the measurement (or degradation) matrix $\bPhi$ in Eq.~(\ref{eq:1}) is constrained to be the identity, denoted as $\bPhi=\mathbf{I}_N$, with a focus on suppressing undesired rain streaks and noise. Generalizing these approaches to address the solution ambiguity inherent in ill-posed image recovery problems is non-trivial and constitutes a major concern to be tackled by our proposed method.
\vspace{-4pt}

\section{Proposed Method}
\label{sec:proposed_method}
\subsection{Dual-Domain Optimization Objective for CS Reconstruction}
In contrast to existing ID-based DUNs, we draw inspiration from convolutional coding methods to enhance NN information transmission capability. Figs.~\ref{fig:dualdomaincomparison}~(a) and (b) exhibit the architectures of ID-based and CCD-based DUNs, respectively. It is evident that the design of ID-based DUNs, where both the input and output at each stage are strictly implemented by single-channel image $\x$ in Eq.~(\ref{eq:1}), limits the data throughput of NN processing. By capitalizing on the inherent advantages of $C$-channel convolutional coefficients $\balpha$, CCD-based DUNs enable high-capacity transmission across stages. Importantly, the prior term in Eq.~(\ref{eq:1}) plays a vital role in the reconstruction process by constraining the feasible solution space. Motivated by this intuitive idea, we perform CS reconstruction based on a model-level dual-domain integrated optimization framework of both the ID and CCD priors that overrides Eq.~(\ref{eq:1}) as follows:
\begin{align} \label{eq:dual-domain-fuction}
\begin{split}
    \{\z, \balpha\}=
    \underset{\{\z, \balpha\}}{\arg\min}~&\frac{1}{2}\|\bPhi\z-\y\|_2^2 +  \frac{\mu_\z}{2}\|\z-\D\circledast\balpha\|_2^2
    \\& + \lambda\phi(\z) + \tau\psi(\balpha).
\end{split}
\end{align}

\textfloatsep = 0.7\baselineskip plus 0.2\baselineskip minus 0.4\baselineskip

\SetCommentSty{textit}
\SetKwComment{tcc}{}{} 
\SetKwInOut{Input}{Input}\SetKwInOut{Output}{Output}
\begin{algorithm}[!t]
\Input{Sampling matrix $\bPhi$, measurement $\y$, globally unified convolutional dictionary $\D$, step size $\rho$, trade-off parameters $\mu_\z$, $\lambda$, $\eta$ and $\beta$, and total number of iterations $T$.}
 \Output{Recovered image $\hat{\x}$.}
\BlankLine
Initialize $\balpha^{(0)}$ from $\y$ and $\bPhi$, and let $\z^{(0)}=\D\circledast\balpha^{(0)}$.\\
\For{$t = 1, 2, \cdots, T$}
{
\begin{minipage}[t]{\linewidth}
$\tilde{\z}^{(t)} = \z^{(t-1)}- \rho \left(\bPhi^{\top}\left(\bPhi\z^{(t-1)}-\y\right)\right.\\
\left.\ \ \ \ \ \ \ \ \ \ \ \ \ \ \ \,  +\mu_\z\left(\z^{(t-1)}-\D\circledast\balpha^{(t-1)}\right) \right)$~(\ref{eq:GDM})\;
\end{minipage}

\vspace{4pt}

\begin{minipage}[t]{\linewidth}
$\z^{(t)} = \mathrm{prox}_{\lambda\phi}^{\rho}(\tilde{\z}^{(t)})=\underset{\z}{\arg\min}~\frac{1}{2\rho}\|\z-\tilde{\z}^{(t)}\|_2^2+\lambda\phi(\z)$~(\ref{eq:PMN})\;
\end{minipage}

\begin{minipage}[t]{\linewidth}
$ \tilde{\balpha}^{(t)} = \underset{\balpha}{\arg\min}~\frac{1}{2}\|\D\circledast\balpha-\z^{(t)}\|_2^2+\frac{\eta}{2}\|\balpha-\balpha^{(t-1)}\|_2^2$~(\ref{eq:data_term})\;
\end{minipage}

$  \balpha^{(t)} = \underset{\balpha}{\arg\min}~ \frac{1}{2}\|\balpha - {\tilde{\balpha}}^{(t)}\|_2^2 +
    \beta \psi({\balpha})$~(\ref{eq:prior_term})\;
}

$\hat{\x}=\D\circledast\balpha^{(T)}$\;
\Return{$\hat{\x}$.}
\caption{Our developed dual-domain (ID \& CCD) iterative optimization framework for CS reconstruction.}\label{alg:optimization}
\end{algorithm}
\newcommand{\tikzmark}[1]{\tikz[overlay,remember picture] \node (#1) {};}

\newcommand*{\AddNote}[4]{%
    \begin{tikzpicture}[overlay, remember picture]
        \draw [decoration={brace,amplitude=0.3em, raise=0.3em},decorate,very thick,blue]
            ($(#3)!(#1.north)!($(#3)-(0,1)$)+(0,0.24)$) --  
            ($(#3)!(#2.south)!($(#3)-(0,1)$)+(0,-0.06)$)
                node [align=left, text width=1.5cm, pos=0.5, anchor=west] {#4};
    \end{tikzpicture}
}%

\definecolor{myIDBcolor}{HTML}{FFF0E6}
\definecolor{myCCDBcolor}{HTML}{FFEBE6}

\SetCommentSty{textit}
\SetKwComment{tcc}{}{} 
\SetKwInOut{Input}{Input}\SetKwInOut{Output}{Output}
\begin{algorithm}[!h]
\Input{Sampling matrix $\bPhi \in \R^{M\times N}$, measurement $\y \in \R^{M}$, globally unified convolutional dictionary $\D \in \R ^{C\times k \times k}$, CS ratio $\gamma \in (0,1]$, and total number of unfolded stages $T\in \mathbb{Z}^{+}$.}
 \Output{Recovered image $\hat{\x} \in \R^{1\times 
 H \times W}$.}
\BlankLine

$\x_{{init}}=\bPhi^{\top}\y\in\R^{1\times H \times W}$\;

$\balpha^{(0)}=\mathcal{G}_{\mathrm{InitNet}}\left(\x_{{init}},\gamma\right)\in\R^{C\times H \times W}$~(\ref{eq:InitNet_impl})\;
$\z^{(0)}=\D\circledast\balpha^{(0)}\in\R^{1\times H \times W}$~(\ref{eq:z_Init})\;

\For{$t = 1, 2, \cdots, T$}{
$\left(\rho^{(t)}, \mu_{\z}^{(t)}, \eta^{(t)}, \beta^{(t)} \right) = \mathcal{G}_{\mathrm{HPN}}^{(t)}\left(\gamma\right) \in \R^{4}$~(\ref{eq:HPN_impl})\;

\vspace{4pt}

\tikzmark{top1}\begin{minipage}[t]{\linewidth}
\tikzmarkin[fill=myIDBcolor, draw=myIDBcolor, rounded corners=3pt]{a}$\tilde{\z}^{(t)} = \mathcal{G}_{\mathrm{GDM}}\left(\balpha^{(t-1)}, \z^{(t-1)} , \right.\\ 
\left. \ \ \ \ \ \ \ \ \ \ \ \ \ \ \ \ \,  \rho^{(t)}, \mu_\z^{(t)}\right)\in\R^{1\times H \times W}$~(\ref{eq:GDM_impl})\;
\end{minipage}

\begin{minipage}[t]{\linewidth}
$\z^{(t)}= \mathcal{G}_{\mathrm{PMN}}^{(t)}\left(\tilde{\z}^{(t)}\right)\in\R^{1\times H \times W}$~(\ref{eq:PMN_impl});\tikzmark{bottom1}\qquad\hspace{14pt}\tikzmarkend{a}
\end{minipage}

\vspace{4pt}

\tikzmark{top2}\begin{minipage}[t]{\linewidth}
\tikzmarkin[fill=myCCDBcolor, draw=myCCDBcolor, rounded corners=3pt]{c}$\tilde{\balpha}^{(t)} = \mathcal{G}_{\mathrm{DTSM}}\left(\balpha^{(t-1)} , \right.\\
\left. \ \ \ \ \ \ \ \ \ \ \ \ \ \ \ \ \  \z^{(t)}, \eta^{(t)}\right)\in\R^{C\times H \times W}$~(\ref{eq:DTSM_impl});\ 
\end{minipage}

\vspace{4pt}

\begin{minipage}[t]{\linewidth}
$\balpha^{(t)}= \mathcal{G}_{\mathrm{PTSN}}^{(t)}\left(\tilde{\balpha}^{(t)}, \beta^{(t)}\right)\in\R^{C\times H \times W}$\tikzmark{bottom2}~(\ref{eq:PTSN_impl});\ \tikzmarkend{c}\hspace{2pt}\tikzmark{right2}\\
\end{minipage}
}

$\hat{\x}=\D\circledast\balpha^{(T)}\in\R^{1\times H \times W}$\;

\Return{$\hat{\x}$.}
\caption{Our implemented unfolded CS recovery of $\DC$-Net. Its NN architecture is illustrated in Fig.~\ref{fig:network}}\label{alg:network}
\end{algorithm}

\AddNote{top1}{bottom1}{right2}{\ \ IDB}
\AddNote{top2}{bottom2}{right2}{\ \ CCDB}

\vspace{-8pt}
Here, $\z\in\R^{H\times W}$ denotes estimated image, $\balpha\in\R^{C\times H \times W}$ contains the convolutional coefficients, while $\lambda \phi(\z)$ and $\tau \psi(\balpha)$ represent the ID and CCD prior terms, respectively. Trade-off parameters $\mu_\z$, $\lambda$, and $\tau$ control the balance among different terms. Fig.~\ref{fig:dualdomaincomparison} (c) illustrates the concept of advantages of the dual-domain priors. It can be seen that the introduction of dual-domain priors constrains the feasible solution space, resulting in more accurate reconstruction compared to single-domain-based models. It is worth noting that unlike the objective functions in \cite{wang2020model} and \cite{zheng2021deep}, where the measurement matrix $\bPhi$ is specifically the identity $\mathbf{I}_N$, our method is applicable to more general cases.

\subsection{Dual-Domain Optimization Algorithm Framework}
Given the sampling matrix $\bPhi$, CS measurement $\y$, globally unified convolutional dictionary $\D$, and trade-off parameters $\mu_\z$, $\lambda$, and $\tau$, along with the initialization $\balpha^{(0)}$ derived from $\y$ and $\bPhi$, we address the optimization of Eq.~(\ref{eq:dual-domain-fuction}) by decoupling it into a $\z$-subproblem in Eq.~(\ref{eq:z_subproblem}, ID) and an $\balpha$-subproblem in Eq.~(\ref{eq:alpha_subproblem}, CCD). The entire optimization process consists of $T$ iterations, where each iteration is formulated as follows:
\begin{subequations} \label{eq:z_and_alpha}
\begin{align}
\label{eq:z_subproblem}
\z^{(t)} = \underset{\z}{\arg\min}~&\frac{1}{2}\|\bPhi\z-\y\|_2^2+\frac{\mu_\z}{2}\|\z-\D\circledast\balpha^{(t-1)}\|_2^2 \nonumber \\
&~+\lambda\phi(\z),\\
\label{eq:alpha_subproblem}
\balpha^{(t)} = \underset{\balpha}{\arg\min}~&\frac{\mu_\z}{2}\|\D\circledast\balpha-\z^{(t)}\|^2_2+\tau \psi(\balpha).
\end{align}
\end{subequations}
Here, $t$ represents the iteration index. The final recovered result is obtained as $\hat{\x} = \D \circledast \balpha^{(T)}$. For a clear and comprehensive overview of the optimization process, Algo.~\ref{alg:optimization} summarizes the complete framework, which is elaborated upon as follows.

\textbf{(1) Solving Eq.~(\ref{eq:z_subproblem}) via the ID Optimization:} Given $\rho$, $\mu_{\z}$, and initialization $\z^{(0)}=\D \circledast \balpha^{(0)}$, we employ the proximal gradient descent (PGD) method~\cite{parikh2014proximal} to solve the $\z$-subproblem in Eq.~(\ref{eq:z_subproblem}) by iterating between the following two update steps:
\vspace{-10pt}
\begin{subequations}
\begin{align}
\label{eq:GDM}
\tilde{\z}^{(t)} 
    &= \z^{(t-1)}- \rho \left(\bPhi^{\top}\left(\bPhi\z^{(t-1)}-\y\right) \right.\nonumber\\
    &\ \ \ \ \ \ \ \ \ \ \ \ \ +\left.\mu_\z\left(\z^{(t-1)}-\D\circledast\balpha^{(t-1)}\right) \right),\\
\label{eq:PMN}
\z^{(t)} &= \mathrm{prox}_{\lambda\phi}^{\rho}(\tilde{\z}^{(t)})= \underset{\z}{\arg\min}\frac{1}{2\rho}\|\z-\tilde{\z}^{(t)}\|_2^2+\lambda\phi(\z).
\end{align}
\end{subequations}

\vspace{-2pt}
\textbf{(2) Solving Eq.~(\ref{eq:alpha_subproblem}) via the CCD Optimization:} We adopt the HQS algorithm~\cite{he2013half} to separate the data term and the prior term by introducing an auxiliary variable $\tilde{\balpha}$. This leads to the following objective function converted from Eq.~(\ref{eq:alpha_subproblem}):
\begin{equation} \label{eq:after_HQS}
\begin{split}
    \{\balpha^{(t)}, \tilde{\balpha}^{(t)}\}=\underset{\{\balpha, \tilde{\balpha}\}}{\arg\min}~&\frac{\mu_\z}{2}\|\D\circledast \tilde{\balpha} - \z^{(t)}\|_2^2 + \tau\psi(\balpha)\\& + \frac{\mu_{\balpha}  }{2}\|\balpha-\tilde{\balpha}\|_2^2,
\end{split}
\end{equation}
where $\mu_{\balpha}$ is the introduced penalty parameter for the distance between $\balpha$ and $\tilde{\balpha}$. By defining $\eta:=({\mu_{\balpha}}/{\mu_{\z}})$ and $\beta:=({\tau}/{\mu_{\balpha}})$, we solve Eq.~(\ref{eq:after_HQS}) by iterating between the following two steps:
\begin{subequations} \label{eq:data_prior_terms}
\begin{align}
\label{eq:data_term}
    \tilde{\balpha}^{(t)} &= \underset{\balpha}{\arg\min}~\frac{1}{2}\|\D\circledast\balpha-\z^{(t)}\|_2^2+\frac{\eta}{2}\|\balpha-\balpha^{(t-1)}\|_2^2,\\
\label{eq:prior_term}
    \balpha^{(t)} &= \underset{\balpha}{\arg\min} ~\frac{1}{2}\|\balpha - {\tilde{\balpha}}^{(t)}\|_2^2 +
    \beta \psi({\balpha}).
\end{align}
\end{subequations}

\textbf{Convergence Analysis:} It is important and non-trivial to know if our above alternating minimization framework for the dual-domain optimization converges as $t\rightarrow \infty$ \cite{beck2009fast}. We first show that the sequence of dual-domain \mrev{objective} function values in Eq.~(\ref{eq:dual-domain-fuction}) is non-increasing and convergent based on the split steps in Eqs.~(\ref{eq:z_subproblem}) and (\ref{eq:alpha_subproblem}) for solving the ID ($\z$)- and CCD ($\balpha$)-subproblems. Our basic assumptions (or settings) are that the regularizers are bounded as $\phi: \R^{H\times W} \rightarrow [\phi^{*},+\infty)$, $\psi: \R^{C\times H\times W} \rightarrow [\psi^{*},+\infty)$ with non-negative lower bounds $\phi^{*}$ and $\psi^{*}$, respectively, and the hyper-parameters $\{\mu_\z, \lambda, \tau\}$ are all non-negative. Based on these general and reasonable assumptions in the context of optimization and inverse problems \cite{yang2018admm}, we present convergence properties of our dual-domain alternating minimization algorithm by the following theorem:

\vspace{5pt}
\noindent \textbf{Theorem 1 (Convergence of the \mrev{Objective} Function Values).} Denote the \mrev{objective} function in Eq.~(\ref{eq:dual-domain-fuction}) as $F(\z,\balpha)=\frac{1}{2}\|\bPhi\z-\y\|_2^2 +  \frac{\mu_\z}{2}\|\z-\D\circledast\balpha\|_2^2 + \lambda\phi(\z) + \tau\psi(\balpha)$, our Algo.~\ref{alg:optimization} based on the alternating minimization in Eqs.~(\ref{eq:z_subproblem}) and (\ref{eq:alpha_subproblem}) leads to the convergence of sequence $\{F(\z^{(t)},\balpha^{(t)})\}$, \textit{i.e.}, $\forall~t\in\{1,2,\cdots\}$, there is $0\le F(\z^{(t)},\balpha^{(t)}) \le F(\z^{(t-1)},\balpha^{(t-1)})$, and the limit $F^{\infty}=F(\z^{(t)},\balpha^{(t)}) \in [\lambda \phi^*+\tau \psi^*,+\infty)$ exists as $t\rightarrow \infty$.

\vspace{3pt}
\noindent \textit{Proof.} Please refer to Sec. D in the \textbf{\textit{\textcolor{blue}{supplemental material}}} for our complete proof of \textbf{Theorem 1} and more theoretical details.

\vspace{5pt}
The above theorem indicates that our alternating minimization, accomplished by iteratively updating using Eqs.~(\ref{eq:z_subproblem}, ID) and (\ref{eq:alpha_subproblem}, CCD), continuously reduces the \mrev{objective} function value and leads to its convergence as the algorithm progresses. To overcome the intricate process of selecting optimal hyper-parameters and exploit the learning capacity of NNs, we will unfold our algorithm framework into an iterative DUN. This transformation aims to enhance recovery efficiency while preserving a well-defined structure and achieving a balance among performance, speed, and interpretability. While \textbf{Theorem 1} presents fundamental properties of our dual-domain algorithm, here we may refrain from delving into an extensive analysis of convergence rates \cite{beck2009fast}, as the current theoretical results suffice to demonstrate the effectiveness and reasonableness of our dual-domain optimization decoupling. Experiments will exhibit the significant superiority of our dual-domain DUN, inspired by this framework, when compared to single-domain (ID/CCD)-based approaches. Moreover, they will provide valuable insights and an understanding of its underlying working principles.
\vspace{-5pt}

\begin{figure*}[t]
  \centering
  \includegraphics[width=\textwidth]{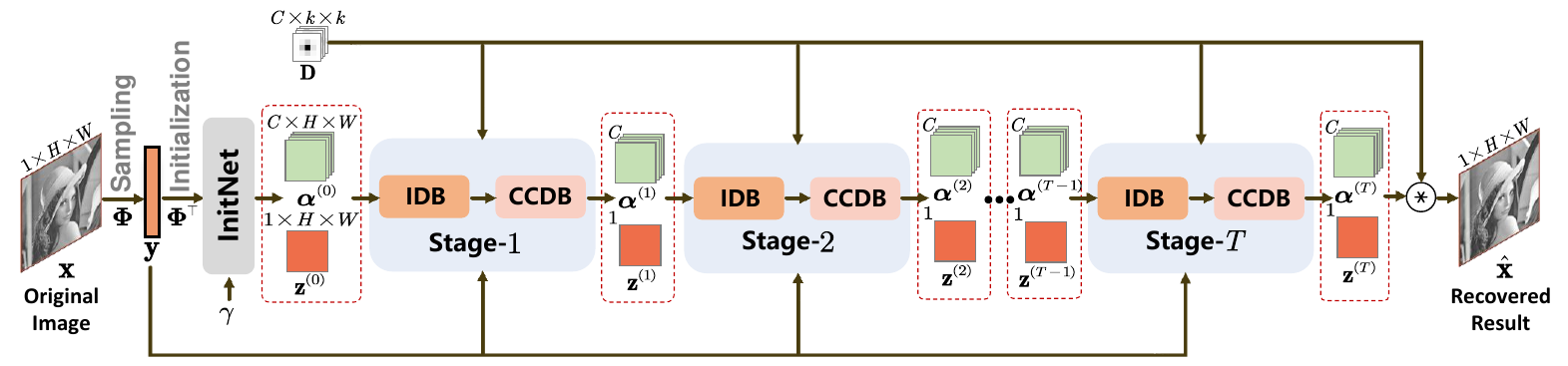}
  \vspace{-15pt}
  \caption{Illustration of the comprehensive architecture of our $\DC$-Net, which encompasses $T$ stages, unfolded from our dual-domain decoupled Algo.~\ref{alg:optimization}. Each stage comprises an image-domain block (IDB) and a convolutional-coding-domain block (CCDB). Within the framework, $\x$ represents the fully-sampled image, $\y$ denotes the under-sampled measurement, and $\hat{\x}$ signifies the final output of $\DC$-Net. The convolutional dictionary and coefficients are denoted by $\D$ and $\balpha$ respectively. Here, $k$ denotes the filter size of $\D$, $H$ and $W$ refer to the height and width of $\x$ and $\balpha$, and $C$ represents the number of channels.}
  \label{fig:network}
\end{figure*}

\begin{figure*}[t]
  \centering
  \includegraphics[width=\textwidth]{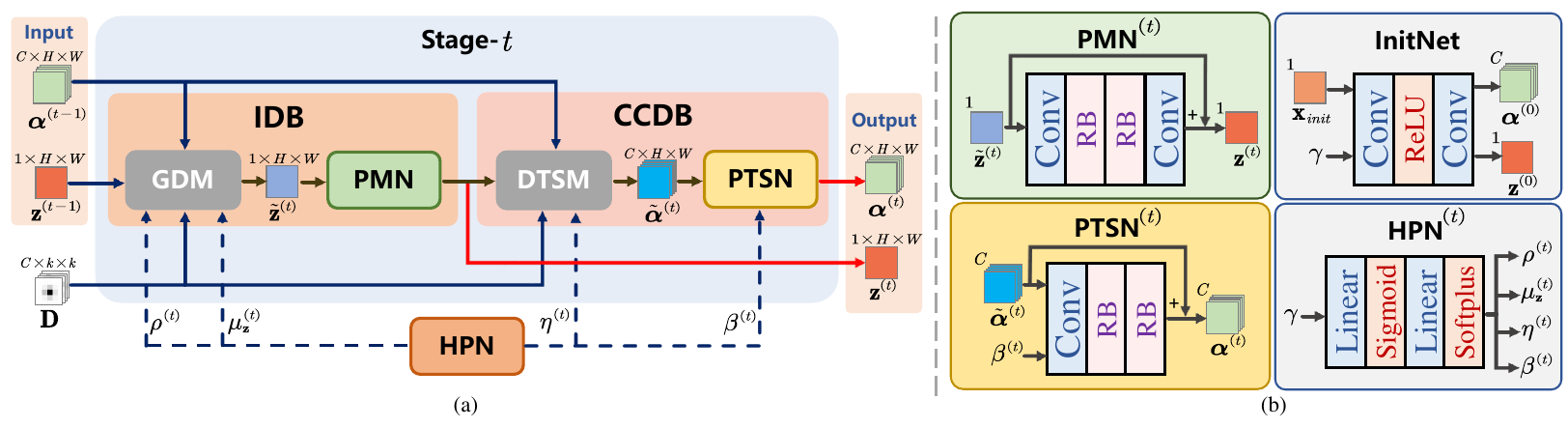}
  \vspace{-20pt}
  \caption{Illustration of the structural design of our unfolded $\DC$-Net stage and its constituent components. In \textcolor{blue}{(a)}, we present the configuration of the $t$-th stage within $\DC$-Net. An image-domain block (IDB) comprises a gradient descent module (GDM) and a proximal mapping network (PMN). Similarly, a convolutional-coding-domain block (CCDB) incorporates a data-term solving module (DTSM) and a prior-term solving network (PTSN). The specifics of the analytic GDM and DTSM are illustrated by Eqs.~(\ref{eq:GDM_impl}) and (\ref{eq:DTSM_impl}) respectively. In \textcolor{blue}{(b)}, we provide the architectural design of the four sub-networks.}
  \label{fig:stage}
\end{figure*}

\subsection{Dual-Domain Unfolding Architecture Design} \label{section:3.3}
Based on Algo.~\ref{alg:optimization}, we develop a dual-domain optimization-inspired DUN, dubbed $\DC$-Net. As illustrated in Fig.~\ref{fig:network}, it consists of a series of $T$ iterative stages, where each stage comprises an image-domain block (IDB) and a convolutional-coding-domain block (CCDB), derived from the Eqs. (\ref{eq:z_subproblem}) and (\ref{eq:alpha_subproblem}), respectively. In this context, $\x \in \R^{1\times H\times W}$ denotes the latent clean original image, $\y$ represents the under-sampled CS measurement at sampling ratio $\gamma$, while the final reconstructed output is obtained as $\hat{\x}=\D\circledast\balpha^{(T)}$. The adaptively learnable convolutional dictionary is denoted as $\D$, and $\balpha$ corresponds to the feature map of convolutional coefficients. Unlike most existing ID-based DUNs, $\DC$-Net incorporates the transmission of an intermediate image $\z^{(t)}$ and a $C$-channel feature-level coefficient map $\balpha^{(t)}$ between each two adjacent stages. This dual-domain algorithm and its unfolding design naturally enable the network to maximize and adaptively process high-throughput information during the NN reconstruction process.

Fig.~\ref{fig:stage} (a) illustrates the stage structure, where each IDB is composed of a gradient descent module (GDM) in conjunction followed by a proximal mapping network (PMN), while each CCDB integrates a data-term solving module (DTSM) with a prior-term solving network (PTSN). Considering the non-trivial challenge of determining the optimal hyper-parameter settings for $\left\{\rho, \mu_\z, \eta, \beta \right\}$, here we draw inspiration from \cite{zheng2021deep}, employing a lightweight fully-connected hyper-parameter network (HPN) to adaptively generate the values from $\gamma$ for each stage. Fig.~\ref{fig:stage} (b) and Algo. \ref{alg:network} summarizes the sub-network architectures and unfolded $\DC$-Net recovery, implemented by the initialization and iterative processing of InitNet, PMN, PTSN, and HPN, which we further elaborate upon as follows.

\textbf{(1) Implementing the Initialization Process:} InitNet takes $\x_{{init}}=\bPhi^\top \y$ and $\gamma$ as input to initialize feature and image as:
\begin{align}
\label{eq:InitNet_impl}
\balpha^{(0)}&=\mathcal{G}_{\mathrm{InitNet}}\left(\x_{{init}},\gamma\right) \in \R^{C\times H\times W}\nonumber\\
&=\mathrm{Conv_2}(\mathrm{ReLU}(\mathrm{Conv_1}(\mathrm{Concat}(\x_{{init}}, \gamma)))),\\
\label{eq:z_Init}
\z^{(0)}&=\D\circledast\balpha^{(0)}.
\end{align}
Specifically:
\begin{itemize}
    \item $\gamma$ is CS ratio map with the same dimension as $\x$, all entries of which are filled by the value of CS ratio and implemented by the \texttt{torch.repeat} API of PyTorch~\cite{paszke2019pytorch} framework.
    \item $\mathrm{Conv_1}$ takes a 2-channel joint input (\textit{i.e.}, the channel-wise concatenation of $\x_{{init}} \in \R^{1\times H \times W}$ and $\gamma \in \R^{1\times H \times W}$) and generates a $C$-channel output with a ReLU~\cite{nair2010rectified} activation.
    \item $\mathrm{Conv_2}$ finally generates a $C$-channel feature output $\balpha^{(0)}$.
    \item $\z^{(0)}$ is generated by a 2D convolution on $\balpha^{(0)}$ with $\D$.
\end{itemize}

\textbf{(2) Predicting Hyper-Parameters via Hyper-Parameter Network (HPN):} In each unfolded stage, HPN takes the CS ratio $\gamma$ as input and predicts the four hyper-parameters as:
\begin{align}
\begin{split}
\label{eq:HPN_impl}
    \left(\rho^{(t)}, \right. &\left.\mu_{\z}^{(t)}, \eta^{(t)}, \beta^{(t)}\right) = \mathcal{G}_{\mathrm{HPN}}^{(t)}\left(\gamma\right) \in \R^{4}\\
    &=\mathrm{Softplus}(\mathrm{Linear_2}(\mathrm{Sigmoid}(\mathrm{Linear_1}(\gamma)))).
\end{split}
\end{align}
Specifically:
\begin{itemize}
    \item $\mathrm{Linear_1}$ takes the CS ratio $\gamma$ as its input and generates a 256-dimensional output with a Sigmoid activation.
    \item $\mathrm{Linear_2}$ then generates four hyper-parameters with a Softplus activation, ensuring all output values are positive.
\end{itemize}

\textbf{(3) Implementing Eq.~(\ref{eq:GDM}) via Gradient Descent Module (GDM):} GDM takes the learned $\balpha^{(t-1)}$, $\z^{(t-1)}$, $\rho^{(t)}$ and $\mu_\z^{(t)}$ as input, and performs the analytic ID gradient descent as:
\begin{align}
\begin{split}
\label{eq:GDM_impl}
    \tilde{\z}^{(t)} 
    &= \mathcal{G}_{\mathrm{GDM}}\left(\balpha^{(t-1)}, \z^{(t-1)}, \rho^{(t)}, \mu_\z^{(t)}\right) \in \R ^{1\times H\times W}\\
    &= \z^{(t-1)}- \rho^{(t)} \left(\bPhi^{\top}\left(\bPhi\z^{(t-1)}-\y\right) \right.\\
    &\ \ \ \ \ \ \ \ \ \ \ \ \ +\left.\mu_\z^{(t)}\left(\z^{(t-1)}-\D\circledast\balpha^{(t-1)}\right) \right)\\
\end{split}
\end{align}

\textbf{(4) Implementing Eq.~(\ref{eq:PMN}) via Proximal Mapping Network (PMN):} PMN solves the ID proximal mapping problem $\mathrm{prox}_{\lambda\phi}^{\rho}(\tilde{\z}^{(t)})$ by a denoising network and is formulated as:
\begin{align}
\begin{split}
\label{eq:PMN_impl}
    \z^{(t)}&= \mathcal{G}_{\mathrm{PMN}}^{(t)}\left(\tilde{\z}^{(t)}\right) \in \R ^{1\times H\times W}\\
    &=\tilde{\z}^{(t)}+\mathrm{Conv_2}(\mathrm{RB_2}(\mathrm{RB_1}(\mathrm{Conv_1}(\tilde{\z}^{(t)})))).
\end{split}
\end{align}
Specifically:
\begin{itemize}
    \item $\mathrm{Conv_1}$ takes $\tilde{\z}^{(t)}$ as input and generates a $C$-channel output.
    \item $\mathrm{RB_1}$ and $\mathrm{RB_2}$ are two residual blocks. Each residual block takes a $C$-channel input and generates a $C$-channel residual output by the classic structure of Conv-ReLU-Conv, \textit{i.e.}, there is $\mathrm{RB}(\x)=\x+\mathrm{Conv}(\mathrm{ReLU(\mathrm{Conv}(\x))})$.
    \item $\mathrm{Conv_2}$ finally converts the $C$-channel input feature to an single-channel output $\z^{(t)}$ under a residual learning strategy.
\end{itemize}

\definecolor{OursColor}{HTML}{FFFFA5}

\newcolumntype{I}{!{\vrule}}

\textbf{(5) Implementing Eq.~(\ref{eq:data_term}) via Data-Term Solving Module (DTSM):} Employing the Fast Fourier Transform (FFT) under the assumption of circular boundary conditions facilitates the solution of Eq.(\ref{eq:data_term}). Let $\mathbfcal{D}=\mathcal{F}(\D)$, $\mathbfcal{Z}^{(t)}=\mathcal{F}(\z^{(t)})$, and $\mathbfcal{A}^{(t-1)}=\mathcal{F}(\balpha^{(t-1)})$, where $\mathcal{F}$ signifies the \mrev{slice-wise} 2D FFT. We have the following closed-form data-term solution:
\begin{align}
\begin{split}
\label{eq:DTSM_impl}
&\tilde{\balpha}^{(t)} = \mathcal{G}_{\mathrm{DTSM}}\left(\balpha^{(t-1)}, \z^{(t)}, \D, \eta^{(t)}\right) \in \R ^{C\times H\times W}\\
&=\frac{1}{\eta^{(t)}}\mathcal{F}^{-1} \left(\mathbfcal{H}^{(t)}-\mathbfcal{D}\circ\left(\frac{\left(\mathbfcal{\bar D} \odot \mathbfcal{H}^{(t)}\right)}{\eta^{(t)}+\left(\mathbfcal{\bar D}\odot\mathbfcal{D}\right)}\uparrow_C  \right) \right).
\end{split}
\end{align}
Here, $\circ$ is the Hadamard product, $\mathbf{X}\odot\mathbf{Y}=\sum_{i=1}^{C}\left(\mathbf{X}_i\circ\mathbf{Y}_i\right)$, $\mathbf{X}\uparrow_C$ expands the channel dimension of $\mathbf{X}$ from 1 to $C$ \mrev{by filling all $C$ feature channels with the value of single-channel $\mathbf{X}$}, $\div$ is the Hadamard division, $\mathcal{F}^{-1}$ denotes the inverse of FFT, $\mathbfcal{\bar D}$ denotes the complex conjugate of $\mathbfcal{D}$, and the intermediate $\mathbfcal{H}^{(t)}$ component is defined as $\mathbfcal{H}^{(t)}=\mathbfcal{D}\circ\left(\mathbfcal{Z}^{(t)}\uparrow_C\right)+\eta\mathbfcal{A}^{(t-1)}$.

\textbf{(6) Implementing Eq.~(\ref{eq:prior_term}) via Prior-Term Solving~Network (PTSN):} PTSN takes features $\tilde{\balpha}^{(t)}$ and $\beta^{(t)}$ as input to adaptively learn the implicit prior for CCD denoising as:
\begin{align}
\begin{split}
\label{eq:PTSN_impl}
    \balpha^{(t)}&= \mathcal{G}_{\mathrm{PTSN}}^{(t)}\left(\tilde{\balpha}^{(t)}, \beta^{(t)}\right) \in \R ^{C\times H\times W}\\
    &=\tilde{\balpha}^{(t)}+\mathrm{RB_2}(\mathrm{RB_1}(\mathrm{Conv_1}(\mathrm{Concat}(\tilde{\balpha}^{(t)}, \beta^{(t)})))).
\end{split}
\end{align}
Specifically:
\begin{itemize}
    \item \mrev{$\beta^{(t)}$ is now represented as a map with same dimension as $\z$. All entries of the new map are filled by the original value of scalar $\beta^{(t)}$ and implemented by the \texttt{torch.repeat} API of PyTorch~\cite{paszke2019pytorch}. Despite the change in representation, the symbol $\beta^{(t)}$ is retained for consistency without confusion.}
    \item $\mathrm{Conv_1}$ takes a $(C+1)$-channel input (\textit{i.e.}, the channel-wise concatenation of $\tilde{\balpha}^{(t)} \in \R^{C\times H \times W}$ and $\beta^{(t)} \in \R^{1\times H \times W}$) and generates a $C$-channel intermediate output.
    \item Two residual blocks $\mathrm{RB_1}$ and $\mathrm{RB_2}$ are then used to extract deep representations as similar to the two RBs in PMN. 
    \item The feature residual learning strategy is employed in PTSN.
\end{itemize}

\subsection{Relationship to Other Works}
In this subsection, we delve into more specific details of the conceptual connection and relationship analysis of our method with other existing works\footnote{For enhanced clarity, we provide a detailed conceptual comparison in the \textbf{\textit{\textcolor{blue}{supplemental material}}}. Please refer to Tab.~\textcolor{red}{VII} of Sec.~\textcolor{red}{A}.}.

\mrev{\textbf{Compared to Existing DNUNs:} Compared with the prior DNUN methods~\cite{kulkarni2016reconnet, sun2020dual, chen2020learning, shi2019image, shi2019scalable}, which aim to learn the inverse mapping $\y \mapsto \x$ through some carefully designed end-to-end ``\textit{black-box}'' NNs. Our method leverages the benefits of both networks and optimization algorithms, thus achieving well-defined interpretability and superior reconstruction accuracy.}

\textbf{Compared to Existing DUNs:} \mrev{Compared with existing DUN methods~\cite{zhang2020optimization, zhang2018ista, you2021ista, zheng2021deep, zhang2020amp, you2021coast, song2021memory, shen2022transcs, chen2022cas, gan2023learned,zhuang2023ucsnet,zhou2023iterative,shi2020video,zhao2016video},} our framework distinguishes itself by formulating the original objective function into an integrated ID- and CCD-optimization. Notably, ISTA-Net$^+$ \cite{zhang2018ista} and ISTA-Net$^{++}$ \cite{you2021ista} emerge as special cases of our approach focusing solely on the ID optimization. In contrast to the recent image denoising method DCDicL~\cite{zheng2021deep}, our $\DC$-Net exhibits three distinctive features: \textcolor{blue}{(1)} Firstly, DCDicL represents a particular instance of our method tailored for image denoising, where the sampling matrix is assumed to be the identity matrix, \textit{i.e.}, $\bPhi =\mathbf{I}_N$. It is also important to note that generalizing DCDicL to CS tasks poses challenges due to the incorporation of non-invertible matrix $\bPhi$ into the optimization model and the potential for numerical instability in implementation. Conversely, $\DC$-Net can be extended to other inverse imaging problems. \textcolor{blue}{(2)} Secondly, while DCDicL exclusively optimizes in CCD, our $\DC$-Net leverages the strengths of both the ID and CCD priors to impose additional constraints on the feasible solution space, which is verified to be crucial for ensuring high performance. \textcolor{blue}{(3)} Lastly, $\DC$-Net employs a universal dictionary as opposed to the adaptive and dynamic one utilized in DCDicL~\cite{zheng2021deep}. This design effectively mitigates instability and collapse during the learning process, enhances training speed and convergence, and can align more closely with the original definition of dictionary learning.

\textbf{Compared to Other Dual-Domain Methods:} Most existing dual-domain CT methods~\cite{wu2021drone, lin2019dudonet} operate within the image ($\x$) and measurement ($\y$) domains, while the MRI methods~\cite{ran2020md, cheng2019model, eo2018kiki} are constructed in k-space and spatial domains. Additionally, these methods only transmit single-channel images among modules. In contrast, our $\DC$-Net can fundamentally diverge from these approaches by establishing connections between the ID ($\x$) and CCD ($\balpha$). This distinctive characteristic enables the transmission of high-throughput feature-level representations and the adaptive capture of features. \mrev{By integrating the strengths of these existing methods, our framework exhibits versatility, empowering $\DC$-Net to flexibly transmit high-capacity coefficients.} The superior performance and generalization ability of our method will be further verified, discussed, and analyzed in the following experimental section.

Moreover, while many existing studies primarily focus on stacking NN modules and rely on some numerical simulations for evaluation, leaving performance drops when it comes to real-world implementations, our work distinguishes itself by offering not only theoretical analysis but also validation on a real established SPI optics system. This significant step forward paves the way for practical CS deployment, thus expanding its potential applications in both daily life and scientific research.

\section{Experiments}
\label{sec:experiments}
\subsection{Implementation Details} 
\label{sec:4.1}
\textbf{Sampling and Initialization Schemes:} In our experiments, we follow~\cite{zhang2020optimization, you2021ista} to adopt the block-based CS setup and simulate CS sampling and initialization processes using bias-free convolutions. Unlike the conventional approach of using split and vectorized image blocks, we employ the entire image $\x\in\R^{1\times H \times W}$ in our implementation, where $H$ and $W$ are multiples of $\sqrt{N}$. In particular, we set block size $N = 1024$. For CS sampling, we reshape the sensing matrix $\bPhi\in\R^{M\times 1024}$ into $M$ filters with kernel size $1\times 32\times 32$. The block-wise sampling process $\y=\bPhi\x$ can be equivalently achieved by a convolution layer with stride 32. Each image $\x \in \R^{1 \times H\times W}$ can be considered as $(H/32) \times (W/32)$ non-overlapping image blocks of size $1\times 32 \times 32$. The resulting measurement $\y$ is a tensor of size $M \times (H/32) \times (W/32)$ after sampling. Moreover, block-wise initialization $\x_{{init}} = \bPhi^\top \y$ is implemented using bias-free transposed convolutions, with the same kernel weights as $\bPhi$. As a result, $\x_{{init}}$ is an image tensor of the same size $1 \times H \times W$ as $\x$. During the learning of $\DC$-Net, random patches of size $96 \times 96$ are cropped and utilized as training samples. Thus, $H$ and $W$ are set to 96 for network training.

\textbf{Network Parameters and Loss Function:} In the case of Fixed Random Gaussian Matrix (\textcolor{blue}{FRGM}), \mrev{the sampling matrix $\bPhi$ undergoes an initial random initialization, followed by an orthogonality process, ensuring that $\bPhi\bPhi^\top=\mathbf{I}_{M}$. During the entire training phase, $\bPhi$ remains constant.} The parameter set of $\DC$-Net, denoted by $\mathbf{\Theta}_{\mathrm{FRGM}}$, is composed of multiple components that can be collaboratively learned and expressed as $\mathbf{\Theta}_{\mathrm{FRGM}} = \{\D, \mathcal{G}_{\mathrm{InitNet}}\}\cup \{\mathcal{G}_{\mathrm{PMN}}^{(t)}, \mathcal{G}_{\mathrm{PTSN}}^{(t)}, \mathcal{G}_{\mathrm{HPN}}^{(t)}\}_{t=1}^{T}$. To minimize the discrepancy between images $\x_j$ and $\hat{\x}_j$, an $\ell_2$-loss is defined using mean squared error (MSE) as $\mathcal{L}_{disc}=[{1}/{(NN_b)}]\sum_{j=1}^{N_b}\|\hat{\x}_j-\x_j\|_F^2$, where $N_b$ and $N$ represent the number of samples in each training batch and the size of each image, respectively. Consequently, the overall loss function is defined as $\mathcal{L}(\mathbf{\Theta}_{\mathrm{FRGM}}) = \mathcal{L}_{disc}$. In the case of Data-driven Adaptive Learned Matrix (\textcolor{blue}{DALM}), $\bPhi$ is \mrev{initialized as a random Gaussian matrix} and considered as a learnable component. \mrev{During the training phase, $\bPhi$ is optimized jointly with the recovery network in an end-to-end manner}, and thus the parameters can be expressed as $\mathbf{\Theta}_{\mathrm{DALM}} = \{\bPhi\} \cup \mathbf{\Theta}_{\mathrm{FRGM}}$. An orthogonal constraint is applied to $\bPhi$ by using $\mathcal{L}_{orth}=({1}/{M^2})\|\bPhi\bPhi^\top-\mathbf{I}_M\|_F^2$. The overall loss function is defined as $\mathcal{L}(\mathbf{\Theta}_{\mathrm{DALM}}) = \mathcal{L}_{disc} + \xi \mathcal{L}_{orth}$, where $\xi$ is the regularization parameter fixed to $0.01$.

\textbf{Network Training and Implementation:} Following~\cite{zheng2021deep}, we use the combination of BSD400~\cite{martin2001database, chen2016trainable}, DIV2K training set~\cite{timofte2017ntire}, and WED~\cite{ma2016waterloo} for NN training. To create training data samples, we extract the luminance component from each image block. Data augmentation is employed to enhance the data diversity. Our $\DC$-Net is implemented with PyTorch~\cite{paszke2019pytorch}. All experiments are conducted on \mrev{a workstation with AMD EPYC 7302 16-Core CPU and NVIDIA GeForce RTX 3090 GPU}. We employ Adam to update the parameters with batch size 32. The network is trained for $2.8\times10^5$ iterations, starting with learning rate $1\times10^{-4}$, which is reduced by 0.1 after $1.6\times10^5$ and $2.4\times10^5$ iterations. The default filter size $k$ and number $C$ for convolutional dictionary $\D$ \mrev{are} set to 5 and $C$, while the numbers of feature maps $C$ and stages $T$ are set to 64 and 8, respectively. \mrev{Our study regarding $k$, $C$, and $T$ of $\DC$-Net can be found in Sec.~\ref{section:4.2}.}

\mrev{\textbf{Enhanced $\DC$-Net Version ($\DC$-Net$^+$):} Firstly, following~\cite{chen2022cas, chen2023deep, song2023deep, song2023optimization}, we observe that the standard $\DC$-Net still has huge potential for improvement. Hence, we make improvements to the network structure and propose an enhanced version $\DC$-Net$^+$. Specifically, similar to MADUN~\cite{song2021memory} and MAPUN~\cite{song2023deep}, we set $C=32$ and $T=25$ for fair comparison. These settings also make NN's effective receptive field larger and the structure more compact. Secondly, following~\cite{chen2023deep, song2023optimization}, we improve training strategies. To be concrete, we replace the original $\ell_2$-loss with a Charbonnier loss~\cite{lai2017deep} $\mathcal{L}_{disc}=[{1}/{(NN_b)}]\sum_{j=1}^{N_b}\sqrt{\|\hat{\x}_j-\x_j\|_F^2 + \epsilon}$, where $\epsilon=1\times 10^{-4}$. The size of training patches is increased from $96 \times 96$ to $128 \times 128$~\cite{chen2022cas,song2023deep, chen2023deep}. Our $\DC$-Net$^+$ is trained for $9 \times 10^5$ iterations, starting with learning rate $1\times 10^{-4}$, which is reduced by 0.1 after $6\times 10^5$ and $8\times 10^5$ iterations.}

\begin{figure}[t!]
    \centering
    \scriptsize
    \hspace{-8pt}
    \setlength{\tabcolsep}{0.2pt}
    \renewcommand{\arraystretch}{0.4}
    \scalebox{1.05}{
    \begin{tabular}{ccc}
         \includegraphics[width=0.31\linewidth]{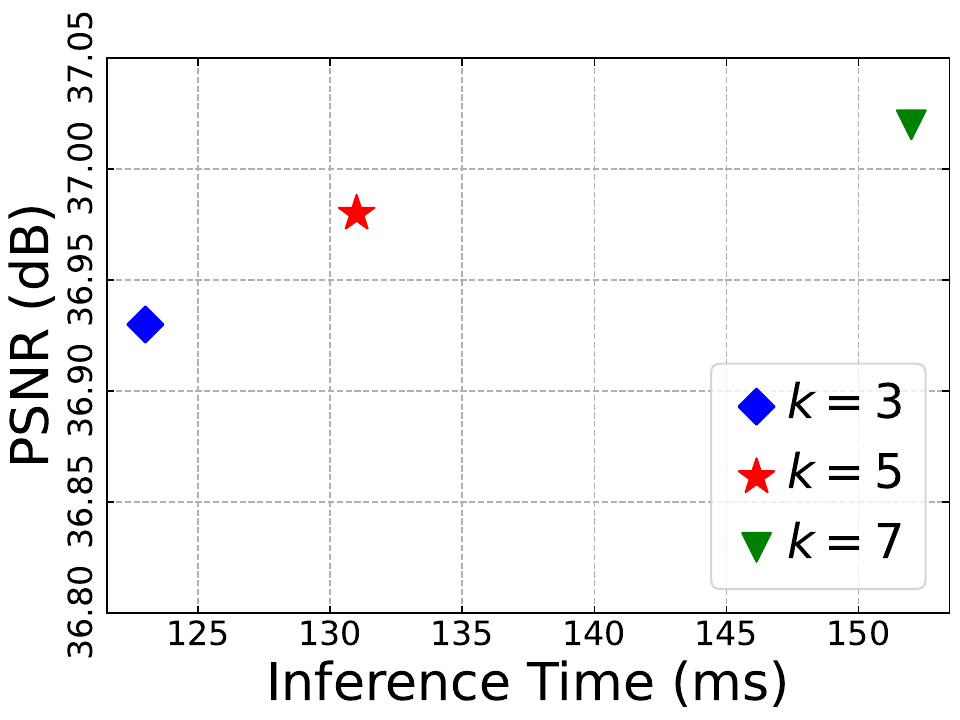}&\includegraphics[width=0.31\linewidth]{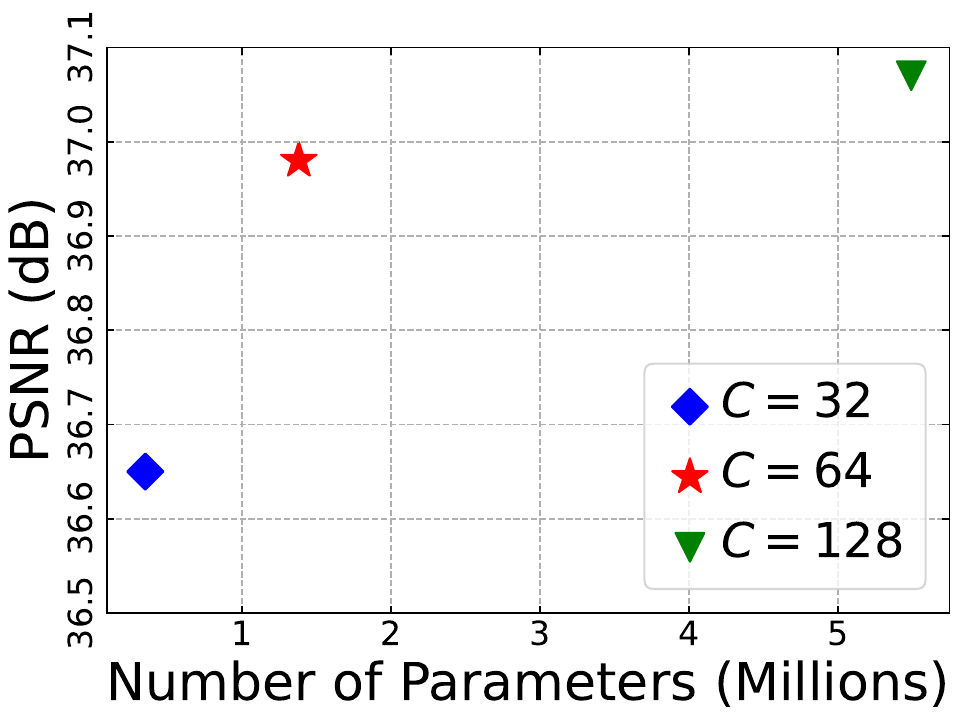}&\includegraphics[width=0.31\linewidth]{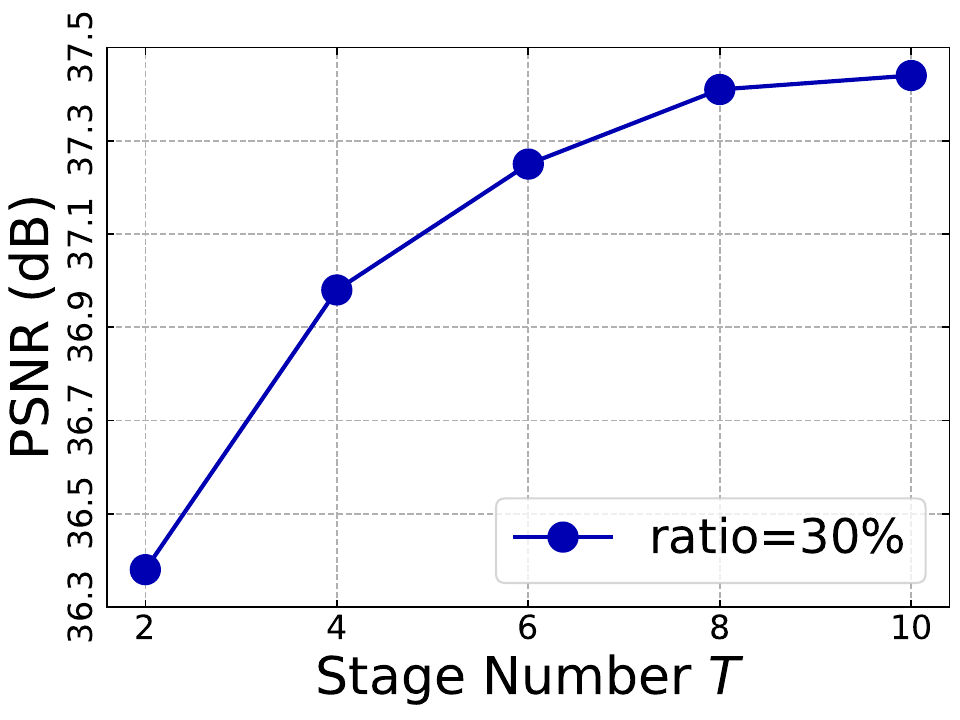} \\
         \hspace{8pt}(a)&\hspace{8pt} (b) &\hspace{8pt}(c)
    \end{tabular}}
    \vspace{-3pt}
    \caption{Ablation studies on \textcolor{blue}{(a)} dictionary filter size $k$, \textcolor{blue}{(b)} feature number $C$, \textcolor{blue}{(c)} unfolded stage number $T$. All experiments are performed on Set11 \cite{kulkarni2016reconnet} with CS ratio $\gamma=30\%$.}
    \label{fig:ablation}
\end{figure}

\vspace{-9pt}
\subsection{\mrev{Study on Hyper-Parameters of Standard $\DC$-Net}}
\label{section:4.2}

In this section, we delve into the selection of key hyper-parameters including the filter size $k$ of convolutional dictionary $\D$, the feature map number $C$, and the stage number $T$ \mrev{of our standard $\DC$-Net. All experiments are performed on the Set11~\cite{kulkarni2016reconnet} benchmark with CS ratio $\gamma=30\%$.}

\textbf{Filter Size $k$ of the Dictionary:} We begin by examining the impact of filter size of $\D$, considering values of $k$ from the set $\{3,5,7\}$. As demonstrated in Fig.~\ref{fig:ablation}~(a), larger values of $k$ yield improved recovery performance at the cost of increased inference time. To obtain a balance between performance and efficiency, we adopt $k=5$ for our $\DC$-Nets by default.

\textbf{Number of Feature Maps $C$:} We evaluate the effect of the number of feature maps by exploring values of $C$ from the set $\{32,64,128\}$. Fig.~\ref{fig:ablation}~(b) presents a comparison of PSNR and parameter number across different $C$ values. With increasing $C$, the throughput of $\DC$-Net improves, leading to enhanced reconstruction performance. However, a surplus of feature maps introduces excessive network parameters, making it challenging to achieve adequate training. To achieve a trade-off between reconstruction performance and computational complexity, we opt for a default setting of $C=64$ in our standard $\DC$-Net.

\textbf{Number of Unfolded Stages $T$:} The stage number $T$ in $\DC$-Net corresponds to the number of unfolded iterations (Algo.~\ref{alg:network}) in our dual-domain optimization framework (Algo.~\ref{alg:optimization}). A larger value of $T$ is expected to yield improved reconstruction accuracy. Fig.~\ref{fig:ablation}~(c) examines the performance of different $\DC$-Net variants with $T$ values ranging from 2 to 10. We observe that the PSNR increases with $T$, but the gains become marginal when $T \geq 8$. Considering the trade-off between final recovery accuracy and efficiency, we select $T=8$ as the default setting.

\subsection{Ablation Studies and Discussions}

\textbf{Effect of High-Throughput Transmission:} To investigate the influence of CCD recovery on NN's transmission capability, starting from our $\DC$-Net, we introduce modifications to the network throughput between stages by adjusting the value of $C$ in Fig.~\ref{fig1:ccmodel}. Specifically, since our network transmits $(C+1)$ channels of information between each of two adjacent stages, smaller $C$ leads to a low-throughput information flow. It is worth noting that the channel number in the intermediate feature space of PMN and PTSN remains unchanged, resulting in an acceptable minor impact on the parameter number. The results in the left part of Tab.~\ref{tab:Ablation-All} exhibit a decrease in recovery accuracy as $C$ is reduced. Notably, our method achieves a PSNR gain of 1.18dB with only a marginal cost of 0.32M parameters compared to the variant with $C=1$. This significant improvement can be attributed to that the information bottleneck limits the transmission while the high-throughput and high-dimensional CCD propagation enhances NN's capacity, ensuring the final superior performance.

\begin{table}[!t]
    \centering
    \caption{Ablation studies of our various structural designs on Urban100~\cite{huang2015single}, started from the standard $\DC$-Net. \textcolor{blue}{(Left)} We progressively decrease the number of feature maps $C$. \textcolor{blue}{(Top right)} We evaluate \mrev{three} single-domain variants. \textcolor{blue}{(Middle right)} \mrev{We investigate the image domain (ID) regularization.} \textcolor{blue}{(Bottom right)} We evaluate the DNUN variant. The average PSNR (dB) and parameter number (M) are provided, with impacts highlighted in purple. The CS ratio is set to 10\% for the left cases and 30\% for the right cases. Throughout this paper, our method is marked with a yellow background for easy reference.}

    \begin{minipage}{0.49\linewidth}
        \centering
        \resizebox{\linewidth}{!}{%
        \begin{tabular}{c|c|c}
        \shline
        \rowcolor[HTML]{EFEFEF}
            $C$ & PSNR (dB) & \#Param. (M) \\
            \hline \hline
            \rowcolor[HTML]{FFFFA5} 64 & 27.56 & 2.72 \\
            \rowcolor[HTML]{FFFFFF} 32 & 27.33~{\scriptsize(\textcolor{purple}{-0.23})} & 2.70~{\scriptsize(\textcolor{purple}{-0.02})}\\
            \rowcolor[HTML]{FFFFFF} 16 & 27.18~{\scriptsize(\textcolor{purple}{-0.38})} & 2.54~{\scriptsize(\textcolor{purple}{-0.18})}\\
            \rowcolor[HTML]{FFFFFF} 8 & 27.02~{\scriptsize(\textcolor{purple}{-0.54})} & 2.46~{\scriptsize(\textcolor{purple}{-0.26})}\\
            \rowcolor[HTML]{FFFFFF} 4 & 26.90~{\scriptsize(\textcolor{purple}{-0.66})} & 2.43~{\scriptsize(\textcolor{purple}{-0.29})}\\
            \rowcolor[HTML]{FFFFFF} 2 & 26.78~{\scriptsize(\textcolor{purple}{-0.78})} & 2.41~{\scriptsize(\textcolor{purple}{-0.31})}\\
            \rowcolor[HTML]{FFFFFF} 1 & 26.38~{\scriptsize(\textcolor{purple}{-1.18})} & 2.40~{\scriptsize(\textcolor{purple}{-0.32})}\\
             \shline
        \end{tabular}%
        }
    \end{minipage}
    \hfill
    \begin{minipage}{0.48\linewidth}
        \centering
        \resizebox{\linewidth}{!}{%
        \begin{tabular}{c|c|c}
        \shline
        \rowcolor[HTML]{EFEFEF}
            Method & PSNR (dB) & \#Param. (M) \\
            \hline 
            \hline
            \rowcolor[HTML]{FFFFA5} Dual-Domain & 34.06 & 2.72\\
            \rowcolor[HTML]{FFFFFF} {\scriptsize ID-HighThroughput} & 33.48~{\scriptsize(\textcolor{purple}{-0.58})} & 2.97~{\scriptsize(\textcolor{purple}{+0.25})}\\
            \rowcolor[HTML]{FFFFFF} ID-Only & 33.31~{\scriptsize(\textcolor{purple}{-0.75})} & 2.39~{\scriptsize(\textcolor{purple}{-0.33})}\\
            \rowcolor[HTML]{FFFFFF} CCD-Only & 31.94~{\scriptsize(\textcolor{purple}{-2.12})} & 2.72~{\scriptsize(\textcolor{purple}{-0.00})}\\

            \hline \hline
            \rowcolor[HTML]{FFFFA5} PMN-2RB & 34.06 & 2.72\\
            \rowcolor[HTML]{FFFFFF} PMN-3RB & 34.09~{\scriptsize(\textcolor{purple}{+0.03})} & 3.32~{\scriptsize(\textcolor{purple}{+0.60})}\\
            \rowcolor[HTML]{FFFFFF} PMN-4RB & 34.18~{\scriptsize(\textcolor{purple}{+0.12})} & 3.91~{\scriptsize(\textcolor{purple}{+1.19})}\\
            \rowcolor[HTML]{FFFFFF} w/o PMN & 33.73~{\scriptsize(\textcolor{purple}{-0.33})} & 1.54~{\scriptsize(\textcolor{purple}{-1.18})}\\
             
            \hline \hline
            \rowcolor[HTML]{FFFFA5} DUN & 34.06 & 2.72\\
            \rowcolor[HTML]{FFFFFF} DNUN & 31.73~{\scriptsize(\textcolor{purple}{-2.33})} & 2.72~{\scriptsize(\textcolor{purple}{-0.00})}\\

             \shline
        \end{tabular}%
        }
        
    \end{minipage}
    \label{tab:Ablation-All}
\end{table}

\textbf{Effect of Dual-Domain Alternating Recovery}: To evaluate the effectiveness of dual-domain priors-based unfolding design, we compare our $\DC$-Net with two variants based on single domains (ID/CCD). Firstly, we evaluate an ID-only network by removing the CCDB from our $\DC$-Net. This results in $\z$ being the sole input and output at each stage, thus reducing the NN's information transmission capacity. To maintain comparable network depth and parameter number, we increase the number of RBs in the PMN from 2 to 4 in each stage. Secondly, we examine a CCD-only network by eliminating the IDB from our $\DC$-Net and obtaining $\z^{(t)}$ through $\z^{(t)} = \D \circledast \balpha^{(t-1)}$ without incorporating the IDB, thereby excluding the iterative injection of $\y$ and $\bPhi$ knowledge in ID. Similarly, we adjust the number of RBs in the PTSN to maintain consistency in network depth and parameter number. The results are presented in the top right section of Tab.~\ref{tab:Ablation-All}. Although both ID-only and CCD-only variants adopt deep unfolding architectures, the former ID-only variant lacks CCD prior and transmits single-channel image-level information between stages, leading to a weakened information transmission. On the other hand, the latter CCD-only variant lacks explicit guidance from the encouragement of measurement consistency and physics-related data in ID. These two ablations bring significant performance drops of 0.75dB and 2.12dB, respectively, verifying the necessity and feasibility of our dual-domain alternating unfolding recovery.

To further study how much spatial information in the initial estimation $\x_{init}$ is utilized for reconstruction, we conduct a visual comparison of the effective receptive field (ERF)~\cite{luo2016understanding}. Specifically, following~\cite{ding2022scaling}, in the last row of Tab.~\ref{tab:ERF}, we compare ERFs of the three $\DC$-Net variants by computing $(\partial\hat{\x}_i/\partial\x_{{init}})$ of recovered center pixel $\hat{\x}_i$ to initialization $\x_{{init}}$. We observe that our default dual-domain network effectively integrates the merits of both ID-/CCD-only variants. It benefits from the gradient descent module (GDM) and our CCD elimination of information bottlenecks to achieve a larger receptive field and powerful non-linear transformations, thus activating more effective regions to facilitate recovery.

\mrev{Furthermore, we evaluate an ``ID-HighThrought'' variant by enhancing the feature maps' number of $\z$ to $C=64$ from an ID-only network and performing the GDM only on the first channel of $\z$. As shown in the top right section of Tab.~\ref{tab:Ablation-All}, the ID-HighThroughput variant outperforms the ID-only one due to its high-throughput transmission capability, but still leaves the performance gap of 0.58dB. This further demonstrates the effectiveness of our introduction of CCD prior.}

\textbf{Effect of ID Prior-Based Recovery}. To validate the effect of ID regularization (\textit{i.e.}, $\lambda \phi(\z)$), we further enhance the PMN by increasing its RB number. \mrev{We also conduct an experiment of removing the PMN module to validate the necessity of ID prior.} Experiments in the middle right part of Tab.~\ref{tab:Ablation-All} show that \mrev{the ID prior brings a significant gain, and} there exists a large improvement room as the depth of ID sub-networks increases. 

\definecolor{Ourscolor}{HTML}{FFFFA5}
\begin{table}[!t]
    \centering
    \caption{Visual comparisons of reconstructing a benchmark image named ``0887'' \cite{agustsson2017ntire} on three $\DC$-Net variants with $\gamma=30\%$. \mrev{We provide their final recoveries with absolute residual maps to ground truth} \textcolor{blue}{(top)} and visualizations of the effective receptive field (ERF)~\cite{luo2016understanding} showing the utilization of informative regions in initialized results \textcolor{blue}{(bottom)}. \mrev{Throughout this paper, lighter pixels in each residual map signify greater absolute values of prediction error. Arrows are utilized to distinctly denote the differences observed across various results.}}
    \label{tab:ERF}
    \resizebox{1.0\linewidth}{!}{
    \begin{NiceTabular}{lIccc}
    \CodeBefore
    \tikz \fill [gray!10] (1-|1) rectangle (3-|4);
    \tikz \fill [Ourscolor] (1-|4) rectangle (9-|5); 
    \Body
    \shline
     \multicolumn{1}{lI}{\multirow{2}{*}{Deep CS Network}}&
  \multicolumn{1}{c}{\multirow{2}{*}{CCD-Only}} & \multicolumn{1}{c}{\multirow{2}{*}{ID-Only}} & \multicolumn{1}{c}{\multirow{2}{*}{\begin{tabular}[c]{@{}c@{}}\textbf{Dual-Domain}\\ \textbf{(Ours)} \end{tabular} }}\\
& & & \\
      \hline \hline 
      & & &\\[-7pt]
      \raisebox{4.0\height}{Recovered Result $\hat{\x}$}   &  
      \includegraphics[width=0.25\linewidth]{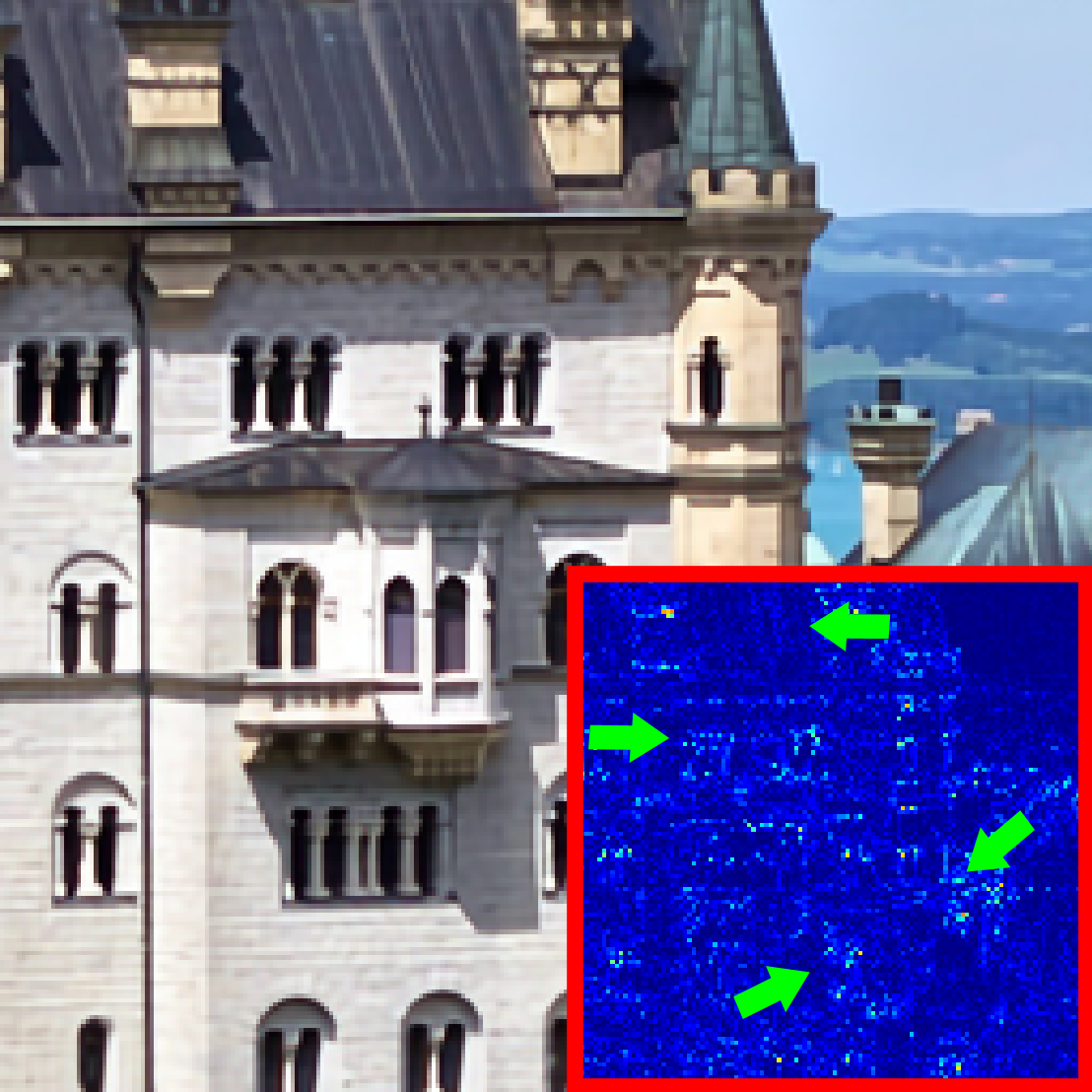} &
      \includegraphics[width=0.25\linewidth]{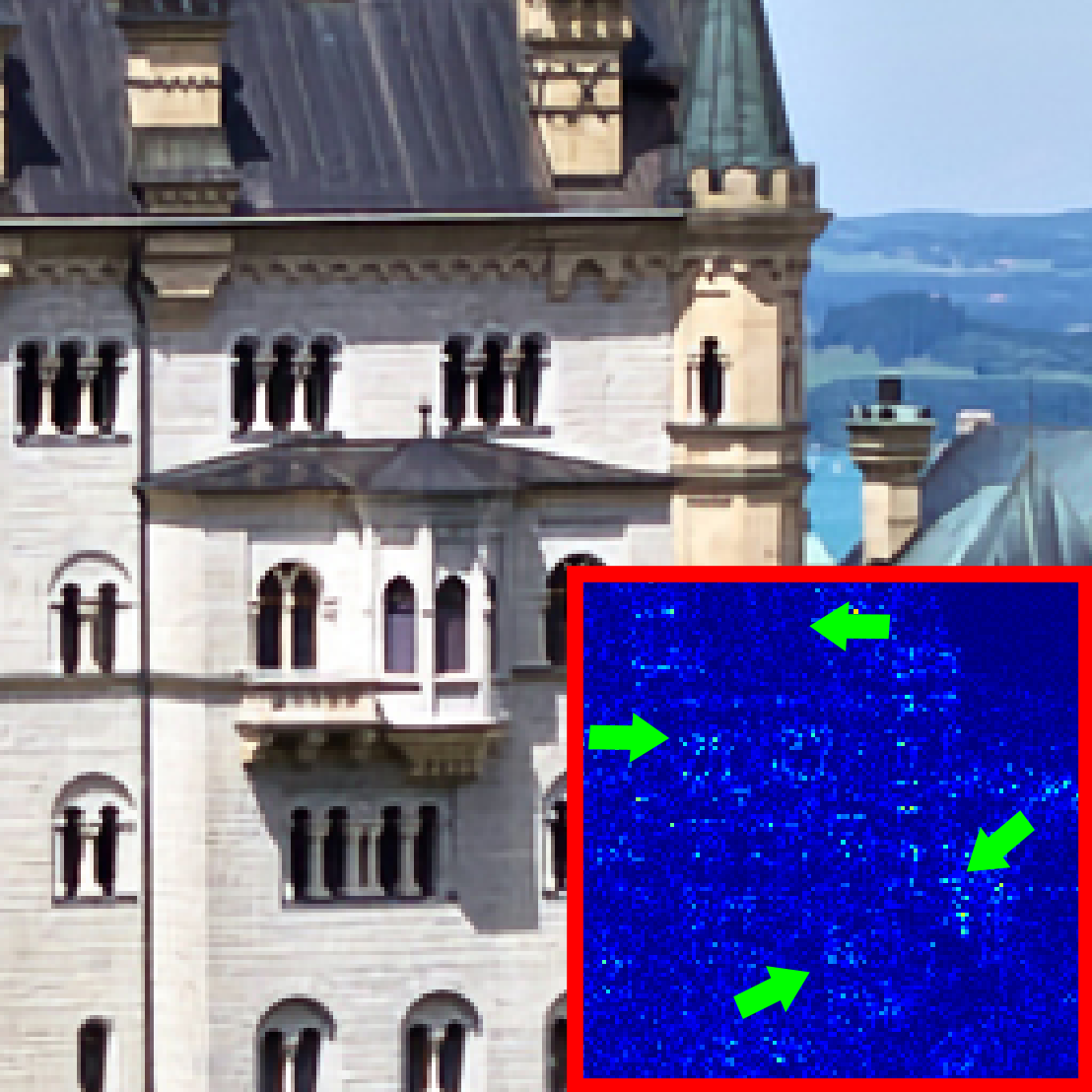} &
      \includegraphics[width=0.25\linewidth]{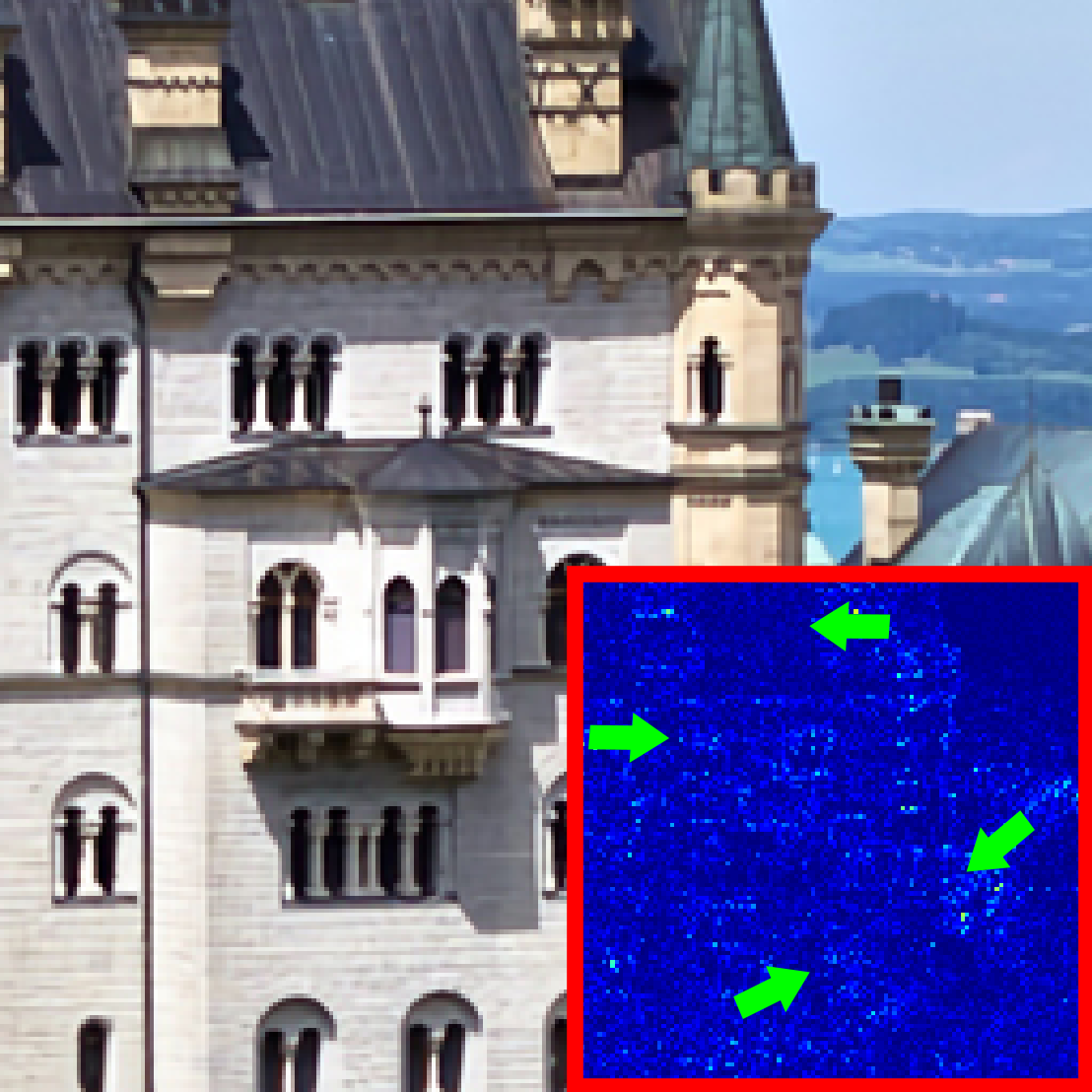} \\
      PSNR~(dB)/SSIM & 31.37/0.9466 & \underline{\textcolor{blue}{32.50}}/\underline{\textcolor{blue}{0.9527}} & \textbf{\textcolor{red}{33.09}}/\textbf{\textcolor{red}{0.9577}} \\ 
      \hline \hline 
      & & &\\[-7pt]
{\multirow{2}{*}[-20pt]{\shortstack{Visualization of $\dfrac{\partial \hat{\mathbf{x}}_i}{\partial \mathbf{x}_{init}}$\\ {\scriptsize(in Logarithm)}}}} &
{\multirow{2}{*}{\includegraphics[width=0.25\linewidth]{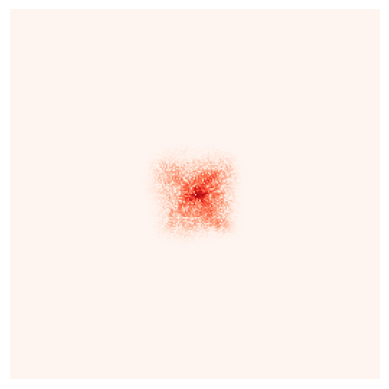}}} &
\raisebox{4.0\height}{{\multirow{2}{*}{\includegraphics[width=0.25\linewidth]{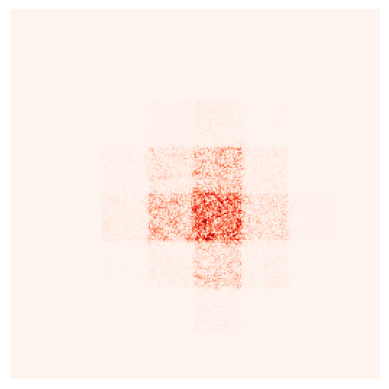}}}} &
\raisebox{4.0\height}{{\multirow{2}{*}{\includegraphics[width=0.25\linewidth]{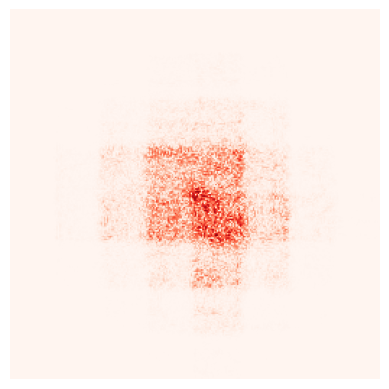}}}} \\
     & & & \\[48pt]
      \shline
    \end{NiceTabular}
    }
\vspace{-5pt}
\end{table}

\textbf{Effect of Deep Algorithm-Unfolding}: To explore the effect of deep unfolding, we introduce a deep non-unfolding variant (DNUN) of our $\DC$-Net by removing the data-terms (GDM and DTSM) in each stage, which can be essentially regarded as a general ``\textit{black-box}" de-aliasing network. Experimental results of ``DUN vs. DNUN" are shown in the bottom right part of Tab.~\ref{tab:Ablation-All}. The notable 2.33dB PSNR decrease brought by DNUN, despite having a close parameter number, demonstrates the efficacy of algorithm-unfolding. It can not only provide a well-defined framework but also explicitly and iteratively incorporate physics- and dictionary-based knowledge into the NN's trunk, thus alleviating the challenges of deep reconstruction learning.

\begin{table*}[t!]
\vspace{-5pt}
  \caption{Comparison of average PSNR (dB), stage number, inference time (ms), GFLOPs, and number of parameters (M) among different CS methods with FRGM and DALM. PSNR is evaluated on Set11~\cite{kulkarni2016reconnet} and Urban100~\cite{huang2015single} datasets with $\gamma \in \{10\%, 30\%, 50\%\}$. More SSIM results regarding natural image CS are provided in our supplemental Sec. C. Throughout this paper, the best and second best results of each setting are highlighted in \textbf{\textcolor{red}{bold red}} and \underline{\textcolor{blue}{underlined blue}}, respectively.}
\centering
\resizebox{1.0\textwidth}{!}{
\begin{minipage}[c]{0.965\textwidth}
\label{tab:compare_sota_psnr}
\resizebox{1.0\textwidth}{!}{
\begin{tabular}{c|l|c|ccc|ccc|c|c|c|c}
\shline
\rowcolor[HTML]{EFEFEF} 
\multicolumn{2}{l|}{\cellcolor[HTML]{EFEFEF}} &
  Test Set &
  \multicolumn{3}{c|}{\cellcolor[HTML]{EFEFEF}Set11 \cite{kulkarni2016reconnet}} &
  \multicolumn{3}{c|}{\cellcolor[HTML]{EFEFEF}Urban100 \cite{huang2015single}} &
  \multicolumn{1}{l|}{\cellcolor[HTML]{EFEFEF}}&
  \multicolumn{1}{l|}{\cellcolor[HTML]{EFEFEF}}&
  \multicolumn{1}{l|}{\cellcolor[HTML]{EFEFEF}}&
  \multicolumn{1}{l}{\cellcolor[HTML]{EFEFEF}}
\\ \cline{3-9}
\rowcolor[HTML]{EFEFEF} 
\multicolumn{2}{l|}{\multirow{-2}{*}{\cellcolor[HTML]{EFEFEF}Method}} &
  CS Ratio $\gamma$ &
  10\% &
  30\% &
  50\% &
  10\% &
  30\% &
  50\% &
  \multicolumn{1}{c|}{\multirow{-2}{*}{\cellcolor[HTML]{EFEFEF}\#Stage}} &
    \multicolumn{1}{c|}{\multirow{-2}{*}{\cellcolor[HTML]{EFEFEF}\begin{tabular}[c]{@{}c@{}}Time\\ (ms) \end{tabular}}}&
\multicolumn{1}{c|}{\multirow{-2}{*}{\cellcolor[HTML]{EFEFEF}GFLOPs}}&
\multicolumn{1}{c}{\multirow{-2}{*}{\cellcolor[HTML]{EFEFEF}
\begin{tabular}[c]{@{}c@{}}\#Param.\\ (M) \end{tabular} }}

 \\ \hline \hline

 \multirow{7}{*}{\rotatebox[origin=c]{90}{FRGM}} &
 \multicolumn{2}{l|}{DPA-Net (TIP 2020) \cite{sun2020dual}}         & 27.66 & 33.60 & -     & 24.00 & 29.04 & - & N/A & 36.52 & - & $9.31 \times 3$    \\
& \multicolumn{2}{l|}{MAC-Net (ECCV 2020) \cite{chen2020learning}}         & 27.92 & 33.87 & 37.76 & 23.71 & 29.03 & 33.10 & 8 & 71.04 & - & $6.12\times 3$ \\
& \multicolumn{2}{l|}{ISTA-Net$^{++}$ (ICME 2021) \cite{you2021ista}} & 28.34 & 34.86 & 38.73 & 24.95 & 31.50 & 35.58 & 20 & 12.35 & 72.44 & $0.76\times 1$\\ 
& \multicolumn{2}{l|}{MADUN (ACM MM 2021) \cite{song2021memory}}         & \msecondbest{29.44} & \msecondbest{36.07} & \msecondbest{39.92} & 26.29 & \msecondbest{33.27} & 37.58 & 25 & 92.15 & 390.03 & $3.12\times 3$ \\
& \multicolumn{2}{l|}{MAPUN (IJCV 2023) \cite{song2023deep}}         & \msecondbest{29.44} & 36.05 & 39.89 & \msecondbest{26.35} & 33.24 & \msecondbest{37.59} & 20 & 91.69 & 345.9 & $2.42\times 3$ \\
& \multicolumn{2}{l|}{\cellcolor[HTML]{FFFFA5}\textbf{$\DC$-Net (Ours)}} &  \cellcolor[HTML]{FFFFA5}28.89   &  \cellcolor[HTML]{FFFFA5}35.41 &\cellcolor[HTML]{FFFFA5}39.14& \cellcolor[HTML]{FFFFA5}26.04 & \cellcolor[HTML]{FFFFA5}32.84 & \cellcolor[HTML]{FFFFA5}36.55 & \cellcolor[HTML]{FFFFA5}8 & \cellcolor[HTML]{FFFFA5}44.50 & \cellcolor[HTML]{FFFFA5}88.74 & \cellcolor[HTML]{FFFFA5}$2.72\times 1$\\
& \multicolumn{2}{l|}{\cellcolor[HTML]{FFFFA5}\textbf{$\DC$-Net$^+$ (Ours)}} &  \cellcolor[HTML]{FFFFA5}\textbf{\textcolor{red}{29.66}}   &  \cellcolor[HTML]{FFFFA5}\textbf{\textcolor{red}{36.12}} &\cellcolor[HTML]{FFFFA5}\textbf{\textcolor{red}{39.93}}& \cellcolor[HTML]{FFFFA5}\textbf{\textcolor{red}{27.25}} & \cellcolor[HTML]{FFFFA5}\textbf{\textcolor{red}{33.72}} & \cellcolor[HTML]{FFFFA5}\textbf{\textcolor{red}{37.62}} & \cellcolor[HTML]{FFFFA5}25 & \cellcolor[HTML]{FFFFA5}58.27 & \cellcolor[HTML]{FFFFA5}68.98 & \cellcolor[HTML]{FFFFA5}$2.12\times 1$\\
\hline \hline
\multirow{12}{*}{\rotatebox[origin=c]{90}{DALM}} 
& \multicolumn{2}{l|}{OPINE-Net$^+$ (JSTSP 2020) \cite{zhang2020optimization}}   & 29.81 & 35.99 & 40.19 & 25.90 & 31.97 & 36.28 & 9 & 17.31 & 36.29 & $0.62\times 3$\\
& \multicolumn{2}{l|}{{\scriptsize GPX-ADMM-Net {\tiny (IEEE Access 2021)}\cite{hu2021gpx}}}   & 28.89 & 35.33 & 39.95 & - & - & - & 10 & 70.00 & - & $0.54 \times 1$\\
& \multicolumn{2}{l|}{AMP-Net (TIP 2021) \cite{zhang2020amp}}         & 29.40 & 36.03 & 40.34 & 25.32 & 31.63 & 35.91 & 9 & 27.38 & 47.93 & $0.58\times 3$\\
& \multicolumn{2}{l|}{COAST (TIP 2021) \cite{you2021coast}}           & 30.02 & 36.33 & 40.33 & 26.17 & 32.48 & 36.56 & 20 & 45.54 & 156.66 & $1.60\times 3$\\
& \multicolumn{2}{l|}{MADUN (ACM MM 2021) \cite{song2021memory}}           & 29.89 & 36.90 & 40.75 & 26.23 & 33.00 & 36.69 & 25 & 92.15 & 390.03 & $3.12\times 3$\\
& \multicolumn{2}{l|}{TransCS (TIP 2022) \cite{shen2022transcs}}          & 29.54 & 35.62 & 40.50 & 25.82 & 31.17 & 36.64 & 6 & 26.00 & 27.1 & $2.03\times 3$\\ 
& \multicolumn{2}{l|}{{\scriptsize QISTA-ImageNet (ECCV 2022) \cite{lin2022qista}}}          & 30.01 & 36.64 & 40.87 & - & - & - & 9 & 86.00 & 100.5 & $0.53\times 3$\\ 
& \multicolumn{2}{l|}{CASNet (TIP 2022) \cite{chen2022cas}}          & 30.36 & 36.92 & 40.93 & 26.84 & 32.85 & 36.95 & 13 & 103.94 & 137.5 & $16.9\times 1$\\ 
& \multicolumn{2}{l|}{MAPUN (IJCV 2023) \cite{song2023deep}}          & 30.19 & 37.08 & 40.98 & 26.44 & 33.28 & 37.00 & 20 & 91.69 & 345.9 & $2.42\times 3$\\ 
& \multicolumn{2}{l|}{OCTUF$^+$ (CVPR 2023) \cite{song2023optimization}}          & 30.73 & 37.32 & \msecondbest{41.35} & 27.28 & 33.87 & 37.82 & 16 & 53.81 & 294.6 & $0.82\times 3$\\ 
& \multicolumn{2}{l|}{\cellcolor[HTML]{FFFFA5}\textbf{$\DC$-Net (Ours)}} &  \cellcolor[HTML]{FFFFA5}\msecondbest{{30.80}}   &  \cellcolor[HTML]{FFFFA5}\msecondbest{{37.41}} & \cellcolor[HTML]{FFFFA5}41.29& \cellcolor[HTML]{FFFFA5}\msecondbest{27.54} & \cellcolor[HTML]{FFFFA5}\msecondbest{34.06} & \cellcolor[HTML]{FFFFA5}\msecondbest{37.89} & \cellcolor[HTML]{FFFFA5}8 & \cellcolor[HTML]{FFFFA5}44.50 & \cellcolor[HTML]{FFFFA5}88.74 & \cellcolor[HTML]{FFFFA5}$2.72\times 1$\\ 
& \multicolumn{2}{l|}{\cellcolor[HTML]{FFFFA5}\textbf{$\DC$-Net$^+$ (Ours)}} &  \cellcolor[HTML]{FFFFA5}\textbf{\textcolor{red}{31.28}}   &  \cellcolor[HTML]{FFFFA5}\textbf{\textcolor{red}{37.67}} & \cellcolor[HTML]{FFFFA5}\textbf{\textcolor{red}{41.65}}& \cellcolor[HTML]{FFFFA5}\textbf{\textcolor{red}{28.38}} & \cellcolor[HTML]{FFFFA5}\textbf{\textcolor{red}{34.67}} & \cellcolor[HTML]{FFFFA5}\textbf{\textcolor{red}{38.61}} & \cellcolor[HTML]{FFFFA5}25 & \cellcolor[HTML]{FFFFA5}58.27 & \cellcolor[HTML]{FFFFA5}68.98 & \cellcolor[HTML]{FFFFA5}$2.12\times 1$\\ \shline
\end{tabular}}
\end{minipage}}
\vspace{-8pt}
\end{table*}

\begin{table}[ht!]
\vspace{-5pt}
\caption{\mrev{Average PSNR (dB)/SSIM comparisons on nature image CS and CS-MRI tasks among different methods at two extremely low ratios $\gamma \in \{1\%, 4\%\}$. The results are evaluated on Set11~\cite{kulkarni2016reconnet} and brain MR images~\cite{clark2013cancer}, respectively.}}
\centering
\resizebox{1.0\linewidth}{!}{
\begin{minipage}[c]{1.0\linewidth}
\label{tab:low_ratios}
\resizebox{0.965\linewidth}{!}{
\begin{tabular}{c|l|c|ccc|ccc}
\shline
\rowcolor[HTML]{EFEFEF} 
\multicolumn{2}{l|}{\cellcolor[HTML]{EFEFEF}Method} &
  CS Ratio $\gamma$ &
  1\% &
  4\% 
 \\ \hline \hline

\multirow{6}{*}{\rotatebox[origin=c]{90}{\scriptsize Nature Image CS}} &
\multicolumn{2}{l|}{OPINE-Net$^+$ \cite{zhang2020optimization}}        & 20.15/0.5340 & 25.69/0.7920\\
& \multicolumn{2}{l|}{{\scriptsize QISTA-ImageNet \cite{lin2022qista}}}    & 21.34/0.5717 & 26.07/0.7869  \\
& \multicolumn{2}{l|}{CASNet \cite{chen2022cas}} & 21.97/0.6140 & 26.41/0.8153 \\ 
& \multicolumn{2}{l|}{OCTUF$^+$ \cite{song2023optimization}}         & 21.63/0.5903 & 26.21/0.8075  \\
& \multicolumn{2}{l|}{\cellcolor[HTML]{FFFFA5}\textbf{$\DC$-Net (Ours)}} &  \cellcolor[HTML]{FFFFA5}\msecondbest{22.18/0.6158}   &  \cellcolor[HTML]{FFFFA5}\msecondbest{26.84/0.8227}  \\
& \multicolumn{2}{l|}{\cellcolor[HTML]{FFFFA5}\textbf{$\DC$-Net$^+$ (Ours)}} &  \cellcolor[HTML]{FFFFA5}\textbf{\textcolor{red}{22.30/0.6251}}   &  \cellcolor[HTML]{FFFFA5}\textbf{\textcolor{red}{27.15/0.8322}} \\
\hline \hline
\multirow{4}{*}{\rotatebox[origin=c]{90}{\scriptsize CS-MRI}}& \multicolumn{2}{l|}{ISTA-Net$^+$~\cite{zhang2018ista}} & 21.03/0.4680 & 32.65/0.8733 \\ 
& \multicolumn{2}{l|}{MADUN~\cite{song2021memory}} & 27.59/0.7153 & 33.15/0.8847 \\ 
& \multicolumn{2}{l|}{MAPUN~\cite{song2023deep}} & \msecondbest{27.67/0.7270} & \msecondbest{33.26/0.8883} \\ 
& \multicolumn{2}{l|}{\cellcolor[HTML]{FFFFA5}\textbf{$\DC$-Net (Ours)}} &  \cellcolor[HTML]{FFFFA5}\textbf{\textcolor{red}{27.83/0.7473}}   &  \cellcolor[HTML]{FFFFA5}\textbf{\textcolor{red}{33.43/0.8911}} \\ \shline
\end{tabular}}
\end{minipage}}
\vspace{-5pt}
\end{table}

\begin{figure*}[t!]
    \centering
    \scriptsize
    \setlength{\tabcolsep}{0.6pt}
    \renewcommand{\arraystretch}{0.6}
    \begin{tabular}{ccccccccccc}
          Ground Truth & ReconNet & ISTA-Net$^+$ & ISTA-Net$^{++}$ &  CSNet$^+$ & SCSNet &  OPINE-Net$^+$ &  AMP-Net & COAST &  MADUN &  $\DC$-Net\\
        \includegraphics[width=0.09\textwidth]{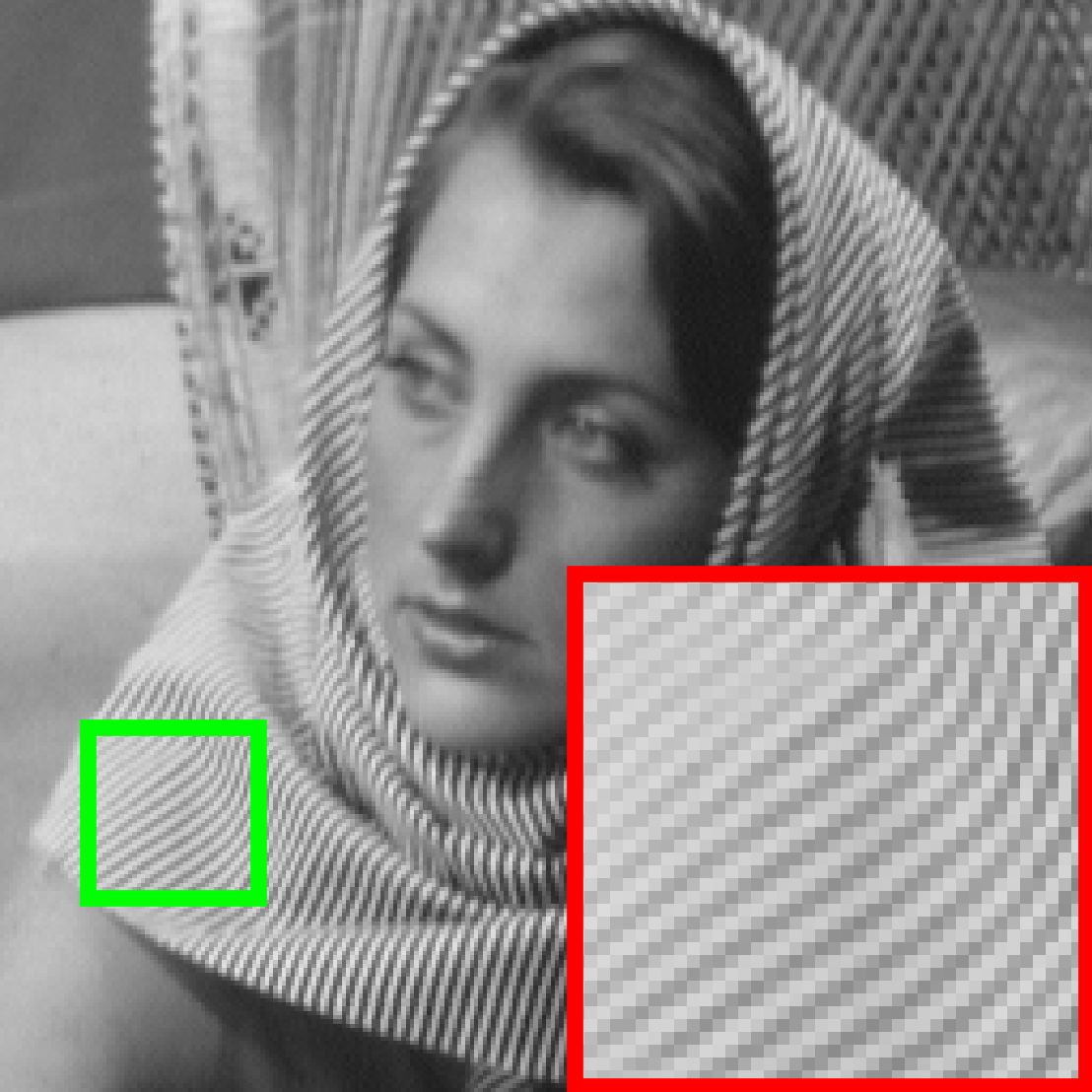} &
        \includegraphics[width=0.09\textwidth]{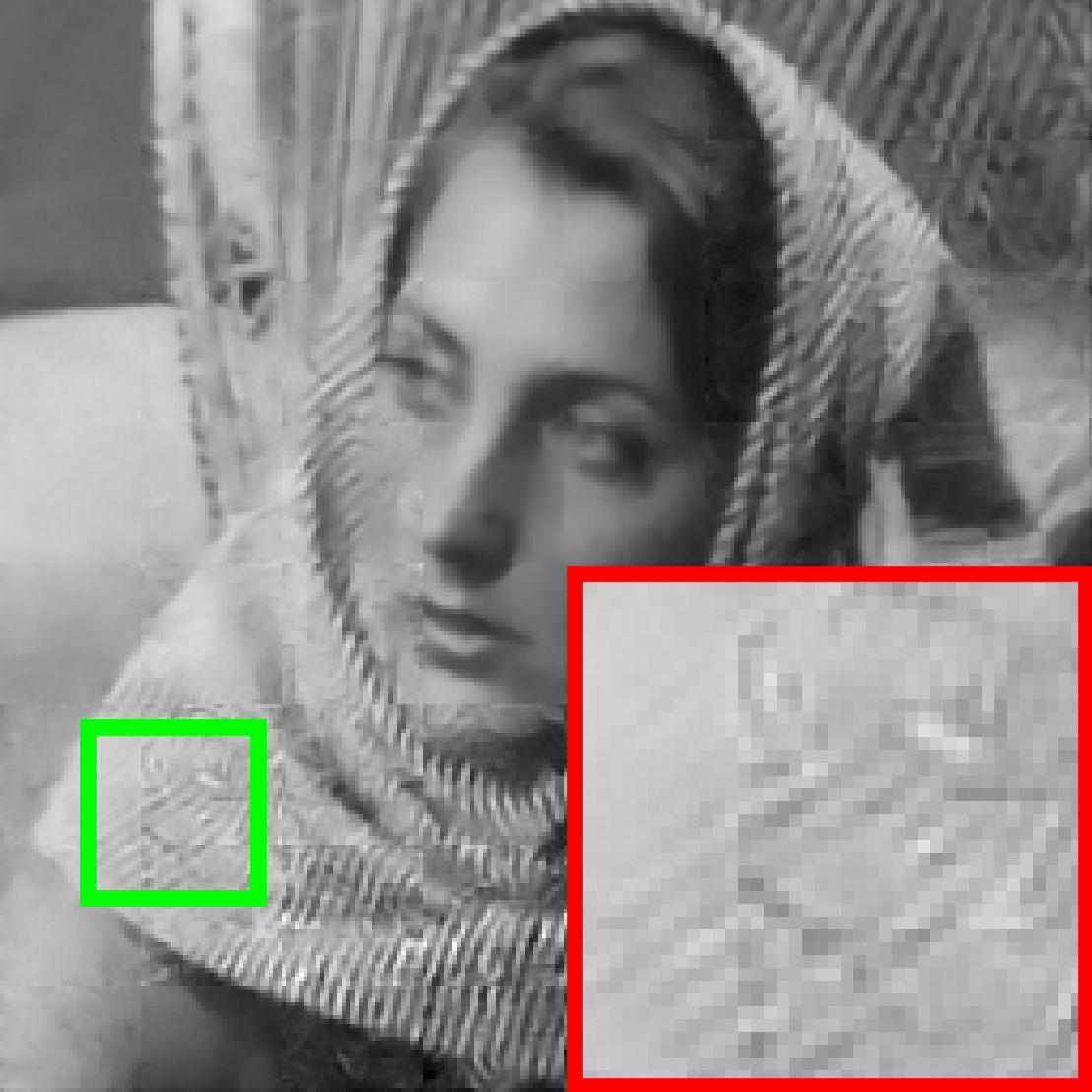} &
        
        \includegraphics[width=0.09\textwidth]{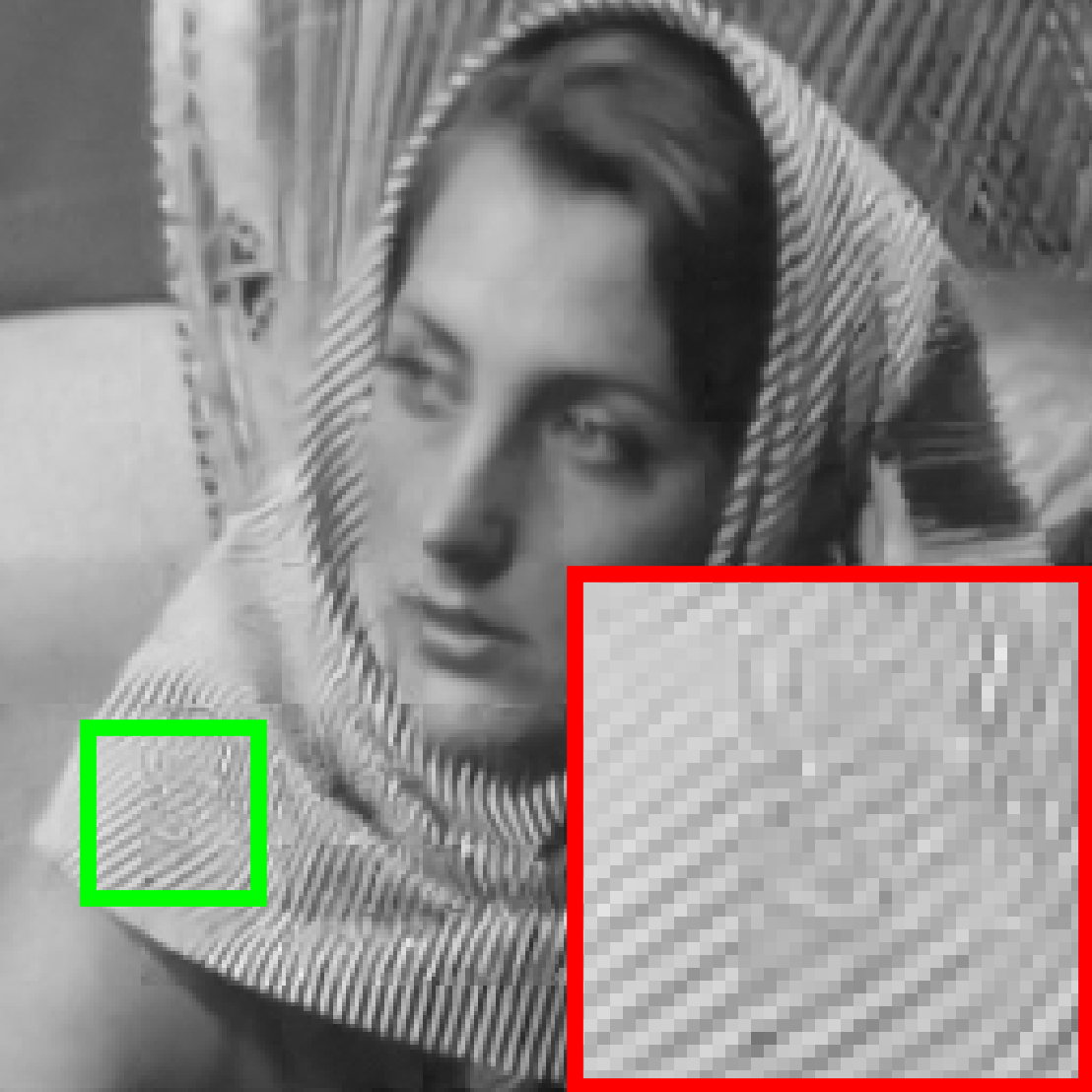} &

        \includegraphics[width=0.09\textwidth]{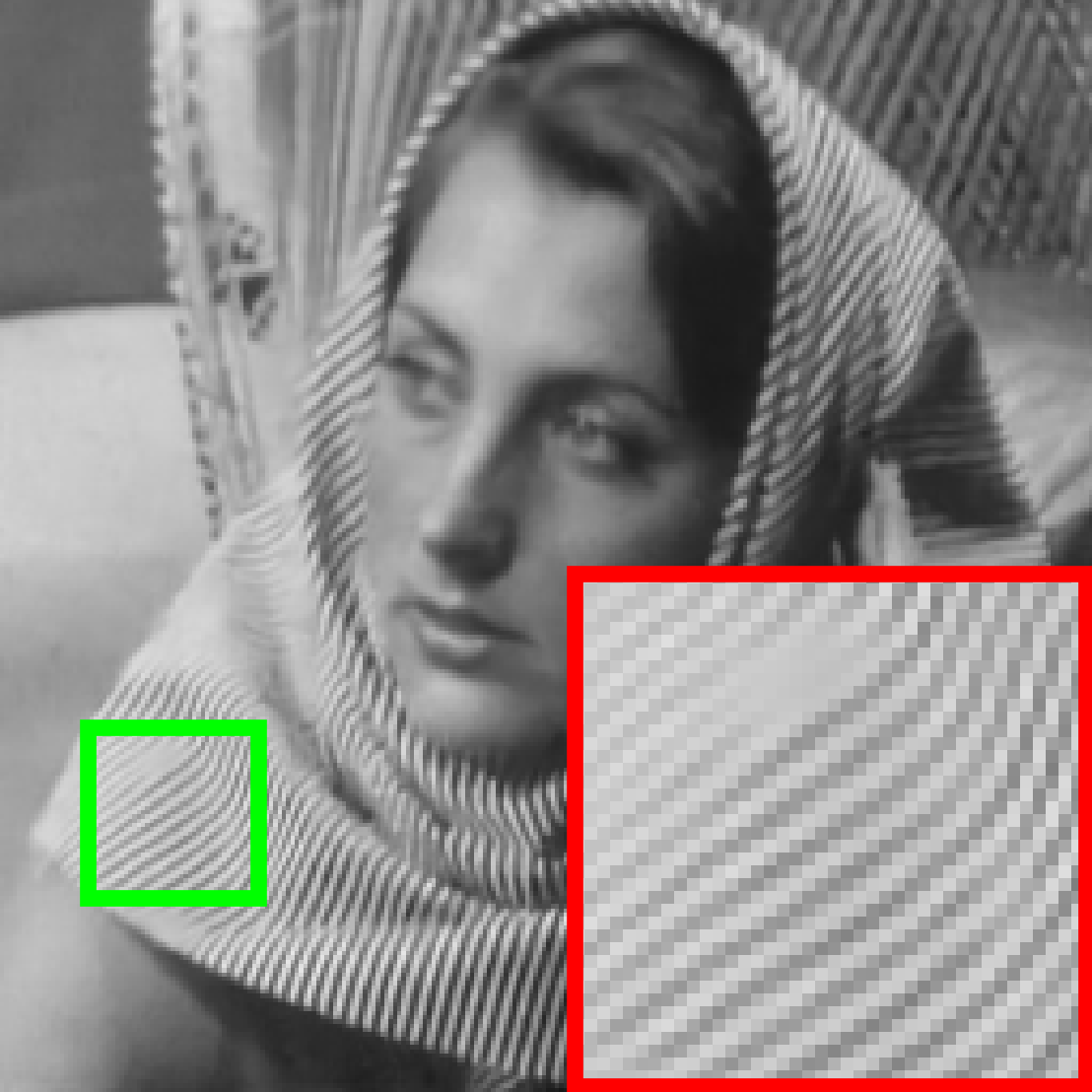} &
        \includegraphics[width=0.09\textwidth]{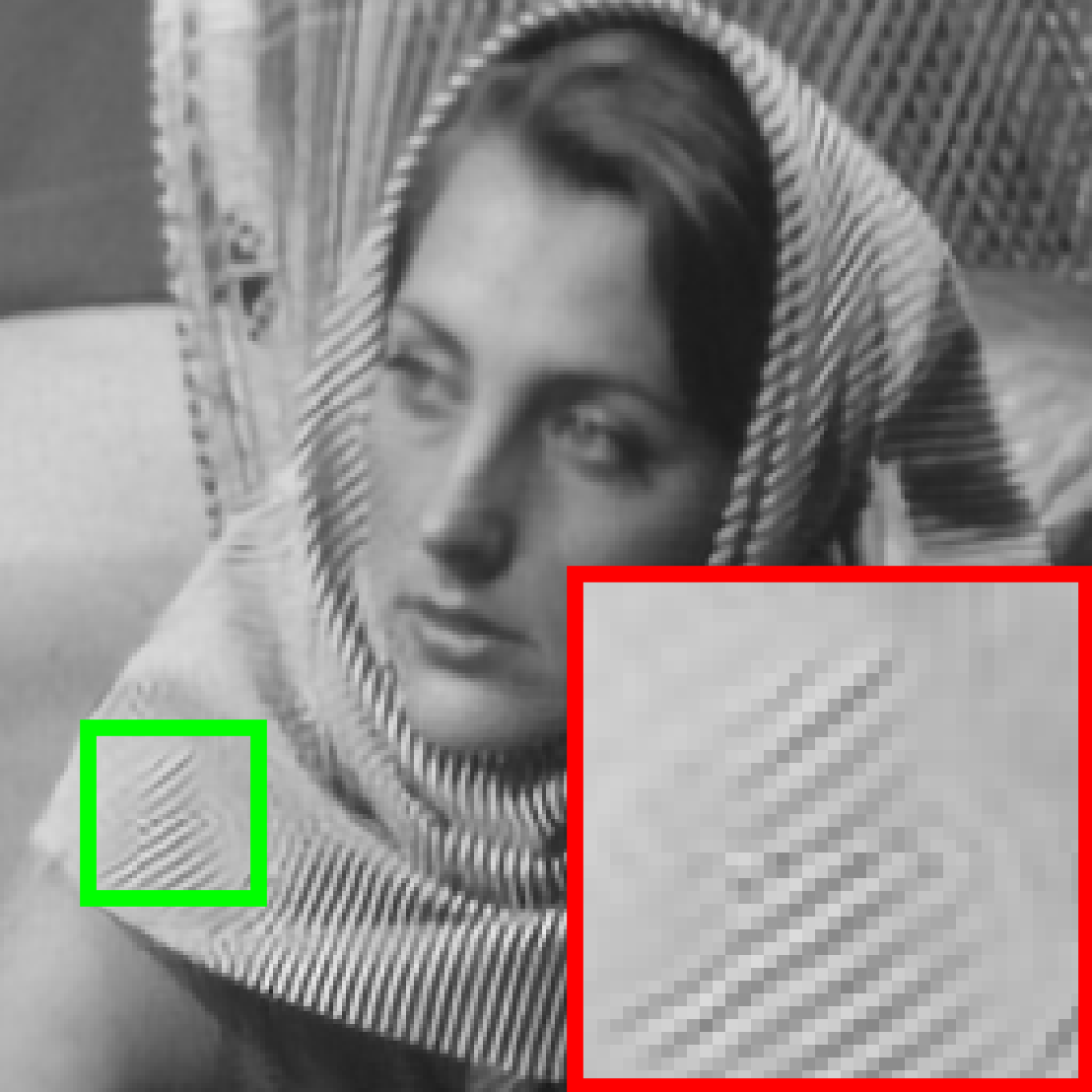} &
        \includegraphics[width=0.09\textwidth]{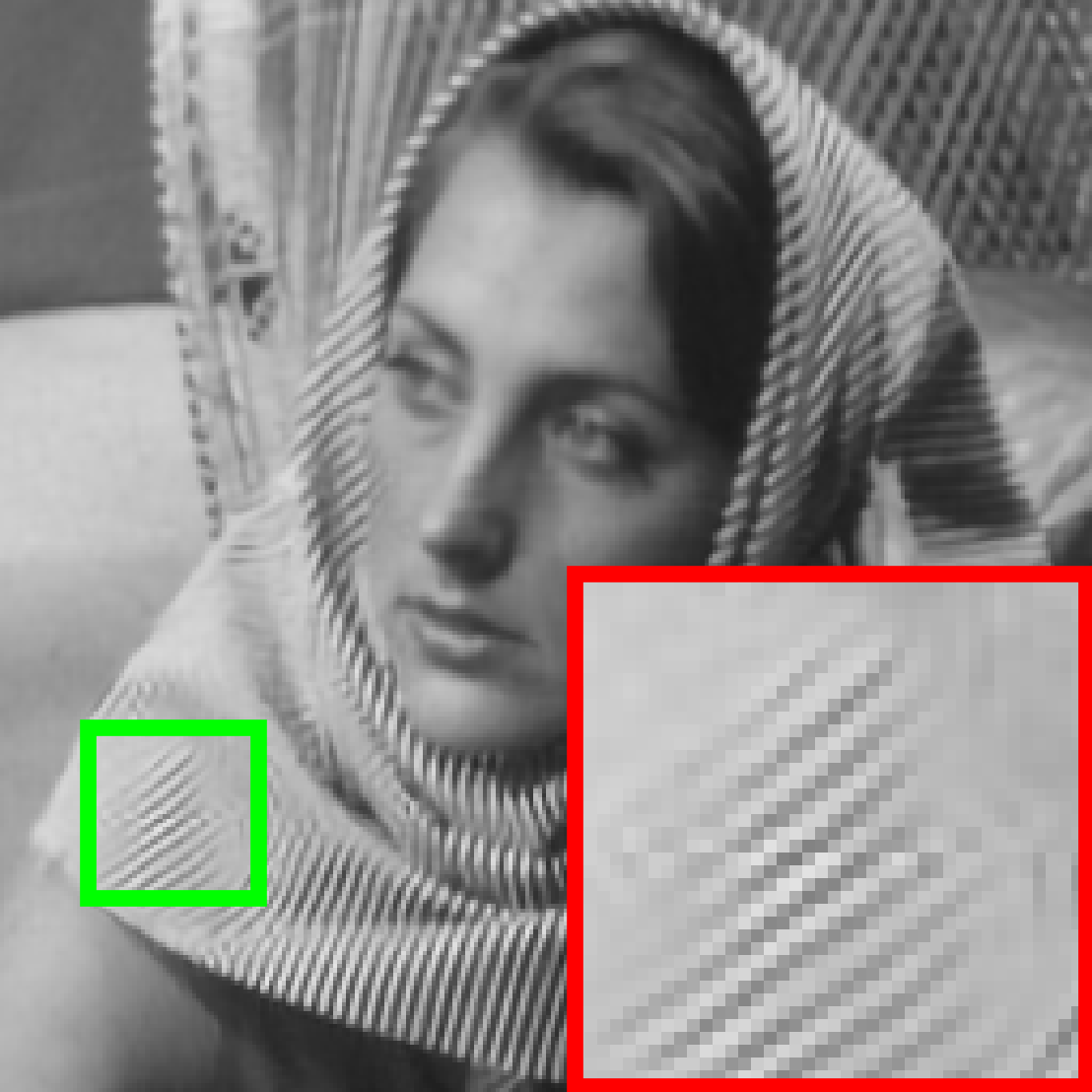} &
        \includegraphics[width=0.09\textwidth]{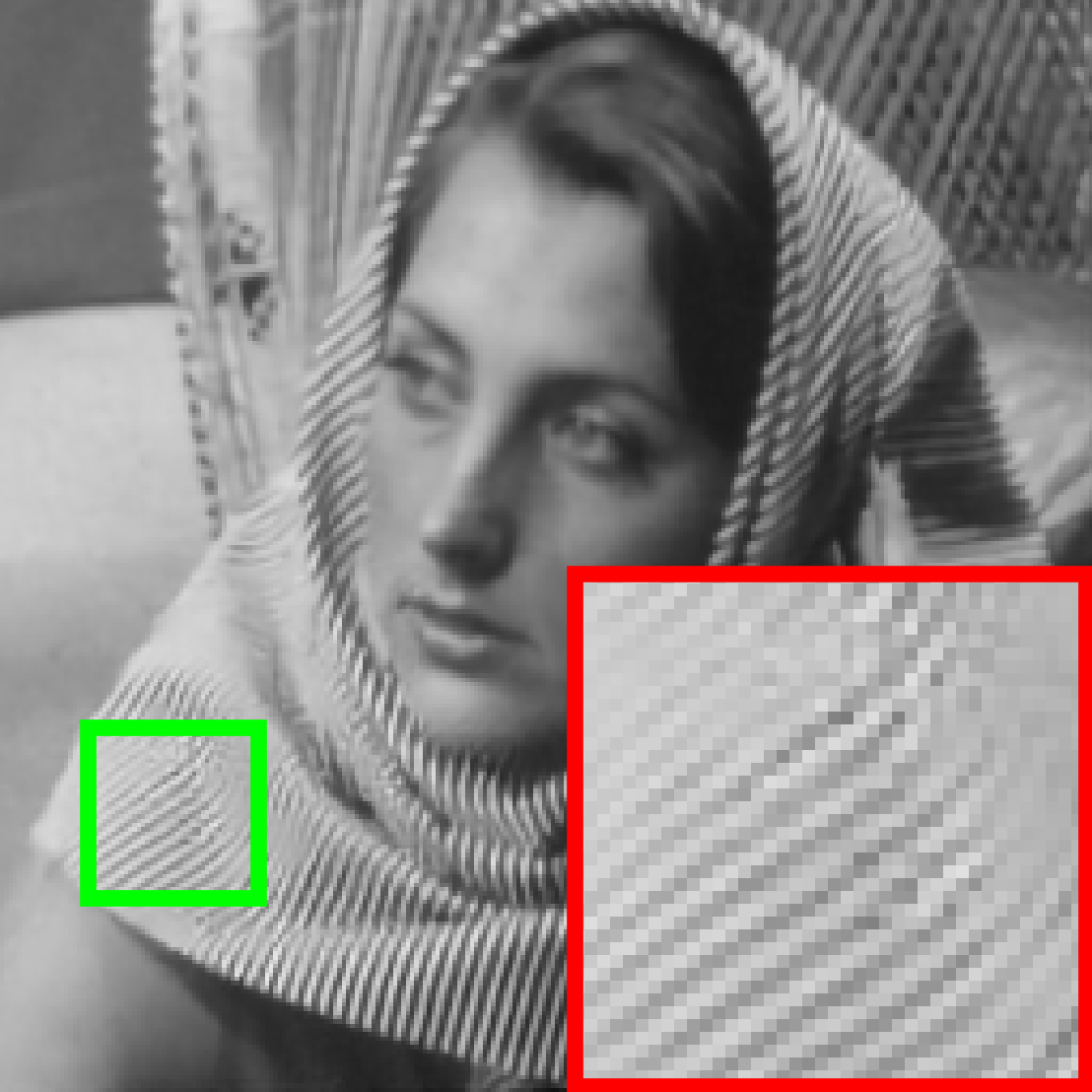} &
        \includegraphics[width=0.09\textwidth]{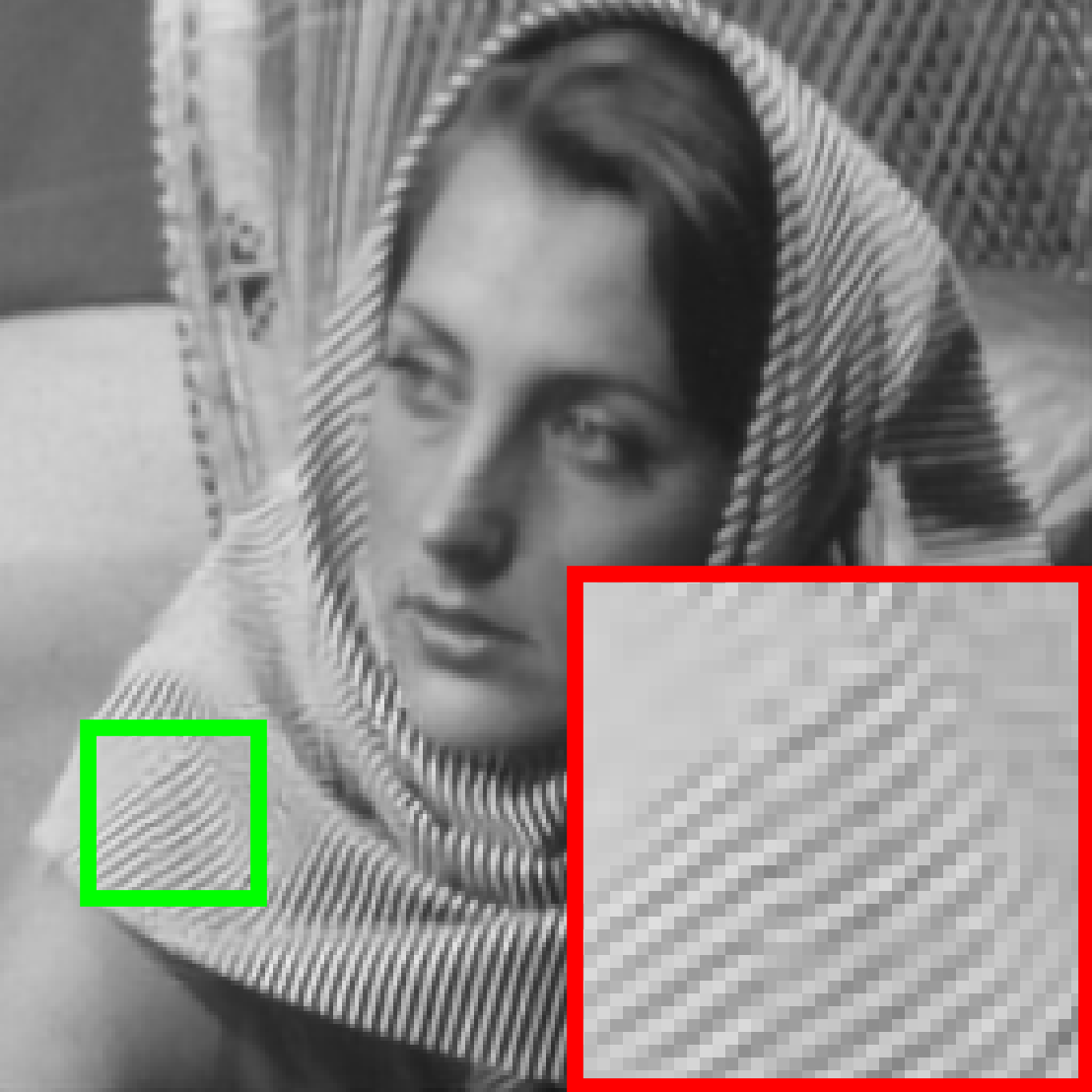} &
        \includegraphics[width=0.09\textwidth]{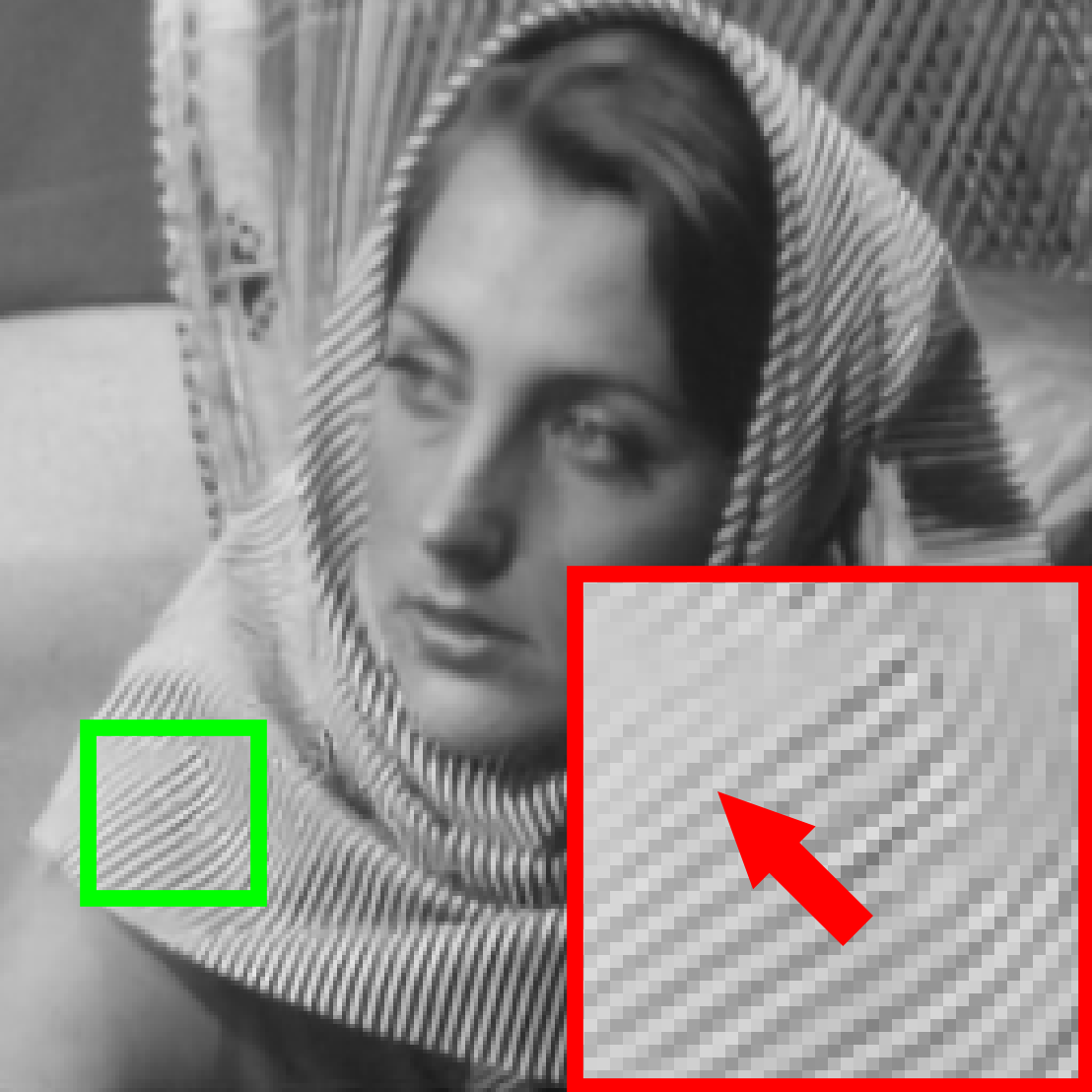} &
        
        \includegraphics[width=0.09\textwidth]{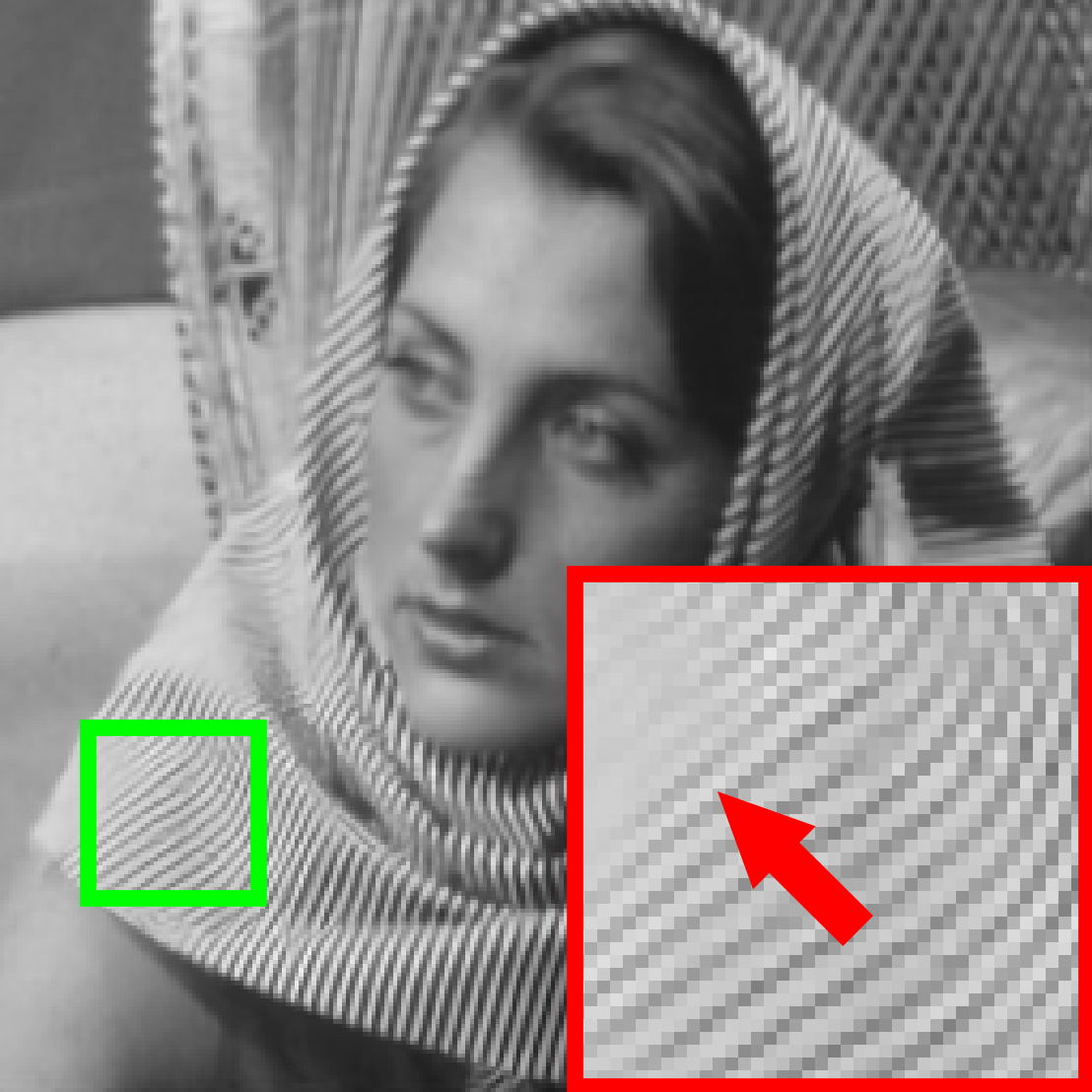} &
        \includegraphics[width=0.09\textwidth]{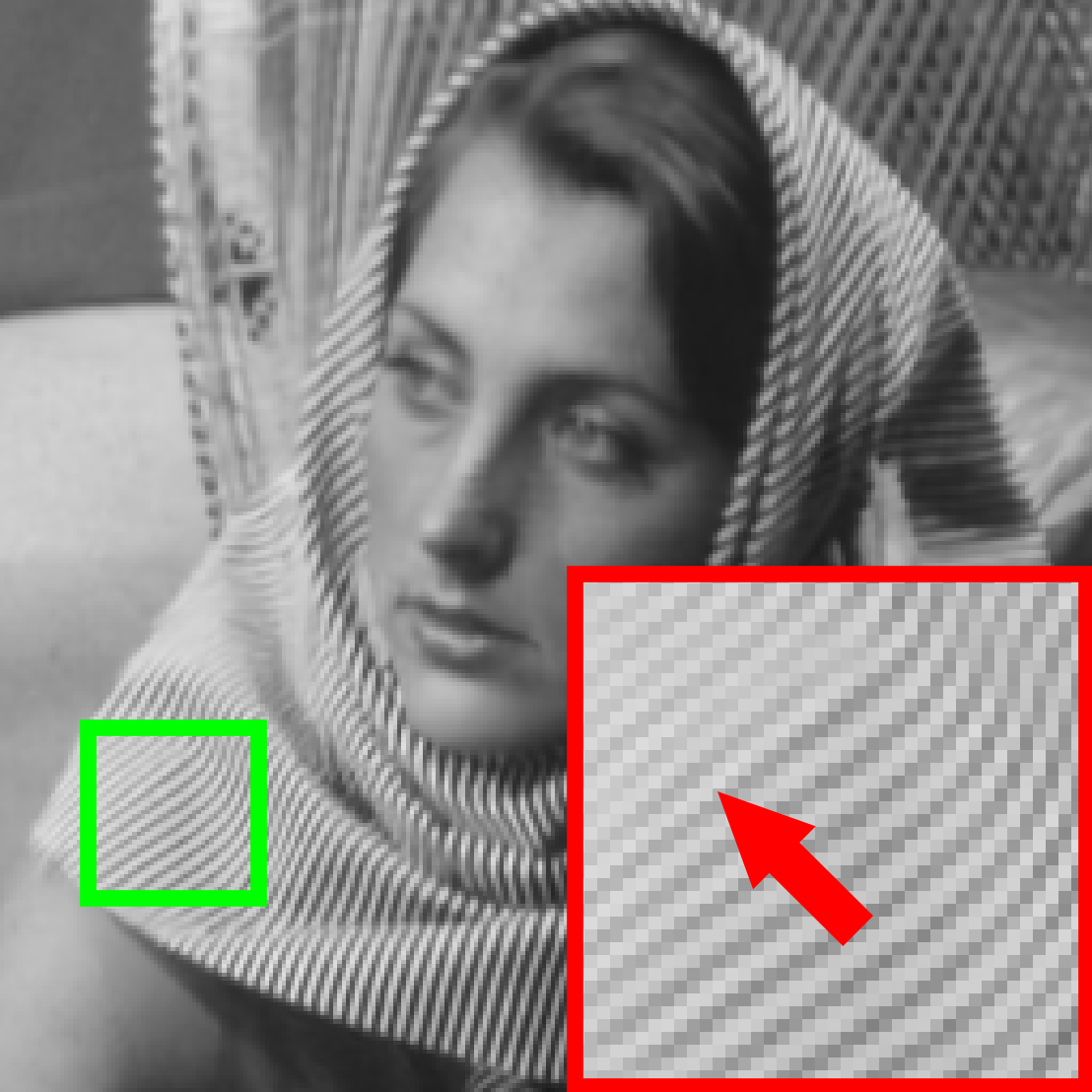}\\
        PSNR/SSIM & 26.64/0.8349 & 30.13/0.9240 & 33.67/0.9581 & 31.22/0.9255 & 31.43/0.9296 & 32.92/0.9561 & 33.53/0.9522 & 34.18/0.9637 & \underline{\textcolor{blue}{35.02}}/\underline{\textcolor{blue}{0.9721}} & \textbf{\textcolor{red}{37.04}}/\textbf{\textcolor{red}{0.9794}}\\
        \includegraphics[width=0.09\textwidth]{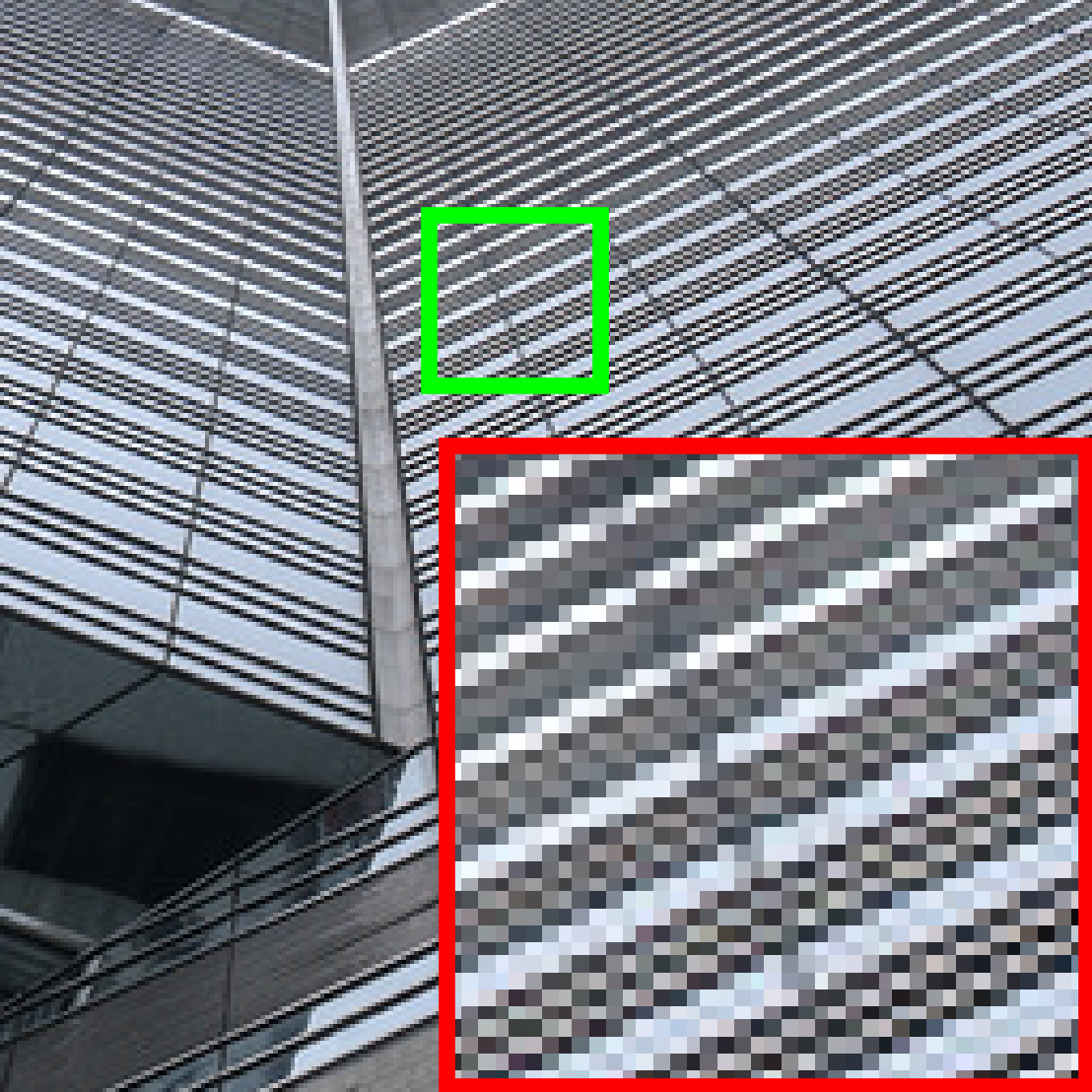} &
        \includegraphics[width=0.09\textwidth]{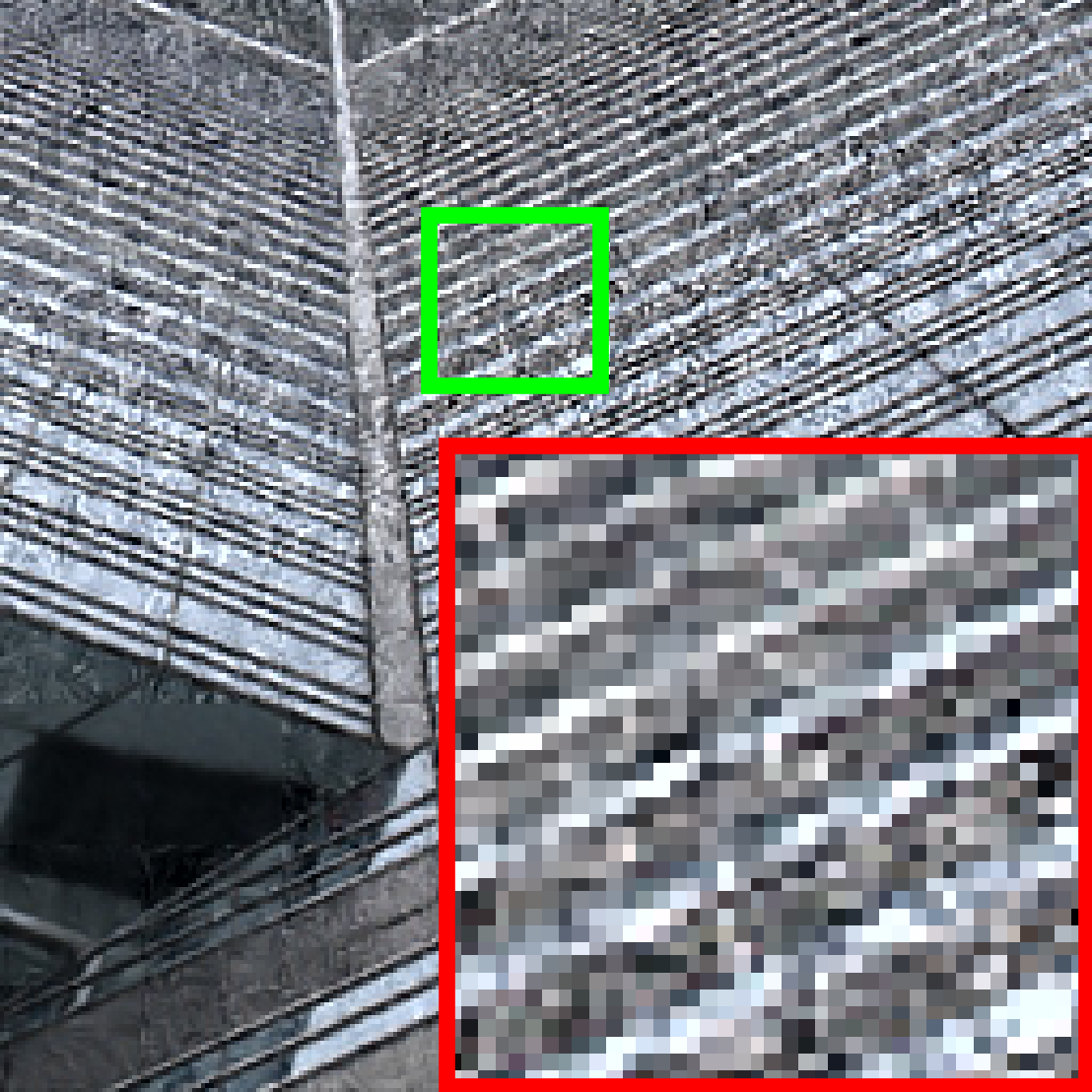} &
        
        \includegraphics[width=0.09\textwidth]{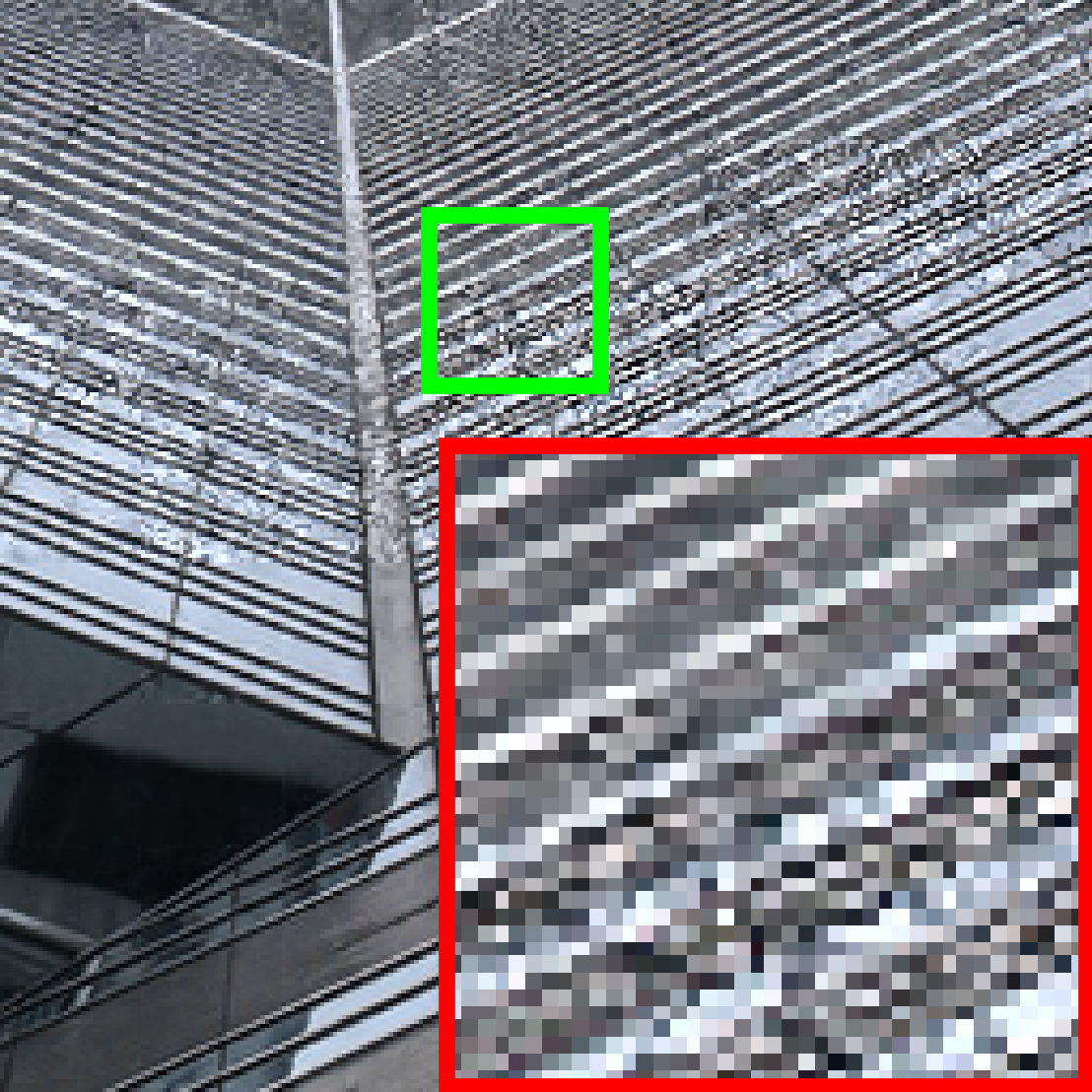} &

        \includegraphics[width=0.09\textwidth]{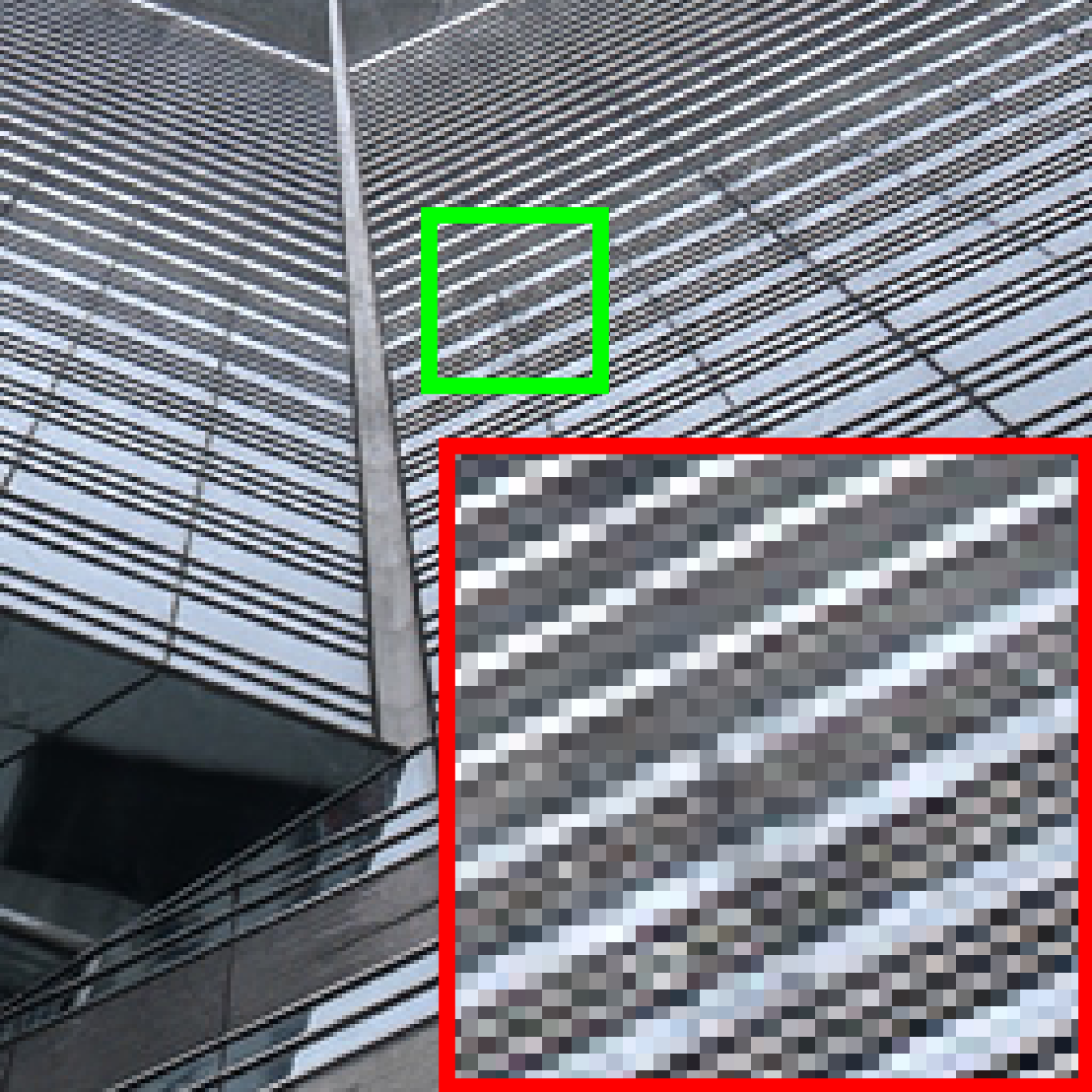} &
        \includegraphics[width=0.09\textwidth]{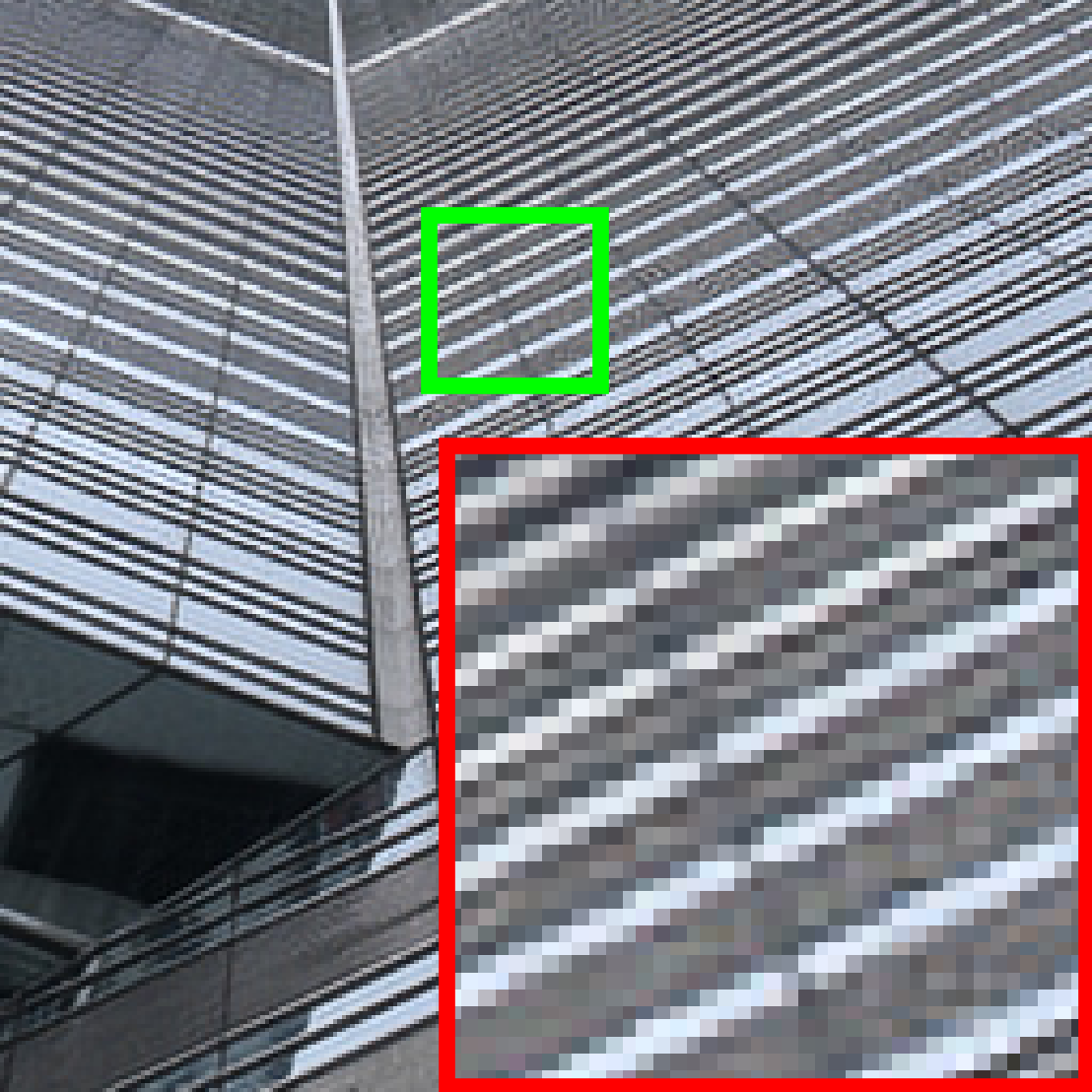} &
        \includegraphics[width=0.09\textwidth]{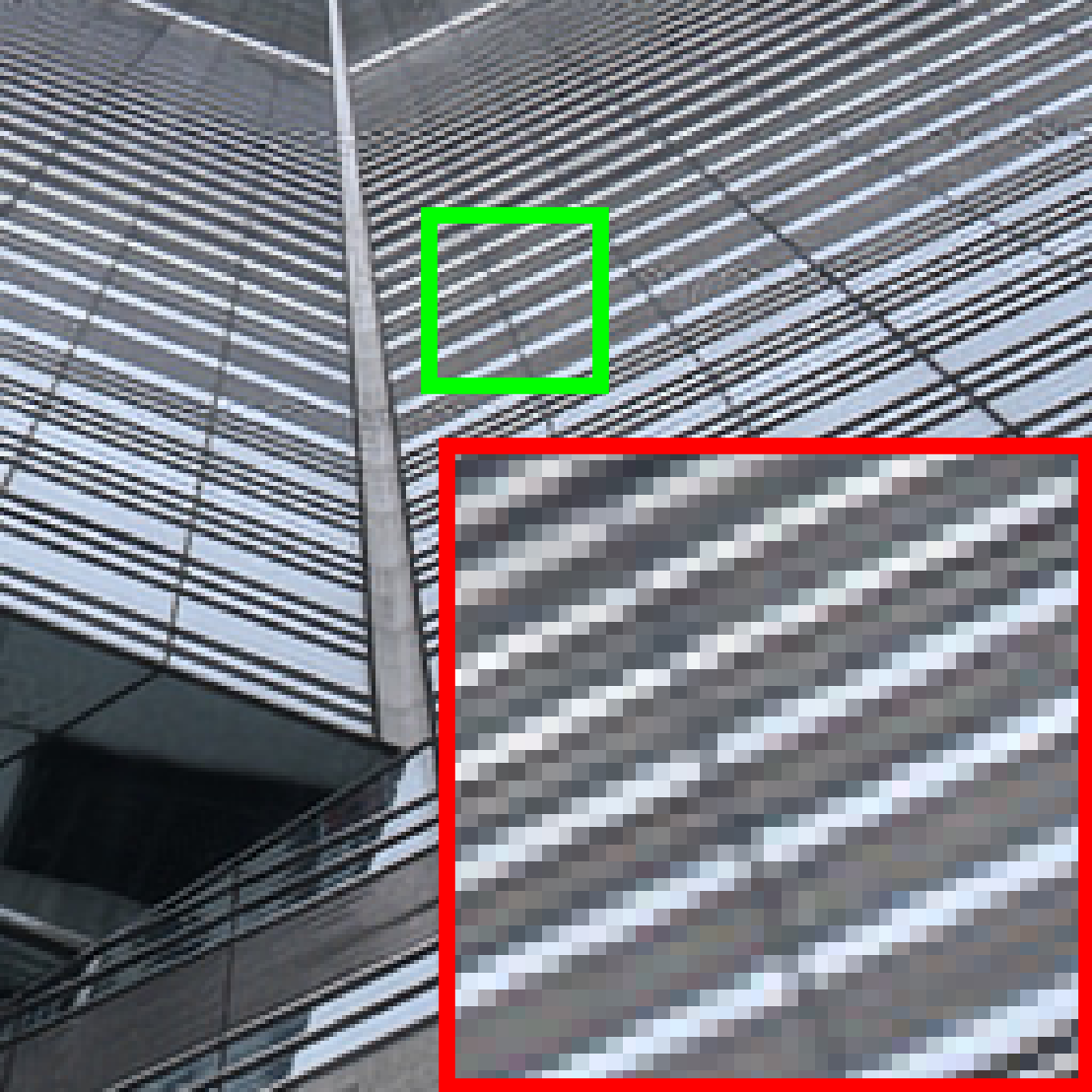} &
        \includegraphics[width=0.09\textwidth]{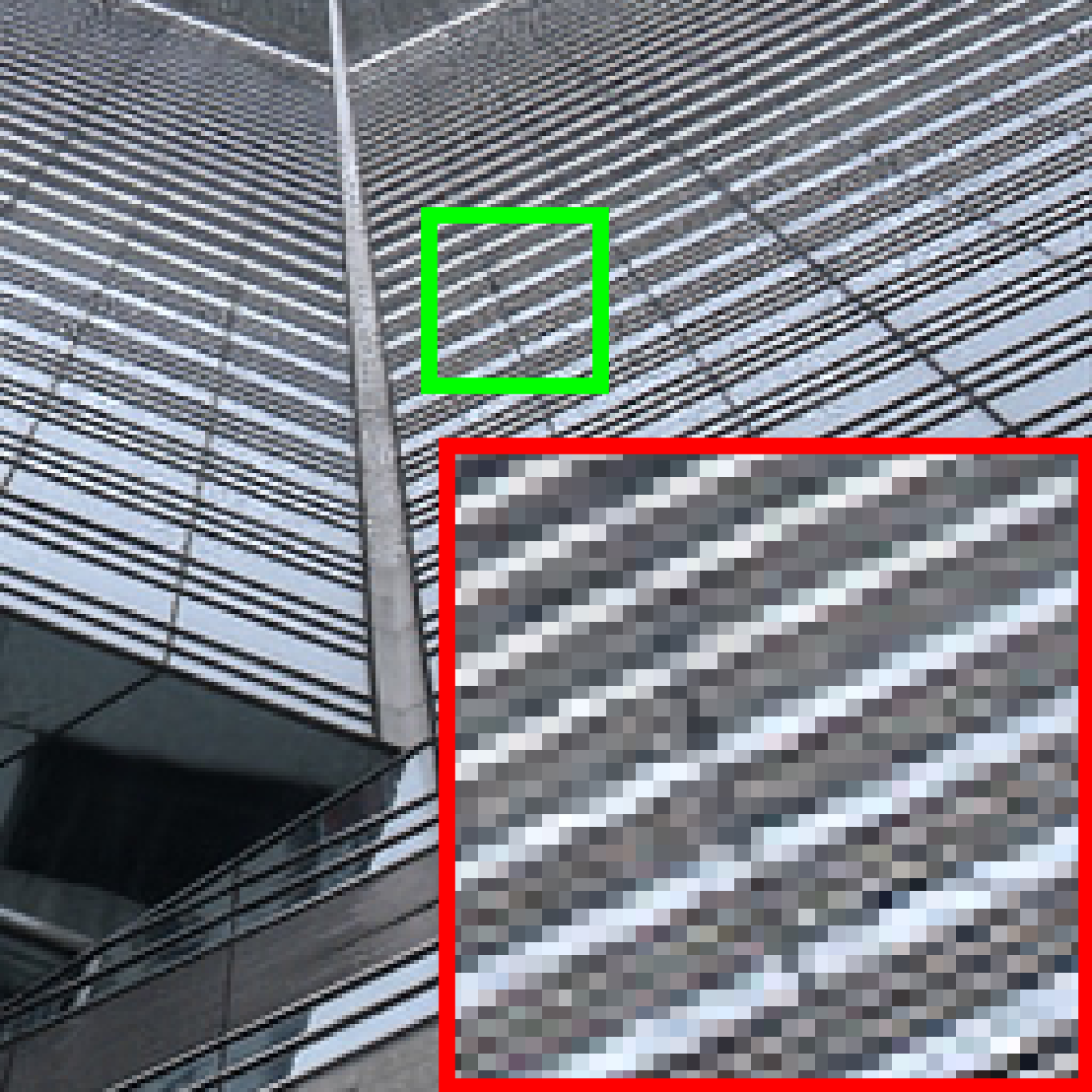} &
        \includegraphics[width=0.09\textwidth]{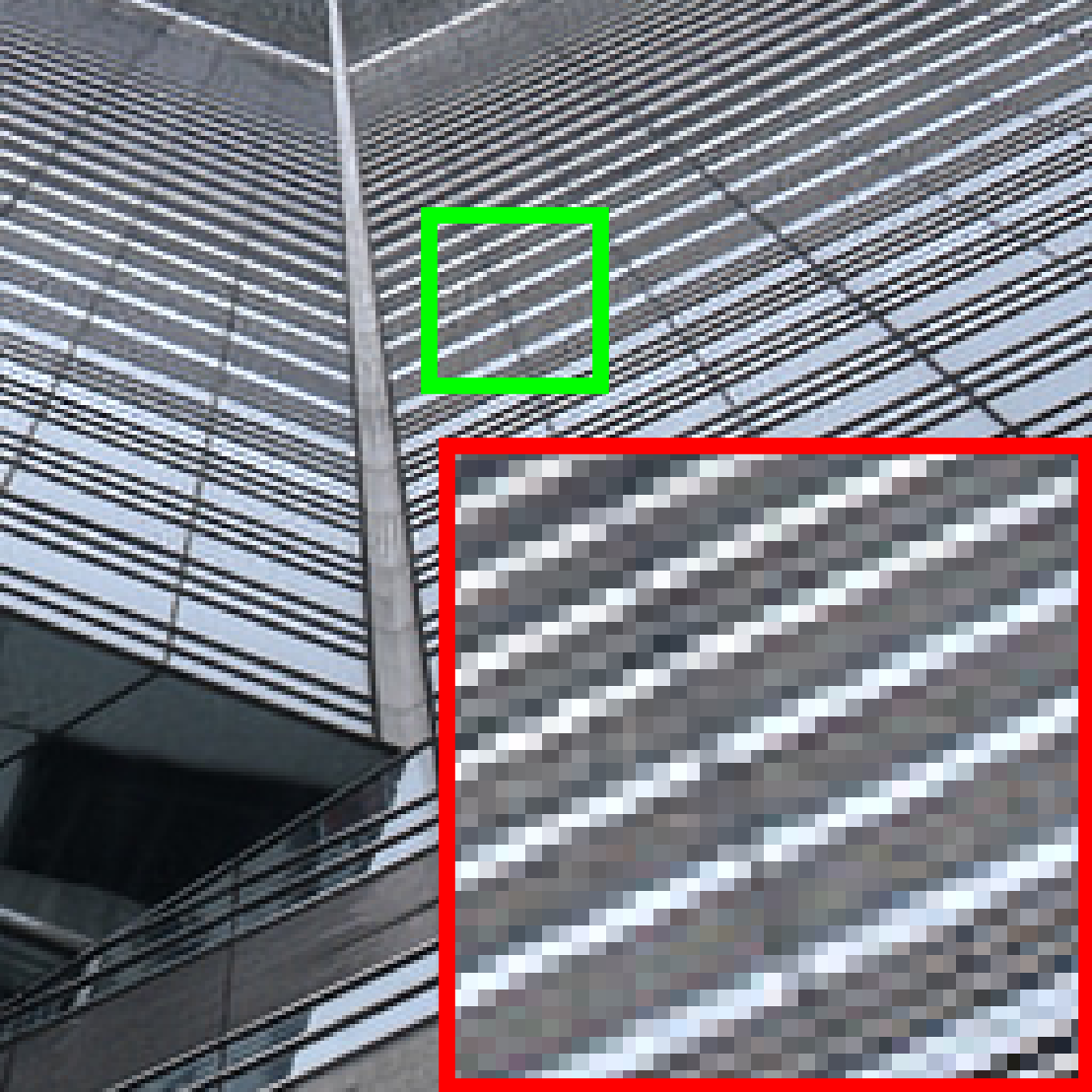} &
        \includegraphics[width=0.09\textwidth]{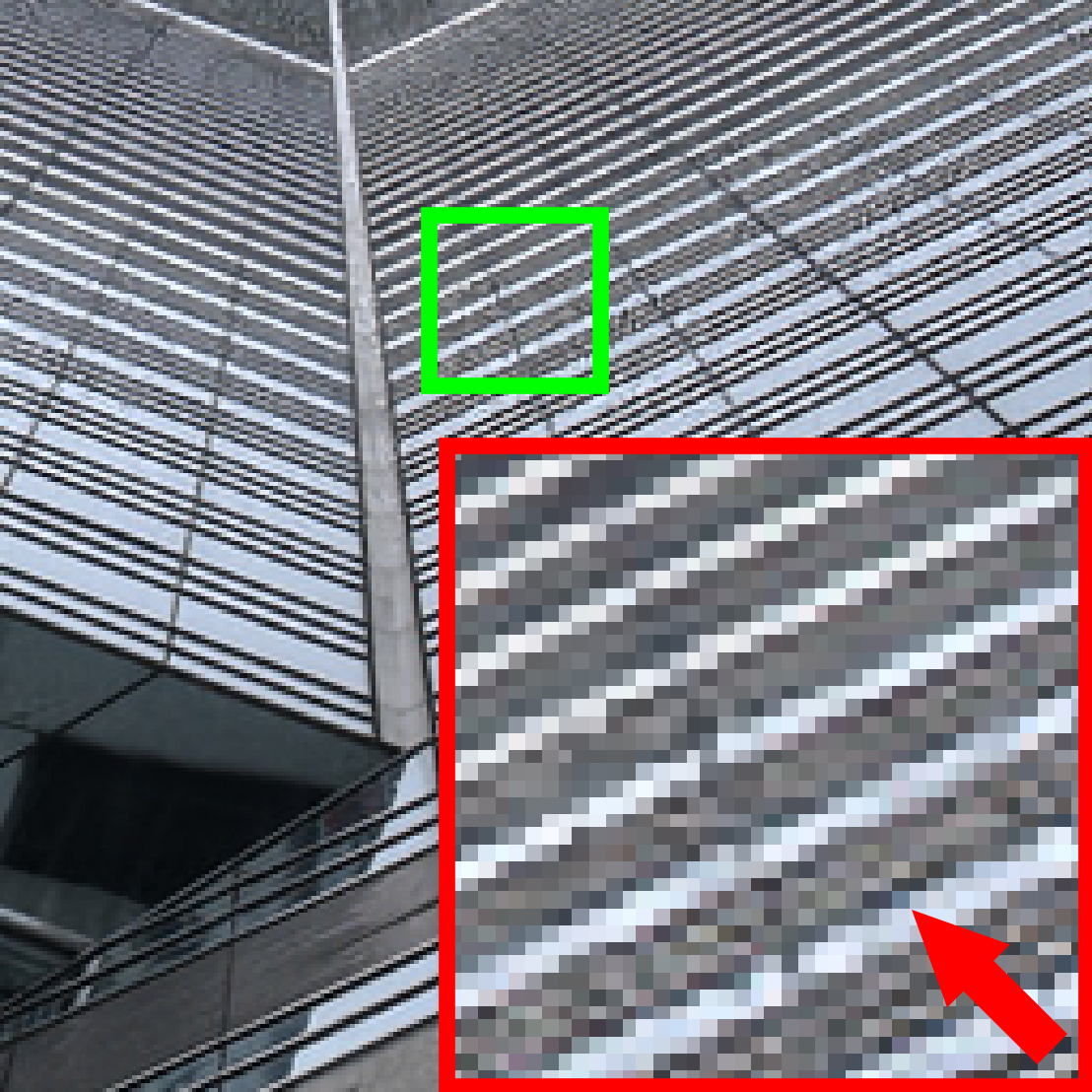} &
        
        \includegraphics[width=0.09\textwidth]{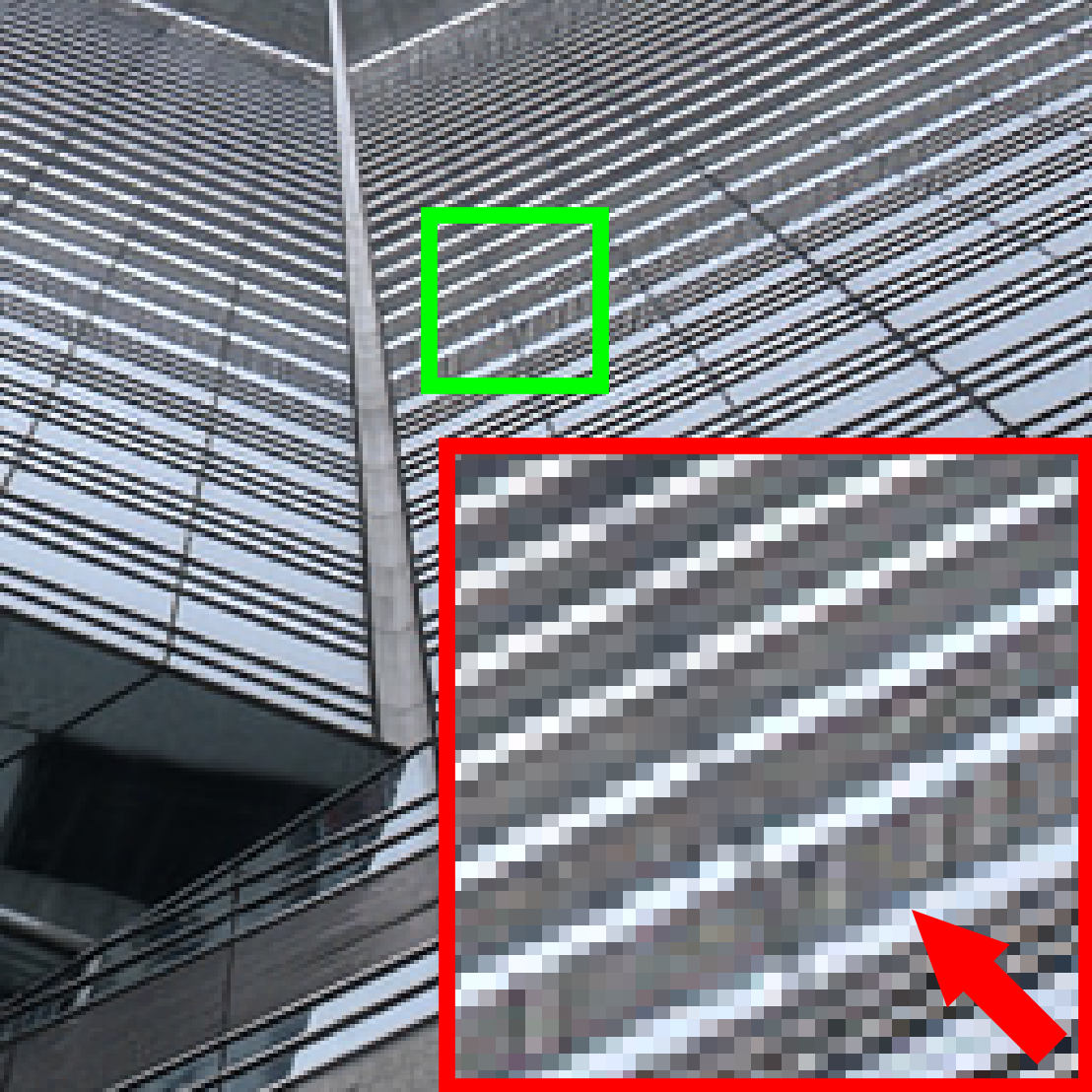} &
        \includegraphics[width=0.09\textwidth]{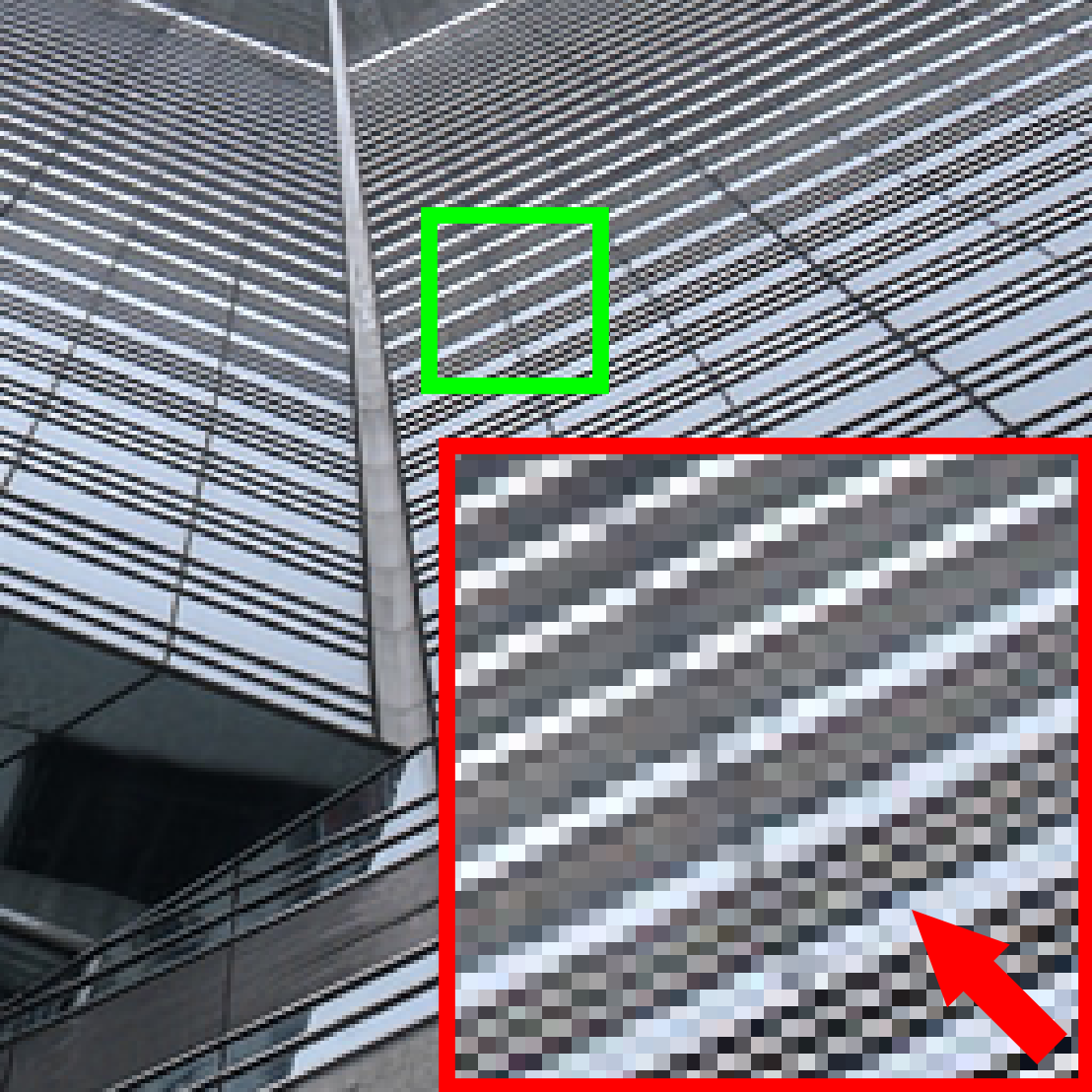}\\

        PSNR/SSIM & 18.23/0.7700 & 21.13/0.9060 & \underline{\textcolor{blue}{29.68}}/\underline{\textcolor{blue}{0.9756}} & 21.65/0.8775 & 22.68/0.8986 & 25.71/0.9554 & 24.93/0.9381 & 25.92/0.9552 & 27.03/0.9657 & \textbf{\textcolor{red}{31.35}}/\textbf{\textcolor{red}{0.9824}}\\
        
    \end{tabular}
    \vspace{-4pt}
    \caption{Visual comparisons on two images from Set11~\cite{kulkarni2016reconnet} (``Barbara'', \textcolor{blue}{top}) and Urban100~\cite{huang2015single} (``059'', \textcolor{blue}{bottom}) datasets, with $\gamma=30\%$ and $50\%$, respectively.}
    \label{fig:images_comparison_all}
\vspace{-12pt}
\end{figure*}

\vspace{-10pt}
\subsection{Comparison with State-of-the-Art Methods}

We conduct a comprehensive comparison \mrev{among our $\DC$-Net, $\DC$-Net$^+$}, and \mrev{seventeen} state-of-the-art end-to-end deep CS methods. \mrev{These methods include ReconNet~\cite{kulkarni2016reconnet}, ISTA-Net$^+$~\cite{zhang2018ista}, DPA-Net~\cite{sun2020dual}, MAC-Net~\cite{chen2020learning}, ISTA-Net$^{++}$~\cite{you2021ista}, CSNet$^+$~\cite{shi2019image}, SCSNet~\cite{shi2019scalable}, OPINE-Net$^+$~\cite{zhang2020optimization}, GPX-ADMM-Net~\cite{hu2021gpx}, AMP-Net~\cite{zhang2020amp}, COAST~\cite{you2021coast}, MADUN~\cite{song2021memory}, TransCS~\cite{shen2022transcs}, QISTA-ImageNet~\cite{lin2022qista}, CASNet~\cite{chen2022cas}, MAPUN~\cite{song2023deep}, and OCTUF$^+$~\cite{song2023optimization}. The former five methods employ the fixed random Gaussian matrix (FRGM), while the latter twelve adopt the data-driven adaptively learned matrix (DALM).} We evaluate the average PSNR of these approaches on Set11~\cite{kulkarni2016reconnet} and Urban100~\cite{huang2015single} across different CS ratios, stage number, inference time, GFLOPs, and parameter number. The results are summarized in Tab.~\ref{tab:compare_sota_psnr}. \mrev{Our more corresponding SSIM results are provided in Tab.~IX in the \textbf{\textit{\textcolor{blue}{supplemental material}}}}. To avoid memory overflow in some large-capacity networks, we perform $256\times 256$ center-cropping on all images from Urban100. When comparing with existing methods, we reference their results directly from the original sources whenever possible. If necessary, we carefully adjust hyper-parameters to obtain comparable results. In cases where replicable algorithms are not available, we leave the corresponding fields empty in tables. \mrev{It is worth noting that our $\DC$-Nets are scalable, which means one set of parameters can handle all ratios $\gamma\in[0,1]$. We further investigate the reconstruction performance under extremely low ratios $\gamma \in \{1\%, 4\%\}$ in Tab.~\ref{tab:low_ratios}. By leveraging our dual-domain high-throughput unfolding principle, $\DC$-Net and $\DC$-Net$^+$ exhibit a remarkable advantage over the existing state-of-the-art CS approaches across all specified CS ratios and two distinct types of sampling matrices, especially on low ratios.} Moreover, our method maintains efficient implementation and manageable complexity. Fig.~\ref{fig:images_comparison_all} visually compares the results of different methods on challenging images from Set11 and Urban100. It demonstrates that the reconstructions of $\DC$-Net achieve superior PSNR and structural similarity index measure (SSIM) metrics. Additionally, $\DC$-Net recovers more reliable textures, with better delineation of lines and stripes and less blurring compared to the results from other competing methods.

\begin{figure}[t!]
\centering
\includegraphics[width=\linewidth]{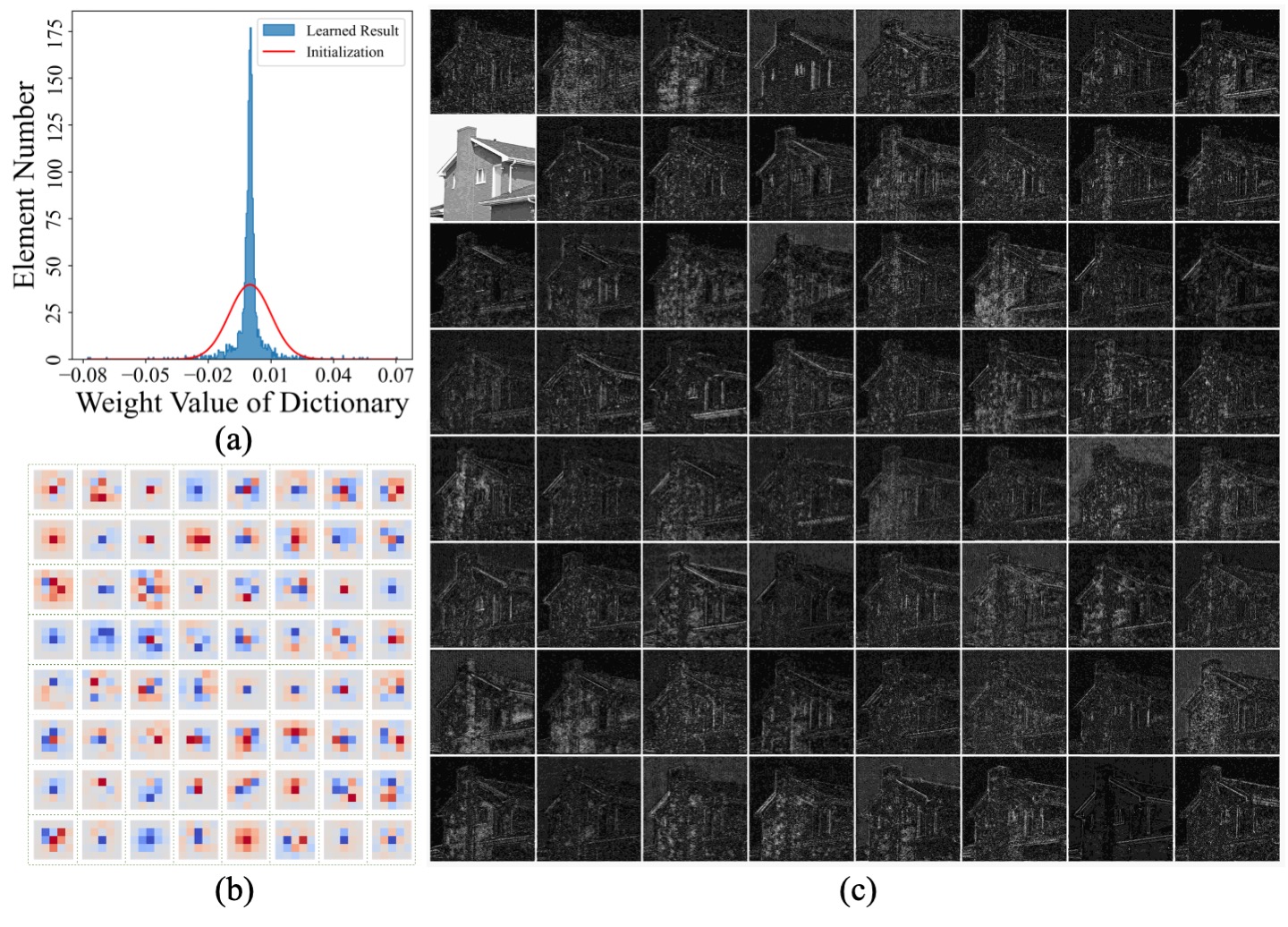}
\vspace{-18pt}
\caption{\mrev{Visualizations for analyzing the learned dictionary $\D \in \mathbb{R}^{64\times 5 \times 5}$ and feature maps $\balpha \in \mathbb{R}^{64\times 256 \times 256}$ with $\gamma=50\%$, including \textcolor{blue}{(a)} the distribution of initialized and final learned weight value of $\D$, \textcolor{blue}{(b)} visualizations of filters of $\D$, and \textcolor{blue}{(c)} all feature maps $\balpha_i$ of an image named ``house'' from Set11~\cite{kulkarni2016reconnet}.}}
\label{fig:DX}
\end{figure}

\begin{figure}[t!]
    \centering
    \scriptsize
    \setlength{\tabcolsep}{0.4pt}
    \renewcommand{\arraystretch}{0.6}
    \vspace{-5pt}
    \resizebox{1.0\linewidth}{!}{
    \begin{tabular}{cc@{\hspace{3pt}}cc@{\hspace{3pt}}cc@{\hspace{4pt}}cc}

       \includegraphics[width=0.11\linewidth]{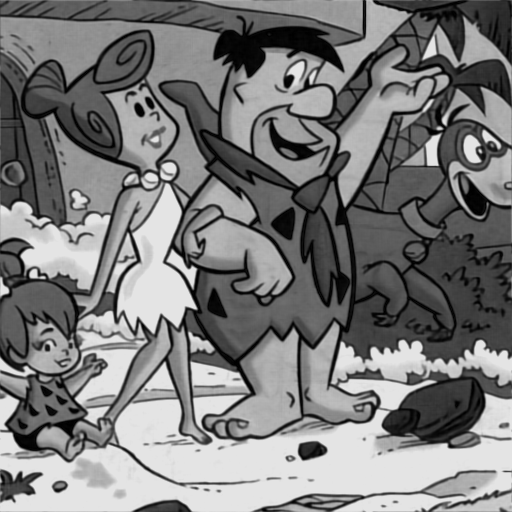} &
       \includegraphics[width=0.11\linewidth]{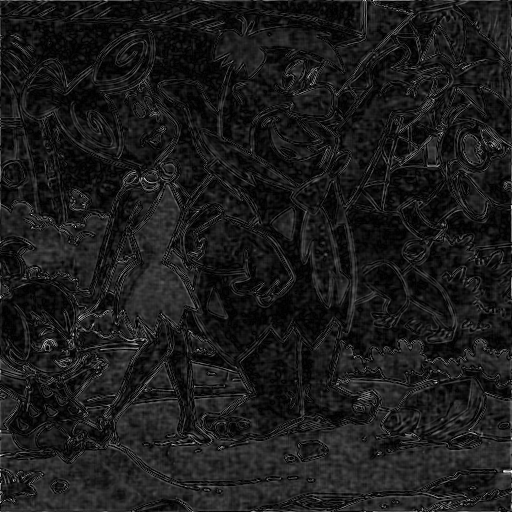} &
       \includegraphics[width=0.11\linewidth]{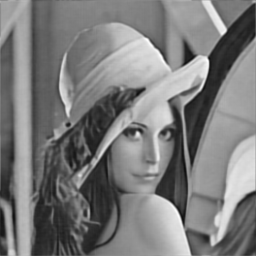} &
       \includegraphics[width=0.11\linewidth]{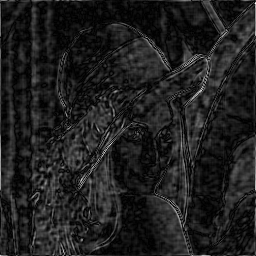} &
       \includegraphics[width=0.11\linewidth]{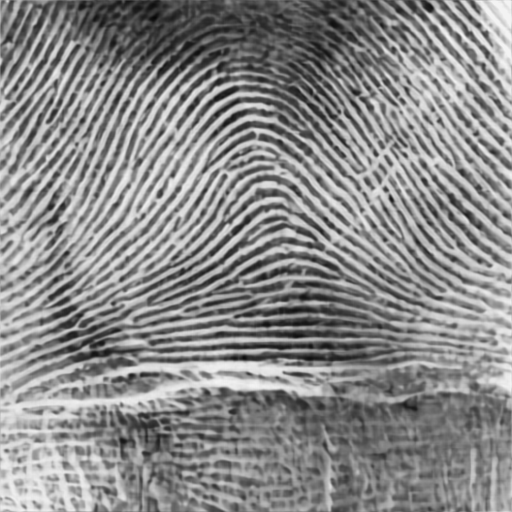} &
       \includegraphics[width=0.11\linewidth]{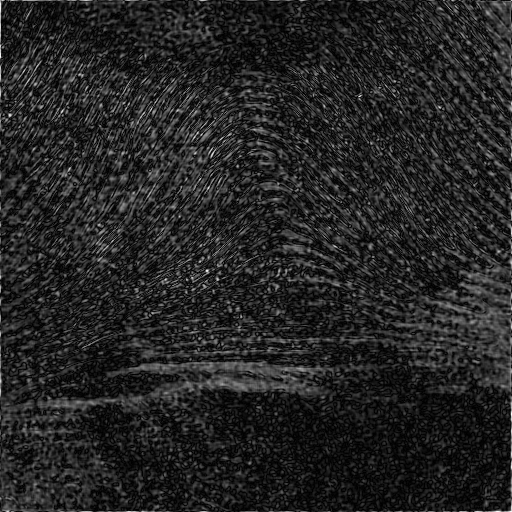} &
       \includegraphics[width=0.11\linewidth]{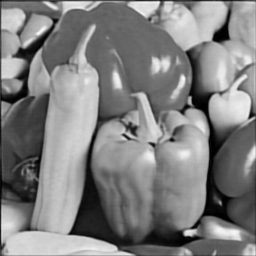} &
       \includegraphics[width=0.11\linewidth]{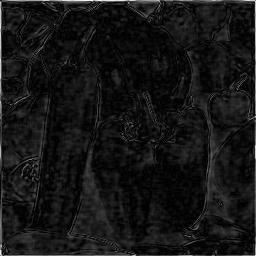} \\
       \multicolumn{2}{c}{flinstones}  &  \multicolumn{2}{c}{lena} &
       \multicolumn{2}{c}{fingerprint} &
       \multicolumn{2}{c}{peppers}\\
        
    \end{tabular}
}
    \caption{Visualizations of low-frequency information $\bd_2\ast\balpha_2$ \textcolor{blue}{(left)} and complementary high-frequency components $\sum_{i\ne2}\left(\bd_i\ast \balpha_i\right)$ \textcolor{blue}{(right)}, with applying $\DC$-Net to four images in Set11~\cite{kulkarni2016reconnet}.}
    \label{fig:lf_hf_0.1}
    \vspace{-5pt}
\end{figure}

\vspace{-10pt}
\subsection{Analysis of the Learned Dictionary $\D$ and Feature Map $\balpha$}

To obtain insights into the representation capability of ID- and CCD-based $\DC$-Nets, we conduct an analysis to explore what has been learned by our network. Firstly, we visualize the learned convolutional dictionary $\D$ and examine its element distributions. \mrev{As illustrated in Fig.~\ref{fig:DX} (a), the weight initialization of $\D$ employs a Gaussian distribution by default. However, over the training process, a majority of the elements in $\D$ progressively converge towards zero, signifying a certain degree of sparsity of the learned $\D$. Additionally, as depicted in Fig~\ref{fig:DX} (b), we observe that the filters in the learned dictionary display their diverse and anisotropic spatial distributions, enabling them to capture gradients in different directions based on the learned patterns.} We then visualize the final refined convolutional coefficients $\balpha^{(T)}$, as illustrated in Fig.~\ref{fig:DX}~(c). In contrast to traditional convolutional sparse coding approaches~\cite{fu2019jpeg, gao2022multi}, which explicitly enforce sparsity priors, our feature maps are implicitly regularized and not excessively sparse. Our other finding is that we consistently observe the presence of one feature channel dedicated to containing the low-frequency information across the unfolded stage-by-stage inferences, exemplified by the 8-th channel in Fig.~\ref{fig:DX}~(c).

\begin{figure}[!t]
    \centering
    \setlength{\tabcolsep}{0.6pt}
    \renewcommand{\arraystretch}{0.6}
    \includegraphics[width=\linewidth]{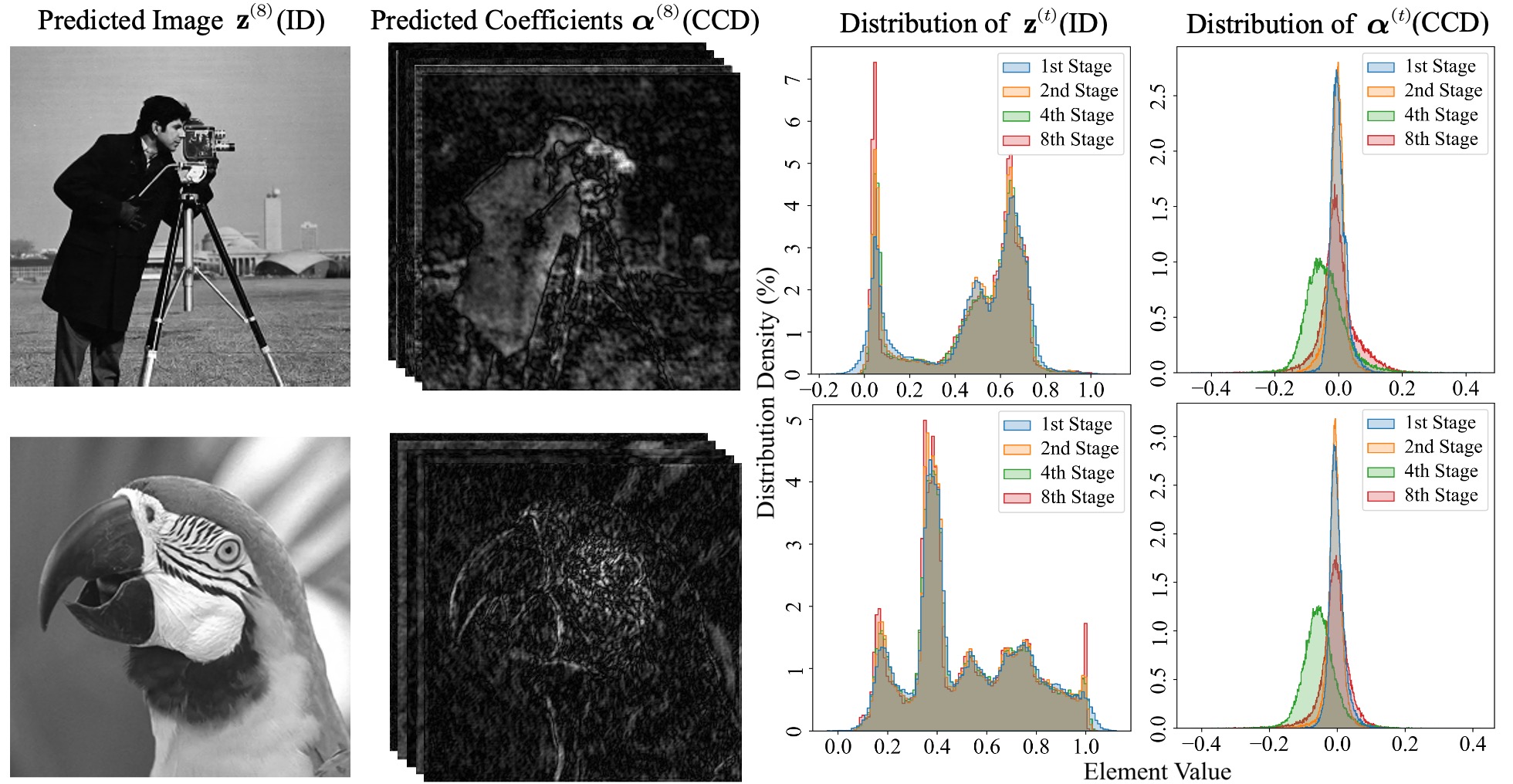}
    \caption{\textcolor{blue}{(Left)} Visualizations of the predicted image $\z^{(8)}$ (ID) and convolutional coefficients $\balpha^{(8)}$ (CCD). \textcolor{blue}{(Right)} The distribution curves of $\z^{(t)}$ and $\balpha^{(t)}$ with $t\in\{1,2,4,8\}$ and $\gamma=50\%$.}
    \label{fig:z_alpha}
\end{figure}

To gain some further insights into the role of low- and high-frequency (sparse) information in our $\DC$-Net, we visualize their distributions in Fig.~\ref{fig:lf_hf_0.1}. Our $\DC$-Net represents the image as a composition of one-channel low-frequency information and multi-channel high-frequency information through convolutional coding. This architectural design empowers our $\DC$-Net to effectively preserve, balance, and propagate different image components throughout the deep network trunk, leading to its remarkable reconstruction accuracy surpassing that of previous state-of-the-art methods. \mrev{Futhermore, the predicted image $\z^{(8)}$ (ID) and convolutional coefficients $\balpha^{(8)}$ (CCD) are visualized in Fig.~\ref{fig:z_alpha}. We observe that there is a distinct difference in the shapes of the distributions for $\z$ and $\balpha$. Specifically, the adaptively learned $\z$ data in different stages is obviously not sparse, while the distribution profile of $\balpha$ exhibits a markedly taller and narrower form, characterized by many values converging near zero. The distributions of $\balpha$ are more stable compared to the ones of $\z$, particularly in the front stages.} We conjecture that the increased feature capacity, sparsity, and stability can facilitate information transmission, well-balanced mid-/high-frequency details~\cite{chen2023deep}, and enhance the NN robustness against noise, thereby reinforcing the superiority of the dual-domain method over other single-domain ones.

\subsection{Extensibility and Generalization Ability of Our Method}

To validate the versatility, adaptability, and generalization ability of our method in addressing various inverse imaging tasks and real CS-based computational imaging systems, we conduct a series of experiments in this subsection. Specifically, we apply our $\DC$-Net to compressive sensing MRI, image inpainting, and sparse-view CT tasks. Furthermore, we establish a real SPI optics system for CS-based fluorescent microscopy.

\definecolor{OursColor}{HTML}{FFFFA5}
\definecolor{myIDBcolor}{HTML}{FFF0E6}
\definecolor{myCCDBcolor}{HTML}{C6FFBF}

\begin{table}[t!]
    \caption{Average PSNR (dB)/SSIM comparisons on CS-MRI with brain MR images \cite{clark2013cancer} \mrev{among eleven methods.}}
    \label{tab:mri}
    \centering
    \centering
    \normalsize
    \resizebox{1.0\linewidth}{!}{
    \begin{NiceTabular}{lIccccc}
    \CodeBefore
    \tikz \fill [gray!10] (1-|1) rectangle (3-|8);
    \tikz \fill [myIDBcolor](3-|1) rectangle (5-|8);
    \tikz \fill [myCCDBcolor](5-|1) rectangle (7-|8);
    \tikz \fill [myIDBcolor](7-|1) rectangle (9-|8);
    \tikz \fill [myCCDBcolor](9-|1) rectangle (11-|8);
    \tikz \fill [myIDBcolor](11-|1) rectangle (13-|8);
    \tikz \fill [myCCDBcolor](13-|1) rectangle (15-|8);
    \tikz \fill [myIDBcolor](15-|1) rectangle (17-|8);
    \tikz \fill [myCCDBcolor](17-|1) rectangle (19-|8);
    \tikz \fill [myIDBcolor](19-|1) rectangle (21-|8);
    \tikz \fill [myCCDBcolor](21-|1) rectangle (23-|8);
    \tikz \fill [OursColor](23-|1) rectangle (25-|8);    
    \Body
    \shline
     \multicolumn{1}{l|}{\multirow{2}{*}{Method}} & \multicolumn{5}{c}{CS Ratio} \\
     \arrayrulecolor[HTML]{CFCFCF}\cline{2-6}
    & 10\%     & 20\%     & 30\% & 40\% & 50\% \\
    \arrayrulecolor{black}\hline \hline 
\multicolumn{1}{l|}{\multirow{2}{*}{
\begin{tabular}[l]{@{}l@{}}\hspace{-3pt}Schlemper et al.~\cite{schlemper2017deep}\\ \hspace{-3pt}(TMI 2017) \end{tabular} }}
     &
\multicolumn{1}{c}{\multirow{2}{*}{
\begin{tabular}[c]{@{}c@{}}34.20\\ /0.8921 \end{tabular} }}
     &
\multicolumn{1}{c}{\multirow{2}{*}{
\begin{tabular}[c]{@{}c@{}}38.47\\ /0.9457 \end{tabular} }}     
     &
\multicolumn{1}{c}{\multirow{2}{*}{
\begin{tabular}[c]{@{}c@{}}40.85\\ /0.9628 \end{tabular} }}       
     &
\multicolumn{1}{c}{\multirow{2}{*}{
\begin{tabular}[c]{@{}c@{}}42.63\\ /0.9724 \end{tabular} }}       
     &
\multicolumn{1}{c}{\multirow{2}{*}{
\begin{tabular}[c]{@{}c@{}}44.19\\ /0.9794 \end{tabular} }}       
     \\
     & & & & & \\
\multicolumn{1}{l|}{\multirow{2}{*}{
\begin{tabular}[l]{@{}l@{}}\hspace{-3pt}Hyun et al.~\cite{hyun2018deep} \\ \hspace{-3pt}(PMB 2018) \end{tabular} }}
&
\multicolumn{1}{c}{\multirow{2}{*}{
\begin{tabular}[c]{@{}c@{}}32.78\\ /0.8385 \end{tabular} }}    
&
\multicolumn{1}{c}{\multirow{2}{*}{
\begin{tabular}[c]{@{}c@{}}36.36\\ /0.9070 \end{tabular} }}   
&
\multicolumn{1}{c}{\multirow{2}{*}{
\begin{tabular}[c]{@{}c@{}}38.85\\ /0.9383 \end{tabular} }}   
&
\multicolumn{1}{c}{\multirow{2}{*}{
\begin{tabular}[c]{@{}c@{}}40.65\\ /0.9539 \end{tabular} }}   
&
\multicolumn{1}{c}{\multirow{2}{*}{
\begin{tabular}[c]{@{}c@{}}42.35\\ /0.9662 \end{tabular} }}   
\\
     & & & & & \\
\multicolumn{1}{l|}{\multirow{2}{*}{
\begin{tabular}[l]{@{}l@{}}\hspace{-3pt}ADMM-Net~\cite{yang2018admm} \\ \hspace{-3pt}(TPAMI 2018) \end{tabular} }}
&
\multicolumn{1}{c}{\multirow{2}{*}{
\begin{tabular}[c]{@{}c@{}}34.42\\ /0.8971 \end{tabular} }}    
&
\multicolumn{1}{c}{\multirow{2}{*}{
\begin{tabular}[c]{@{}c@{}}38.60\\ /0.9478 \end{tabular} }}   
&
\multicolumn{1}{c}{\multirow{2}{*}{
\begin{tabular}[c]{@{}c@{}}40.87\\ /0.9633 \end{tabular} }}   
&
\multicolumn{1}{c}{\multirow{2}{*}{
\begin{tabular}[c]{@{}c@{}}42.58\\ /0.9726 \end{tabular} }}   
&
\multicolumn{1}{c}{\multirow{2}{*}{
\begin{tabular}[c]{@{}c@{}}44.19\\ /0.9796 \end{tabular} }}   
\\
     & & & & & \\
\multicolumn{1}{l|}{\multirow{2}{*}{
\begin{tabular}[l]{@{}l@{}}\hspace{-3pt}RDN~\cite{sun2018compressed} \\ \hspace{-3pt}(AAAI 2018) \end{tabular} }}
&
\multicolumn{1}{c}{\multirow{2}{*}{
\begin{tabular}[c]{@{}c@{}}34.59\\ /0.8968 \end{tabular} }}    
&
\multicolumn{1}{c}{\multirow{2}{*}{
\begin{tabular}[c]{@{}c@{}}38.58\\ /0.9470 \end{tabular} }}   
&
\multicolumn{1}{c}{\multirow{2}{*}{
\begin{tabular}[c]{@{}c@{}}40.82\\ /0.9625 \end{tabular} }}   
&
\multicolumn{1}{c}{\multirow{2}{*}{
\begin{tabular}[c]{@{}c@{}}42.64\\ /0.9723 \end{tabular} }}   
&
\multicolumn{1}{c}{\multirow{2}{*}{
\begin{tabular}[c]{@{}c@{}}44.18\\ /0.9793\end{tabular} }}   
\\
     & & & & & \\
\multicolumn{1}{l|}{\multirow{2}{*}{
\begin{tabular}[l]{@{}l@{}}\hspace{-3pt}ISTA-Net$^+$~\cite{zhang2018ista} \\ \hspace{-3pt}(CVPR 2018) \end{tabular} }}
&
\multicolumn{1}{c}{\multirow{2}{*}{
\begin{tabular}[c]{@{}c@{}}34.65\\ /0.9038 \end{tabular} }}    
&
\multicolumn{1}{c}{\multirow{2}{*}{
\begin{tabular}[c]{@{}c@{}}38.67\\ /0.9480 \end{tabular} }}   
&
\multicolumn{1}{c}{\multirow{2}{*}{
\begin{tabular}[c]{@{}c@{}}40.91\\ /0.9631 \end{tabular} }}   
&
\multicolumn{1}{c}{\multirow{2}{*}{
\begin{tabular}[c]{@{}c@{}}42.65\\ /0.9727 \end{tabular} }}   
&
\multicolumn{1}{c}{\multirow{2}{*}{
\begin{tabular}[c]{@{}c@{}}44.24\\ /0.9798\end{tabular} }}   
\\
     & & & & & \\
\multicolumn{1}{l|}{\multirow{2}{*}{
\begin{tabular}[l]{@{}l@{}}\hspace{-3pt}MoDL~\cite{aggarwal2018modl} \\ \hspace{-3pt}(TMI 2018)\end{tabular} }}
&
\multicolumn{1}{c}{\multirow{2}{*}{
\begin{tabular}[c]{@{}c@{}}35.18\\ /0.9091 \end{tabular} }}    
&
\multicolumn{1}{c}{\multirow{2}{*}{
\begin{tabular}[c]{@{}c@{}}38.51\\ /0.9457 \end{tabular} }}   
&
\multicolumn{1}{c}{\multirow{2}{*}{
\begin{tabular}[c]{@{}c@{}}40.97\\ /0.9636 \end{tabular} }}   
&
\multicolumn{1}{c}{\multirow{2}{*}{
\begin{tabular}[c]{@{}c@{}}42.38\\ /0.9705 \end{tabular} }}   
&
\multicolumn{1}{c}{\multirow{2}{*}{
\begin{tabular}[c]{@{}c@{}}44.20\\ /0.9776\end{tabular} }}   
\\
     & & & & & \\
\multicolumn{1}{l|}{\multirow{2}{*}{
\begin{tabular}[l]{@{}l@{}}\hspace{-3pt}CDDN~\cite{zheng2019cascaded} \\ \hspace{-3pt}(NeurIPS 2019)\end{tabular} }}
&
\multicolumn{1}{c}{\multirow{2}{*}{
\begin{tabular}[c]{@{}c@{}}34.63\\ /0.9002 \end{tabular} }}    
&
\multicolumn{1}{c}{\multirow{2}{*}{
\begin{tabular}[c]{@{}c@{}}38.59\\ /0.9474 \end{tabular} }}   
&
\multicolumn{1}{c}{\multirow{2}{*}{
\begin{tabular}[c]{@{}c@{}}40.89\\ /0.9633 \end{tabular} }}   
&
\multicolumn{1}{c}{\multirow{2}{*}{
\begin{tabular}[c]{@{}c@{}}42.59\\ /0.9725 \end{tabular} }}   
&
\multicolumn{1}{c}{\multirow{2}{*}{
\begin{tabular}[c]{@{}c@{}}44.15\\ /0.9795\end{tabular} }}   
\\
     & & & & & \\    
\multicolumn{1}{l|}{\multirow{2}{*}{
\begin{tabular}[l]{@{}l@{}}\hspace{-3pt}HiTDUN~\cite{zhang2022high} \\ \hspace{-3pt}(JSTSP 2022)\end{tabular} }}
&
\multicolumn{1}{c}{\multirow{2}{*}{
\begin{tabular}[c]{@{}c@{}}35.71\\ /0.9179 \end{tabular} }}    
&
\multicolumn{1}{c}{\multirow{2}{*}{
\begin{tabular}[c]{@{}c@{}}39.27\\ /0.9529 \end{tabular} }}   
&
\multicolumn{1}{c}{\multirow{2}{*}{
\begin{tabular}[c]{@{}c@{}}41.37\\ /0.9660 \end{tabular} }}   
&
\multicolumn{1}{c}{\multirow{2}{*}{
\begin{tabular}[c]{@{}c@{}}-\\ /- \end{tabular} }}   
&
\multicolumn{1}{c}{\multirow{2}{*}{
\begin{tabular}[c]{@{}c@{}}-\\ /-\end{tabular} }}   
\\
     & & & & & \\ 
\multicolumn{1}{l|}{\multirow{2}{*}{
\begin{tabular}[l]{@{}l@{}}\hspace{-3pt}MADUN~\cite{song2021memory} \\ \hspace{-3pt}(ACM MM 2021)\end{tabular} }}
&
\multicolumn{1}{c}{\multirow{2}{*}{
\begin{tabular}[c]{@{}c@{}}\underline{\textcolor{blue}{36.15}}\\ /0.9237 \end{tabular} }}    
&
\multicolumn{1}{c}{\multirow{2}{*}{
\begin{tabular}[c]{@{}c@{}}\underline{\textcolor{blue}{39.44}}\\ /\underline{\textcolor{blue}{0.9542}} \end{tabular} }} 
&
\multicolumn{1}{c}{\multirow{2}{*}{
\begin{tabular}[c]{@{}c@{}}\underline{\textcolor{blue}{41.48}}\\ /\underline{\textcolor{blue}{0.9666}} \end{tabular} }}   
&
\multicolumn{1}{c}{\multirow{2}{*}{
\begin{tabular}[c]{@{}c@{}}\underline{\textcolor{blue}{43.06}}\\ /\underline{\textcolor{blue}{0.9746}} \end{tabular} }}   
&
\multicolumn{1}{c}{\multirow{2}{*}{
\begin{tabular}[c]{@{}c@{}}\underline{\textcolor{blue}{44.60}}\\ /\underline{\textcolor{blue}{0.9810}}\end{tabular} }}   
\\
     & & & & & \\ 
\multicolumn{1}{l|}{\multirow{2}{*}{
\begin{tabular}[l]{@{}l@{}}\hspace{-3pt}MAPUN~\cite{song2023deep} \\ \hspace{-3pt}(IJCV 2023)\end{tabular} }}
&
\multicolumn{1}{c}{\multirow{2}{*}{
\begin{tabular}[c]{@{}c@{}}36.14\\ /\msecondbest{0.9242} \end{tabular} }}    
&
\multicolumn{1}{c}{\multirow{2}{*}{
\begin{tabular}[c]{@{}c@{}}39.40\\ /0.9541 \end{tabular} }}   
&
\multicolumn{1}{c}{\multirow{2}{*}{
\begin{tabular}[c]{@{}c@{}}41.42\\ /0.9664 \end{tabular} }}   
&
\multicolumn{1}{c}{\multirow{2}{*}{
\begin{tabular}[c]{@{}c@{}}43.02\\ /0.9745 \end{tabular} }}   
&
\multicolumn{1}{c}{\multirow{2}{*}{
\begin{tabular}[c]{@{}c@{}}44.55\\ /0.9809\end{tabular} }}   
\\
     & & & & & \\ 
\multicolumn{1}{l|}{\multirow{2}{*}{
\begin{tabular}[l]{@{}l@{}}\hspace{-3pt}\textbf{$\DC$-Net}\\ \hspace{-3pt}\textbf{(Ours)}\end{tabular} }}
&
\multicolumn{1}{c}{\multirow{2}{*}{
\begin{tabular}[c]{@{}c@{}}\textbf{\textcolor{red}{36.48}}\\ /\textbf{\textcolor{red}{0.9289}} \end{tabular} }}    
&
\multicolumn{1}{c}{\multirow{2}{*}{
\begin{tabular}[c]{@{}c@{}}\textbf{\textcolor{red}{39.66}}\\ /\textbf{\textcolor{red}{0.9558}} \end{tabular} }} 
&
\multicolumn{1}{c}{\multirow{2}{*}{
\begin{tabular}[c]{@{}c@{}}\textbf{\textcolor{red}{41.59}}\\ /\textbf{\textcolor{red}{0.9671}} \end{tabular} }}   
&
\multicolumn{1}{c}{\multirow{2}{*}{
\begin{tabular}[c]{@{}c@{}}\textbf{\textcolor{red}{43.14}}\\ /\textbf{\textcolor{red}{0.9748}} \end{tabular} }}   
&
\multicolumn{1}{c}{\multirow{2}{*}{
\begin{tabular}[c]{@{}c@{}}\textbf{\textcolor{red}{44.63}}\\ /\textbf{\textcolor{red}{0.9811}}\end{tabular} }}   
\\
     & & & & & \\   
     \shline
    \end{NiceTabular}
    }
\vspace{-6pt}
\end{table}
\begin{figure}[t!]
    \centering
    \small
    \setlength{\tabcolsep}{0.4pt}
    \renewcommand{\arraystretch}{0.4}
    \scriptsize
    \begin{tabular}{cccccc}
         ADMM-Net &  CDDN &  ISTA-Net$^{+}$ & MoDL&  MADUN&  $\DC$-Net\\
        \includegraphics[width=0.16\linewidth]{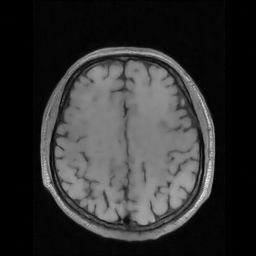} &
        \includegraphics[width=0.16\linewidth]{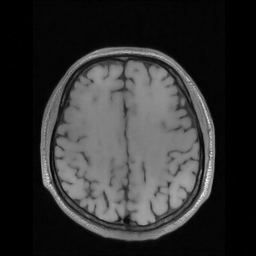} &
        \includegraphics[width=0.16\linewidth]{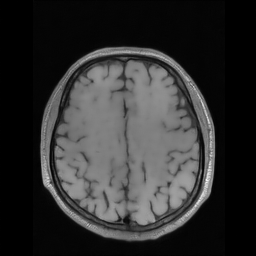} &
        \includegraphics[width=0.16\linewidth]{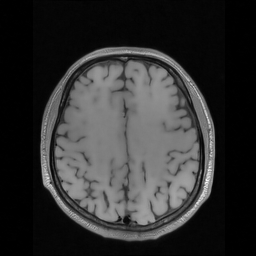}&
        \includegraphics[width=0.16\linewidth]{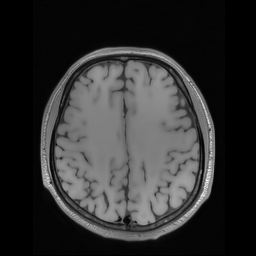}&
        \includegraphics[width=0.16\linewidth]{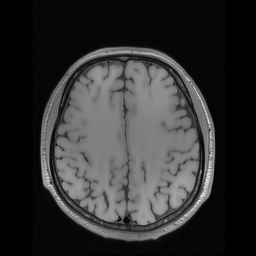}\\
        
        \includegraphics[width=0.16\linewidth]{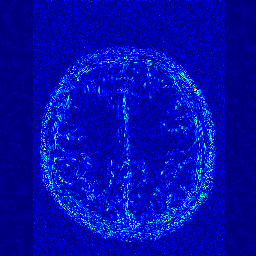} &
        \includegraphics[width=0.16\linewidth]{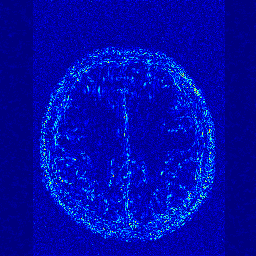} &
        \includegraphics[width=0.16\linewidth]{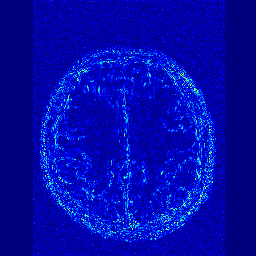} &
        \includegraphics[width=0.16\linewidth]{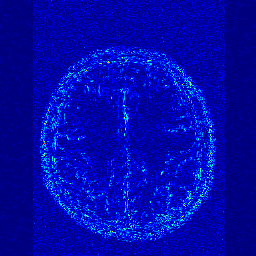}&
        \includegraphics[width=0.16\linewidth]{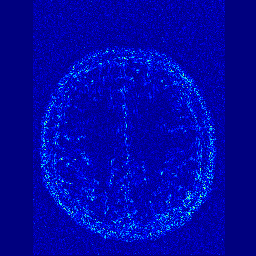}&
        \includegraphics[width=0.16\linewidth]{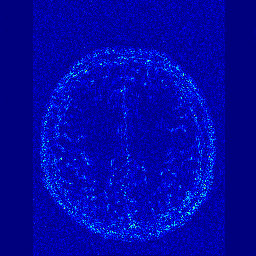}\\
        33.09 dB& 33.44 dB&33.36 dB& 34.03 dB& \underline{\textcolor{blue}{35.13}} dB & \textbf{\textcolor{red}{35.66}} dB\\
    \end{tabular}
    \vspace{-5pt}
    \caption{Visual comparisons on CS-MRI recovery~\textcolor{blue}{(top)} and absolute residual maps to ground truth~\textcolor{blue}{(bottom)} of six methods on the image named ``brain-test-13''~\cite{clark2013cancer} with $\gamma= 10\%$.}
    \label{fig:MRI13}
    \vspace{-5pt}
\end{figure}

\textbf{Application to Compressive Sensing MRI:} We extend the application of $\DC$-Net to the practical inverse problem of reconstructing MR images in the context of compressive sensing MRI (CS-MRI). The goal is to restore MR images from a limited number of under-sampled data points in $k$-space. To address this problem, we adopt a common approach by setting the measurement matrix as $\bPhi:= \mathbf{B}\mathbf{F}$, where $\mathbf{B}$ represents an under-sampling matrix and $\mathbf{F}$ denotes the discrete Fourier transform. Following the methodology of MADUN~\cite{song2021memory}, we utilize a training set consisting of 100 fully sampled brain MR images. To mitigate the risk of overfitting with this limited dataset, we adjust the architecture of $\DC$-Net for MRI by reducing its width and increasing its depth. The resulting $\DC$-Net for MRI has $C=32$ and $T=20$, with parameter number 1.72M, which is lower than 2.72M of standard $\DC$-Net for natural images. We also apply geometric data augmentation techniques to enhance the diversity of training data.

\mrev{As demonstrated in Tabs.~\ref{tab:low_ratios} and~\ref{tab:mri}, our $\DC$-Net surpasses state-of-the-art methods across all the seven given CS ratios.} It is worth noting that achieving higher PSNR leading becomes more challenging as the ratio increases since the NN performance becomes saturated. Visual comparisons of reconstructed MR images and corresponding error maps, compared with the ground truth, are shown in Fig.~\ref{fig:MRI13}. The results clearly illustrate that $\DC$-Net produces highly accurate recoveries with reduced overall errors and enhanced details of brain tissue when compared to other competing methods. This confirms the superiority and generalizability of our proposed approach. Furthermore, thanks to the incorporation of CS ratio information in InitNet and HPN, $\DC$-Net for MRI exhibits scalability for different ratios. This means that a single model can handle multiple ratios, resulting in a significant reduction in overall parameter count.  Compared with MADUN (3.13M parameters for each ratio), $\DC$-Net leverages only about $(1/9) \times$ number of parameters while achieving better reconstruction performance on CS-MRI.

\begin{figure}
    \centering
    \scriptsize
    \setlength{\tabcolsep}{0.6pt}
    \renewcommand{\arraystretch}{0.6}
        \vspace{-6pt}
    
    \resizebox{1.0\linewidth}{!}{
    
    \begin{tabular}{ccccc}
        Ground Truth & $\mathbf{A}^{\dagger}\mathbf{y}$ & ID-Only & CCD-Only & Ours \\
        \includegraphics[width=0.19\linewidth]{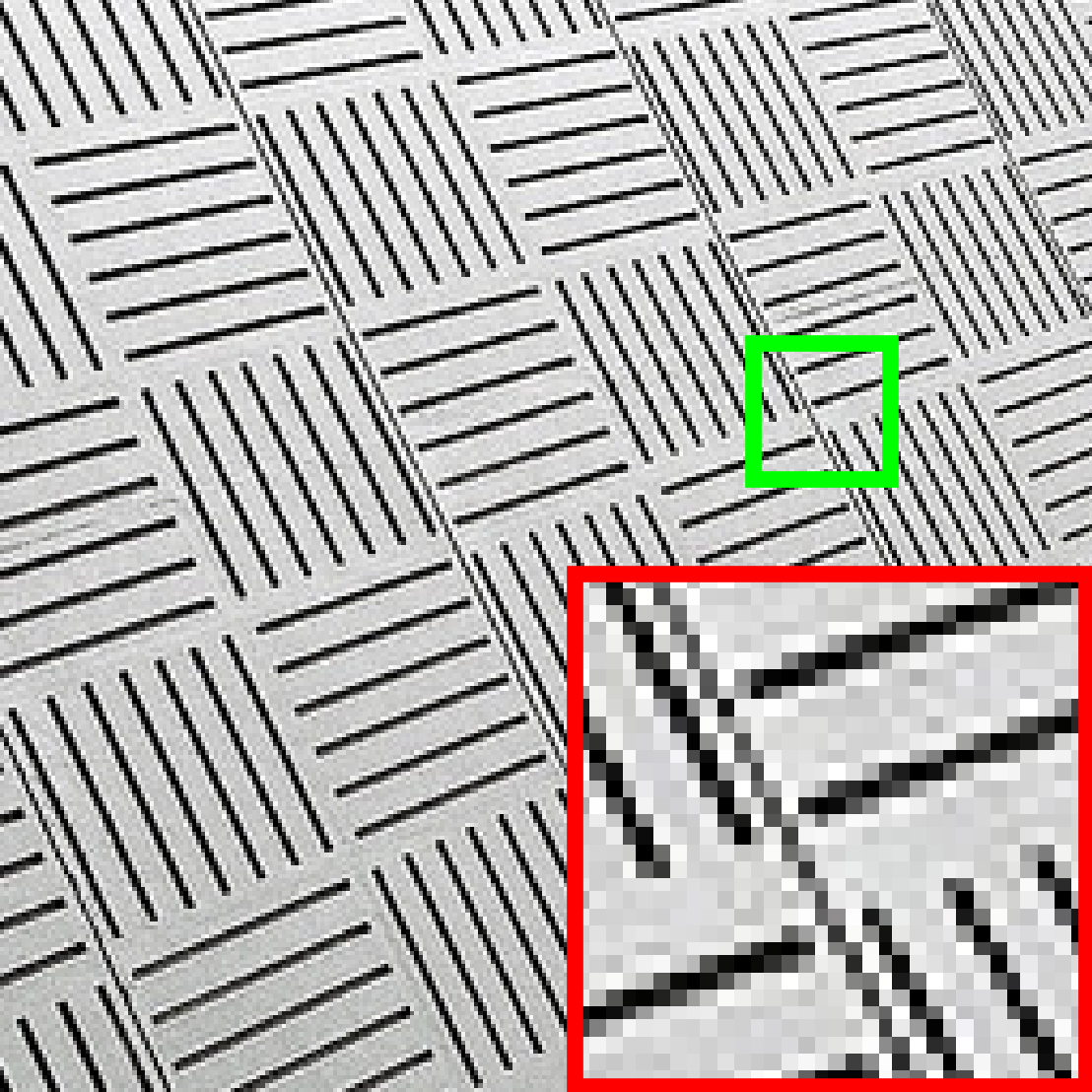} &         \includegraphics[width=0.19\linewidth]{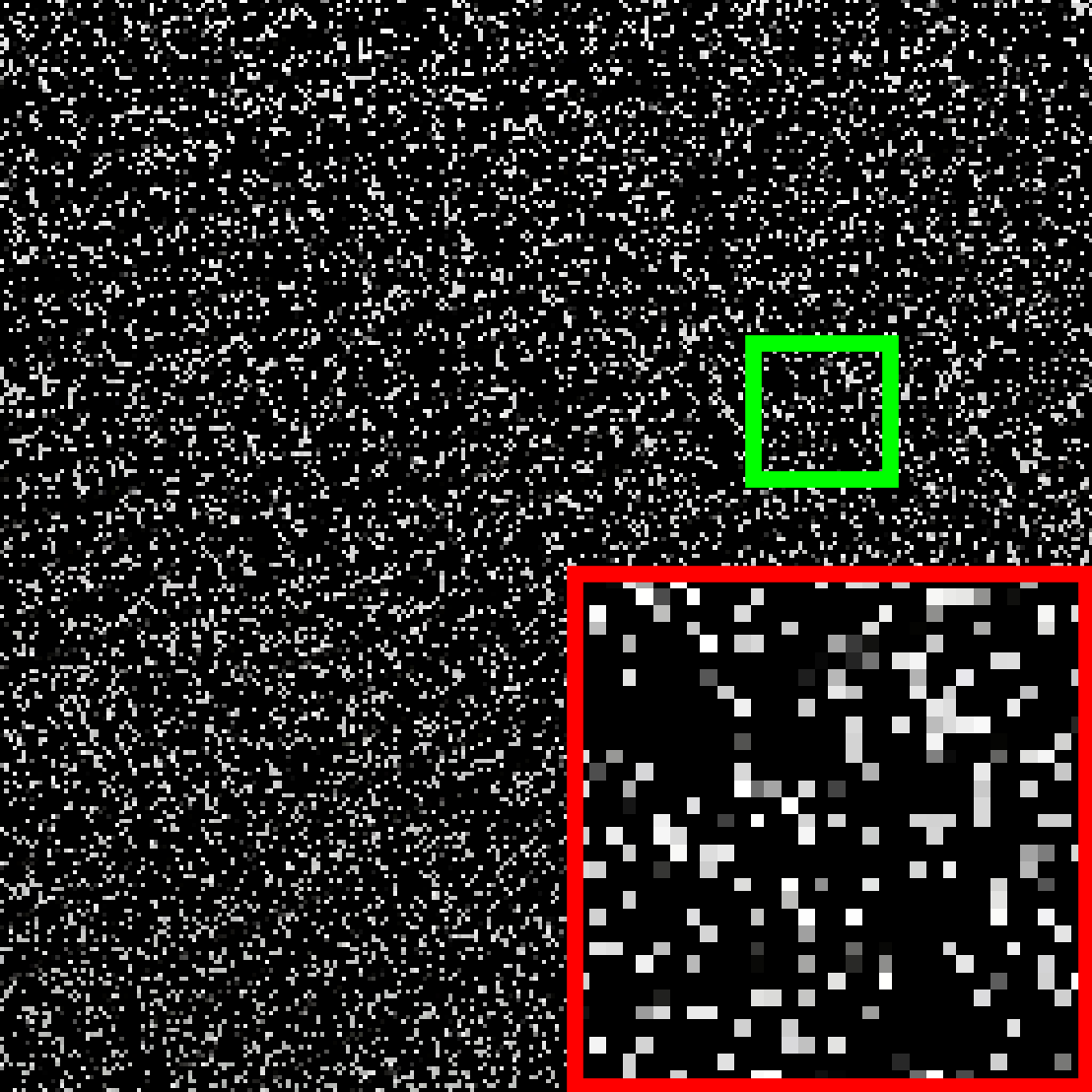}&     \includegraphics[width=0.19\linewidth]{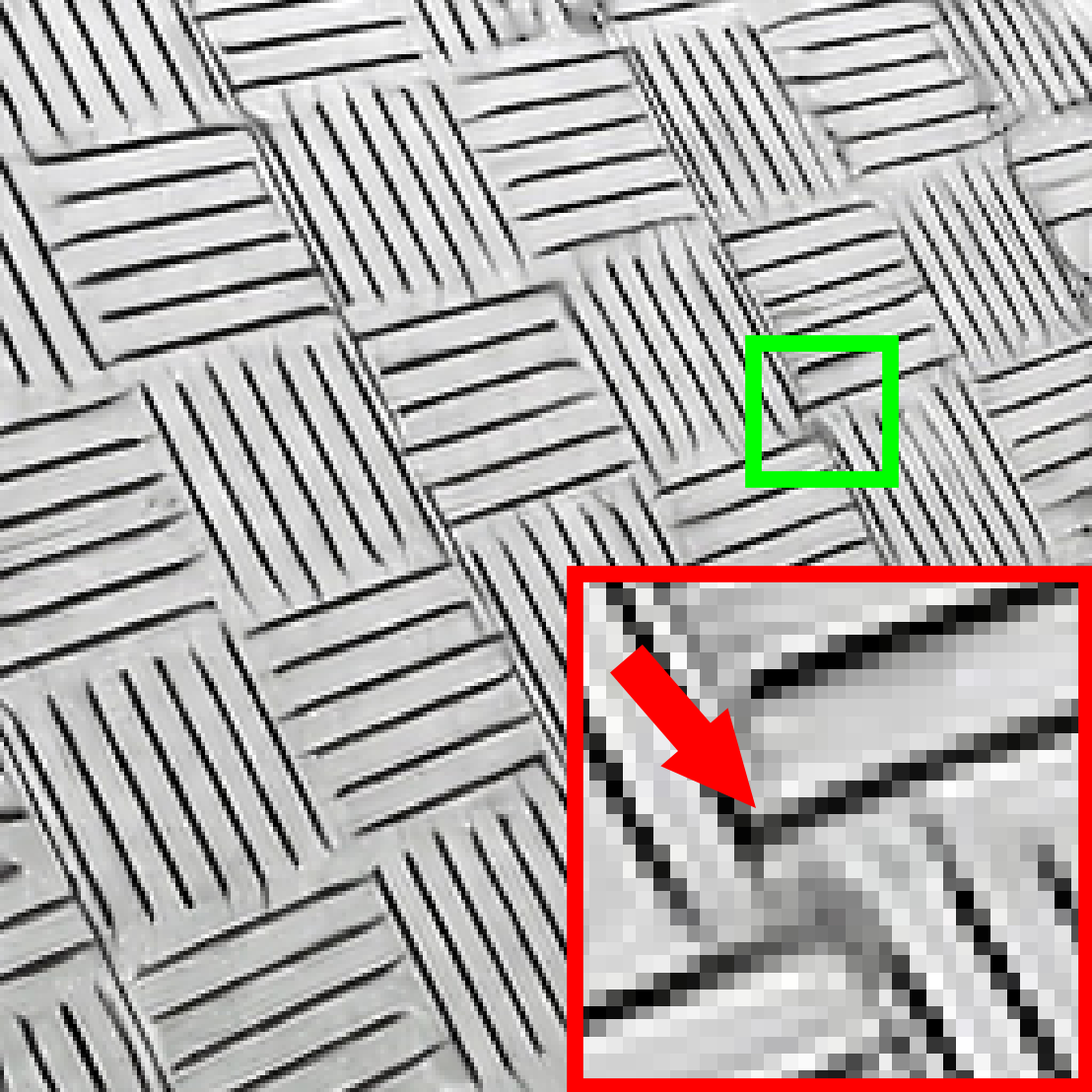}&     \includegraphics[width=0.19\linewidth]{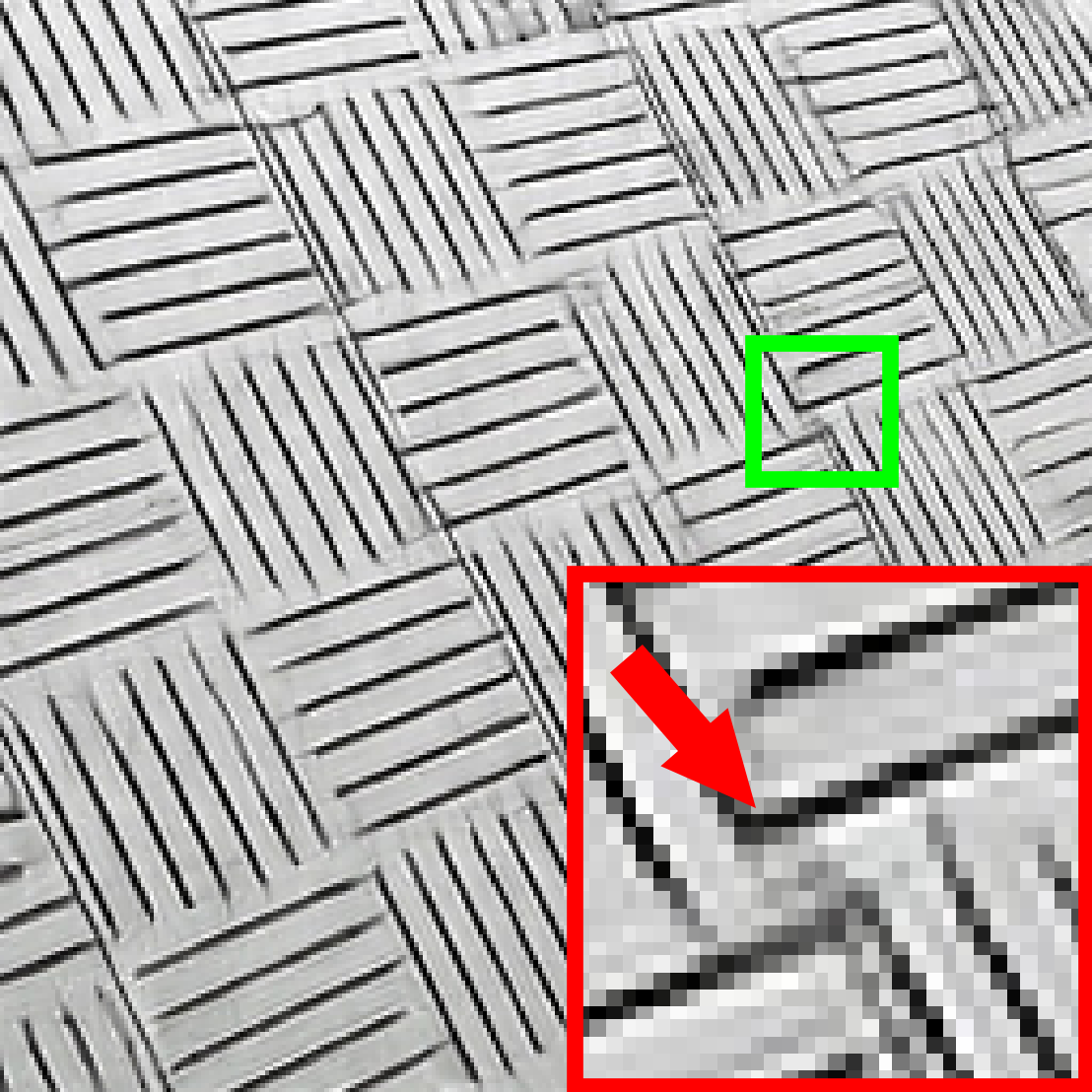} &    \includegraphics[width=0.19\linewidth]{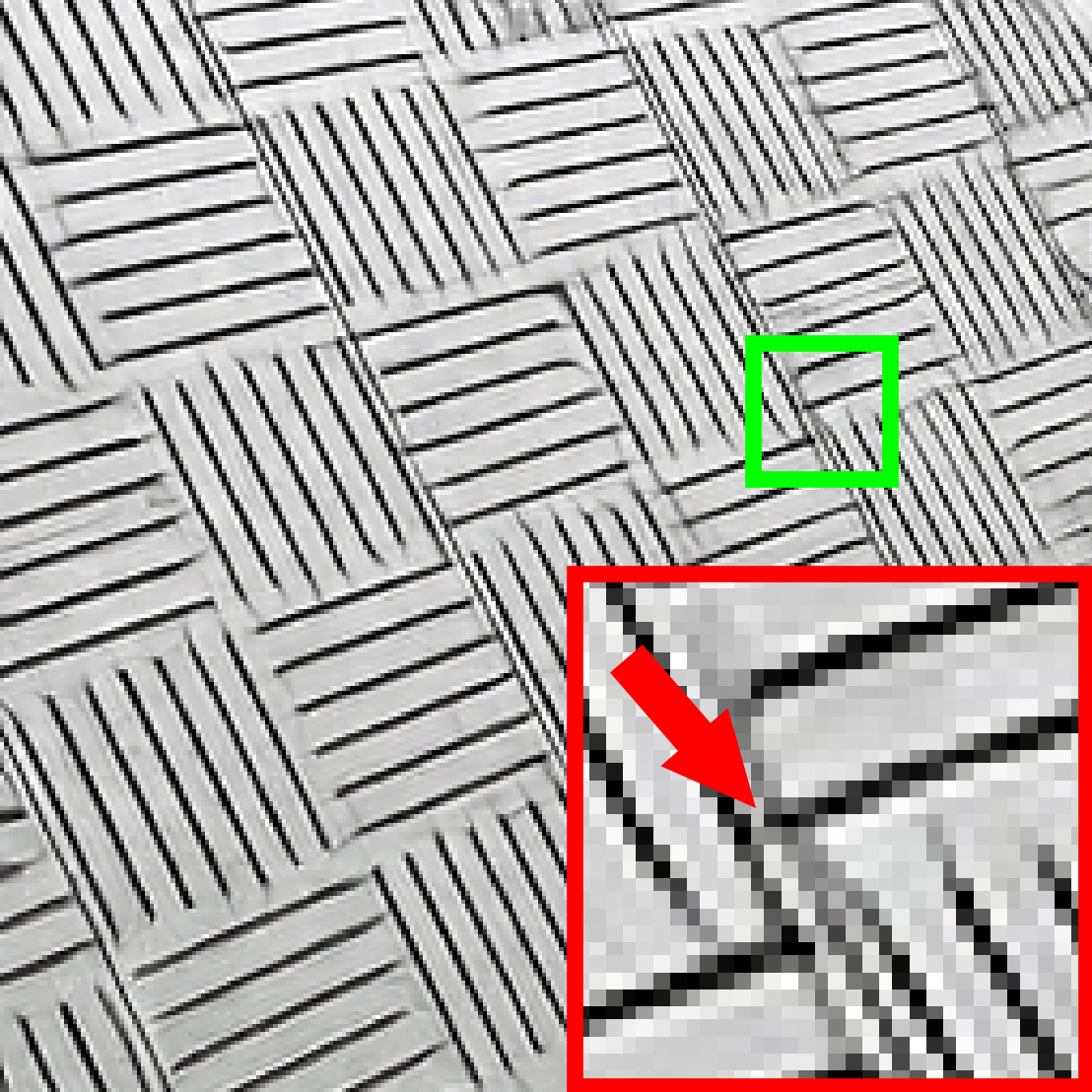} \\
        PSNR/SSIM & 3.60/0.0697 & \underline{\textcolor{blue}{19.74}}/\underline{\textcolor{blue}{0.8858}} & 19.30/0.8757 & \textbf{\textcolor{red}{20.76}}/\textbf{\textcolor{red}{0.9088}} \\ 

        \includegraphics[width=0.19\linewidth]{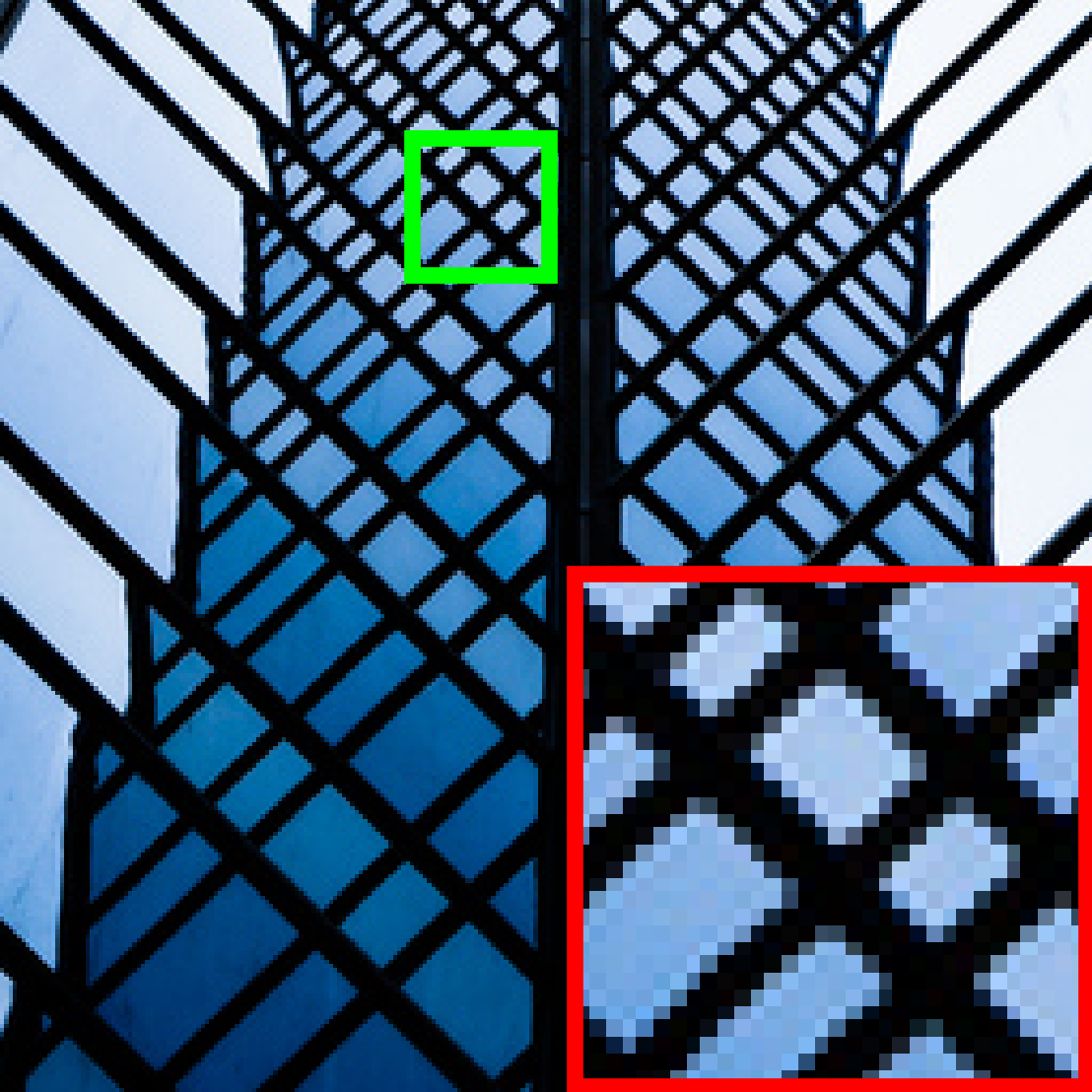} &         \includegraphics[width=0.19\linewidth]{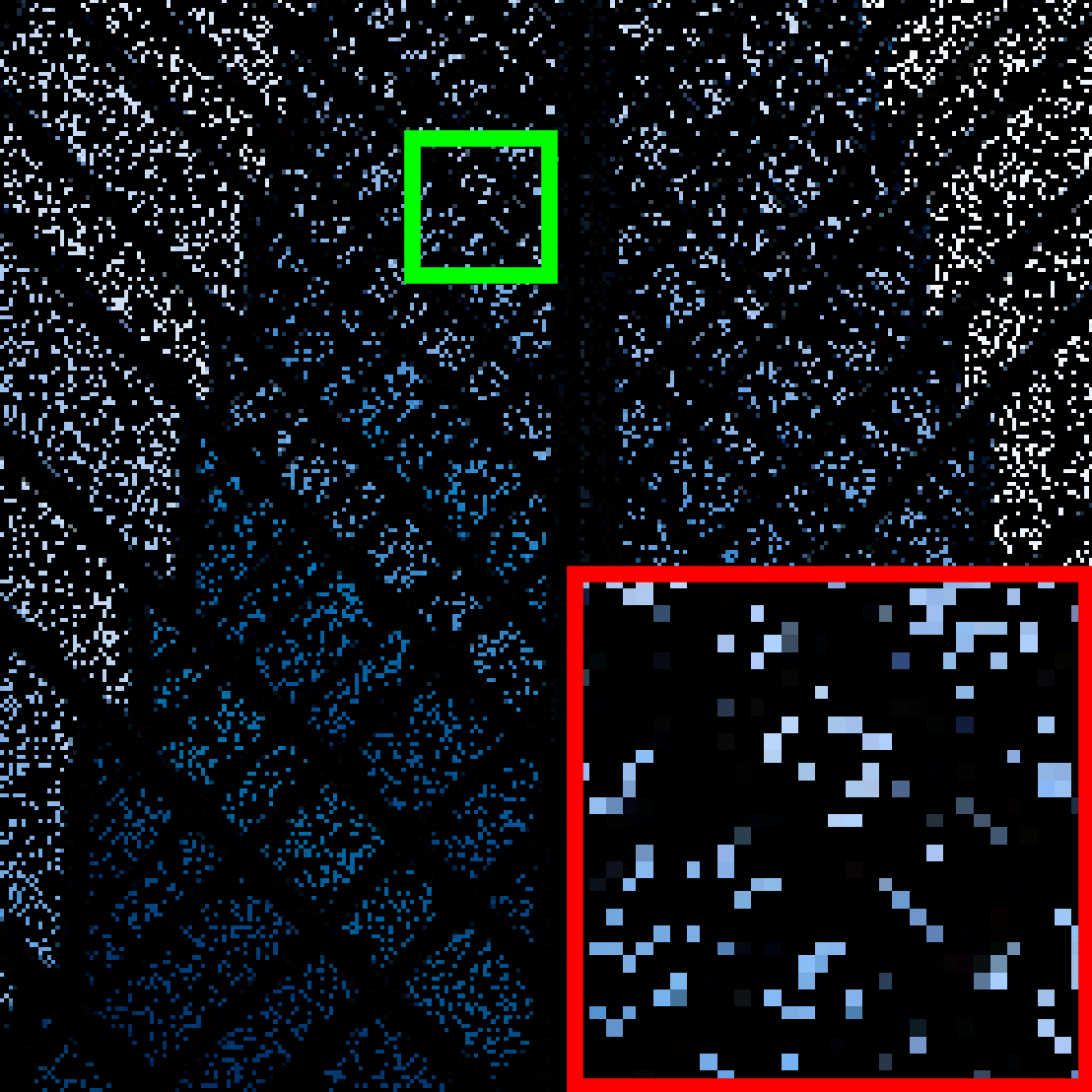}&     \includegraphics[width=0.19\linewidth]{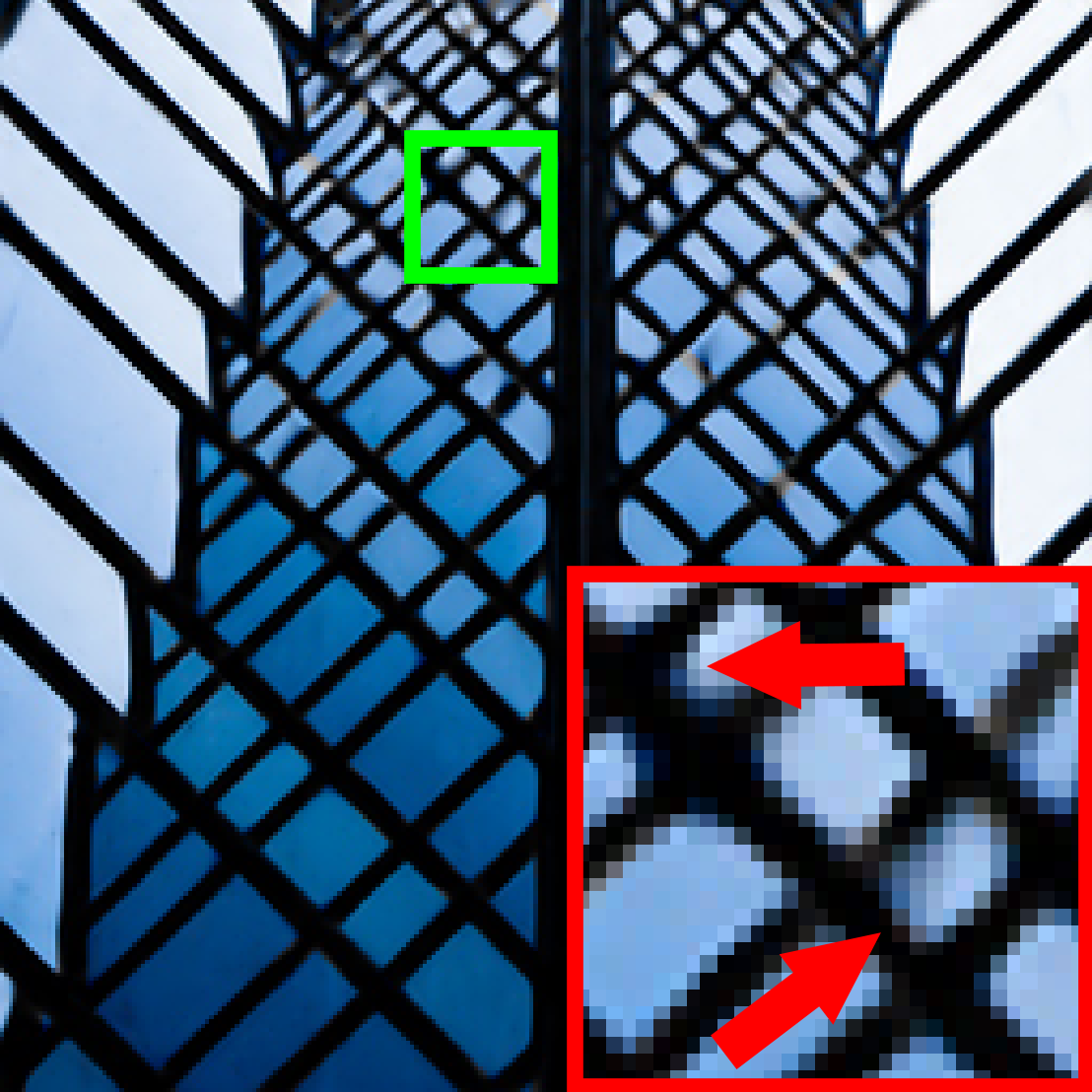}&     \includegraphics[width=0.19\linewidth]{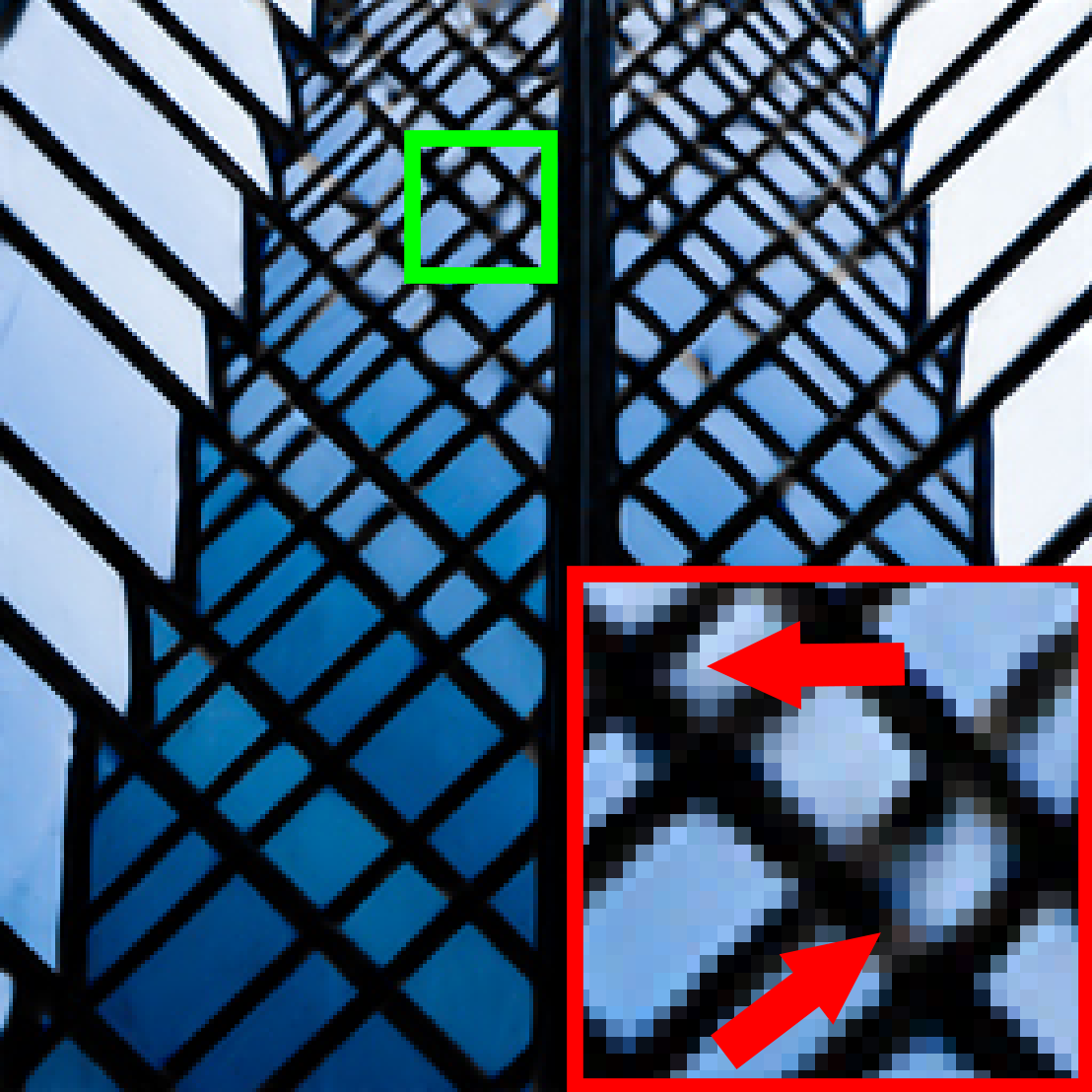} &    \includegraphics[width=0.19\linewidth]{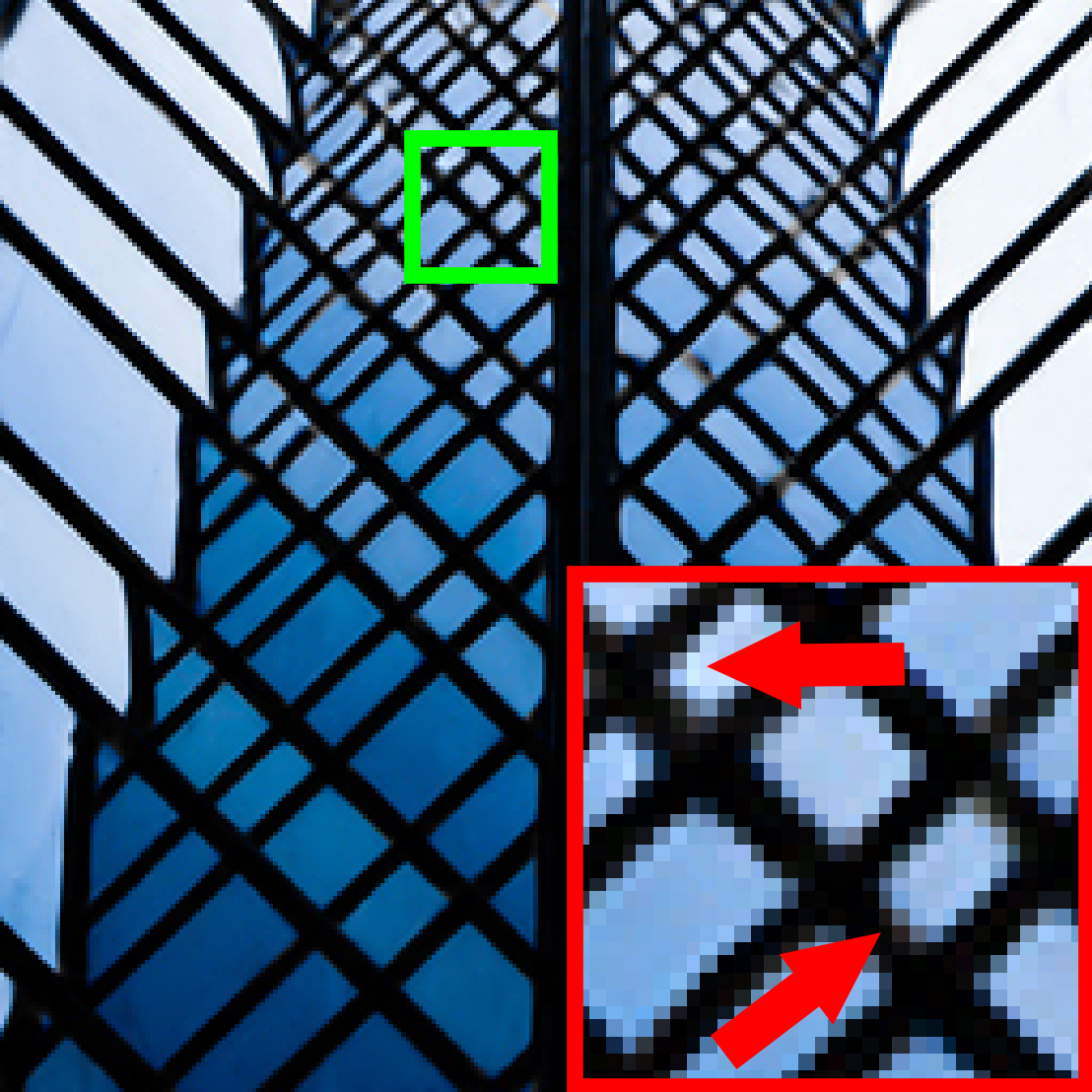} \\
        PSNR/SSIM & 7.62/0.1258 & 23.07/0.9414 & \underline{\textcolor{blue}{23.41}}/\underline{\textcolor{blue}{0.9446}} & \textbf{\textcolor{red}{24.64}}/\textbf{\textcolor{red}{0.9565}} \\

        \includegraphics[width=0.19\linewidth]{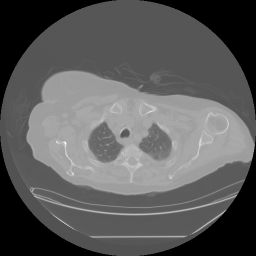} &         \includegraphics[width=0.19\linewidth]{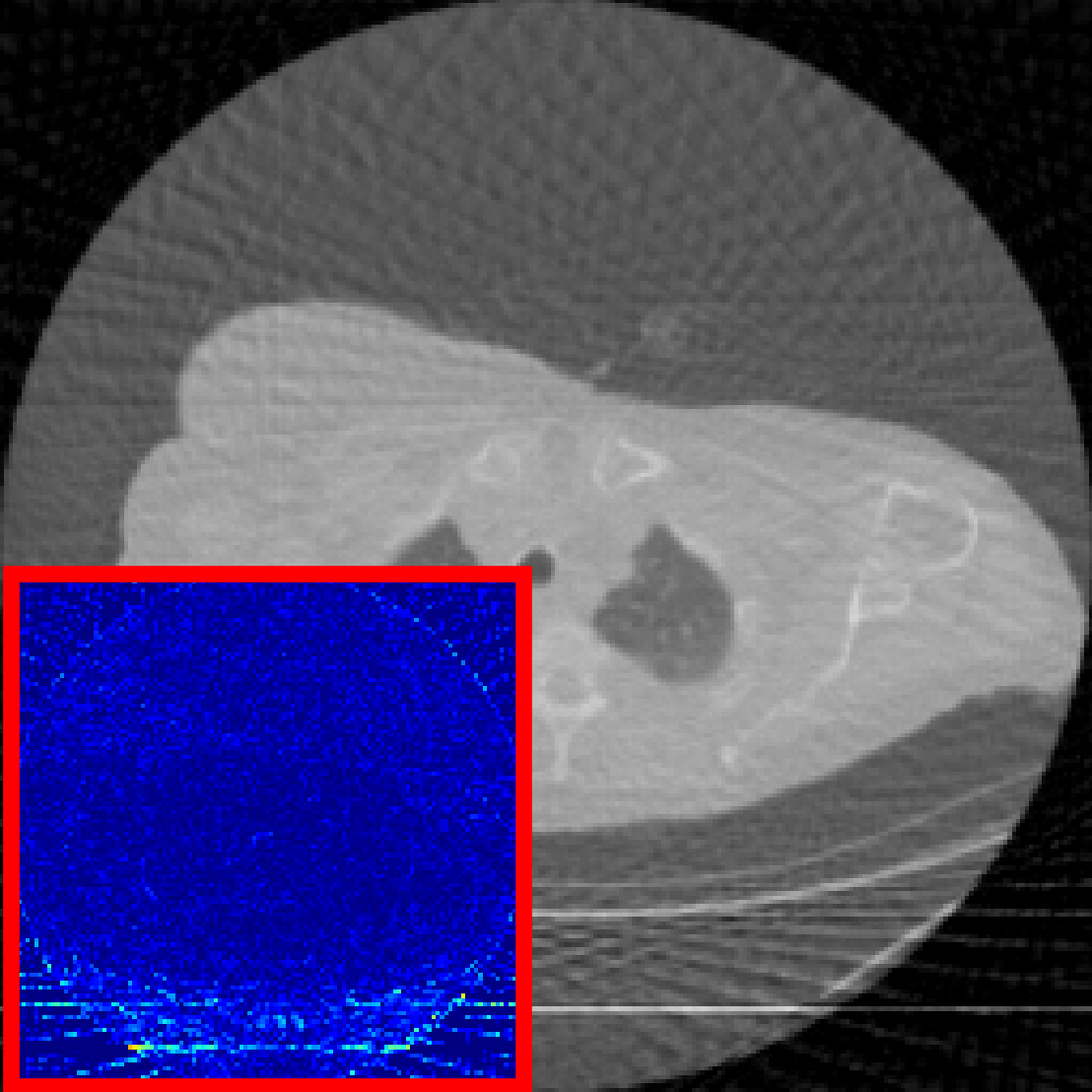}&     \includegraphics[width=0.19\linewidth]{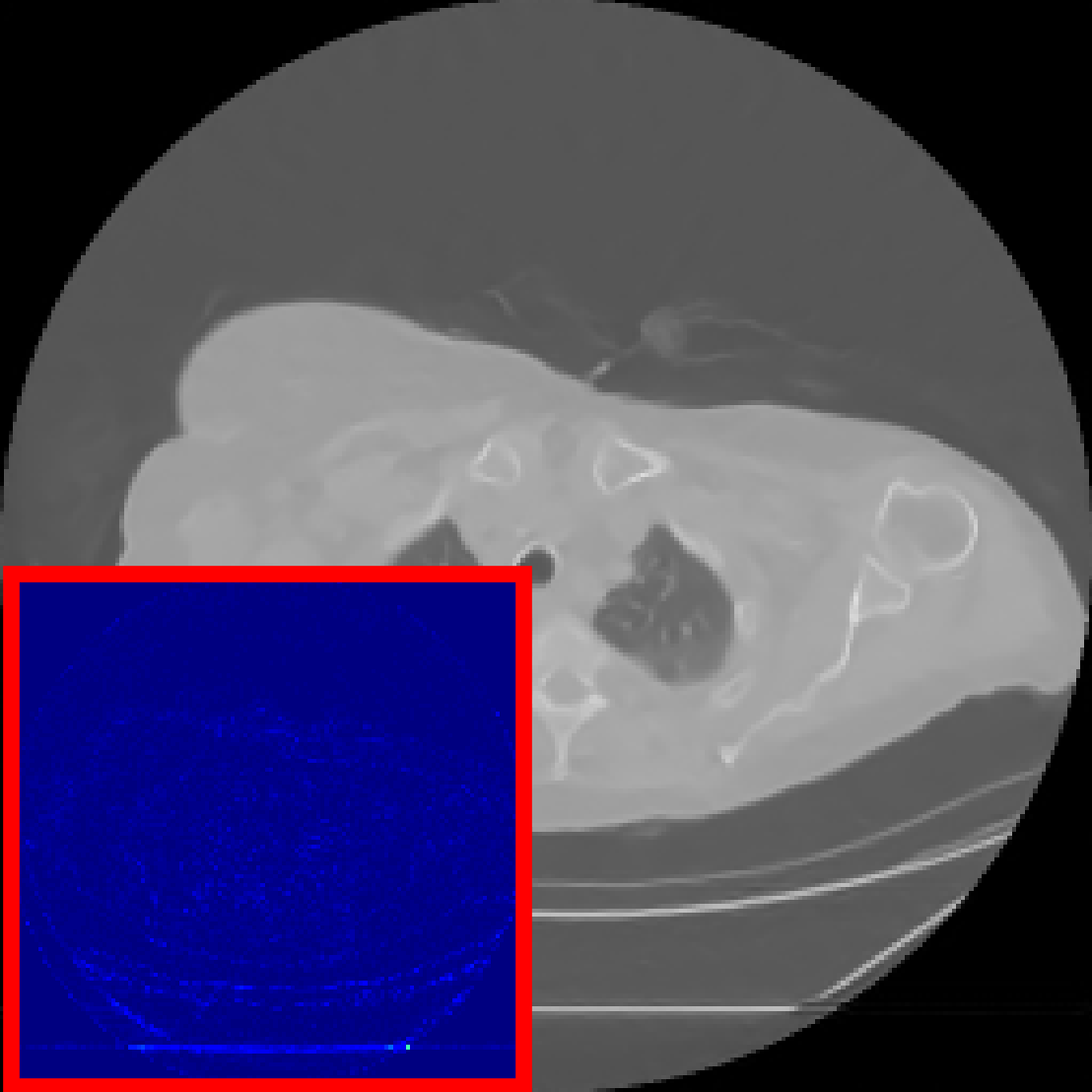}&     \includegraphics[width=0.19\linewidth]{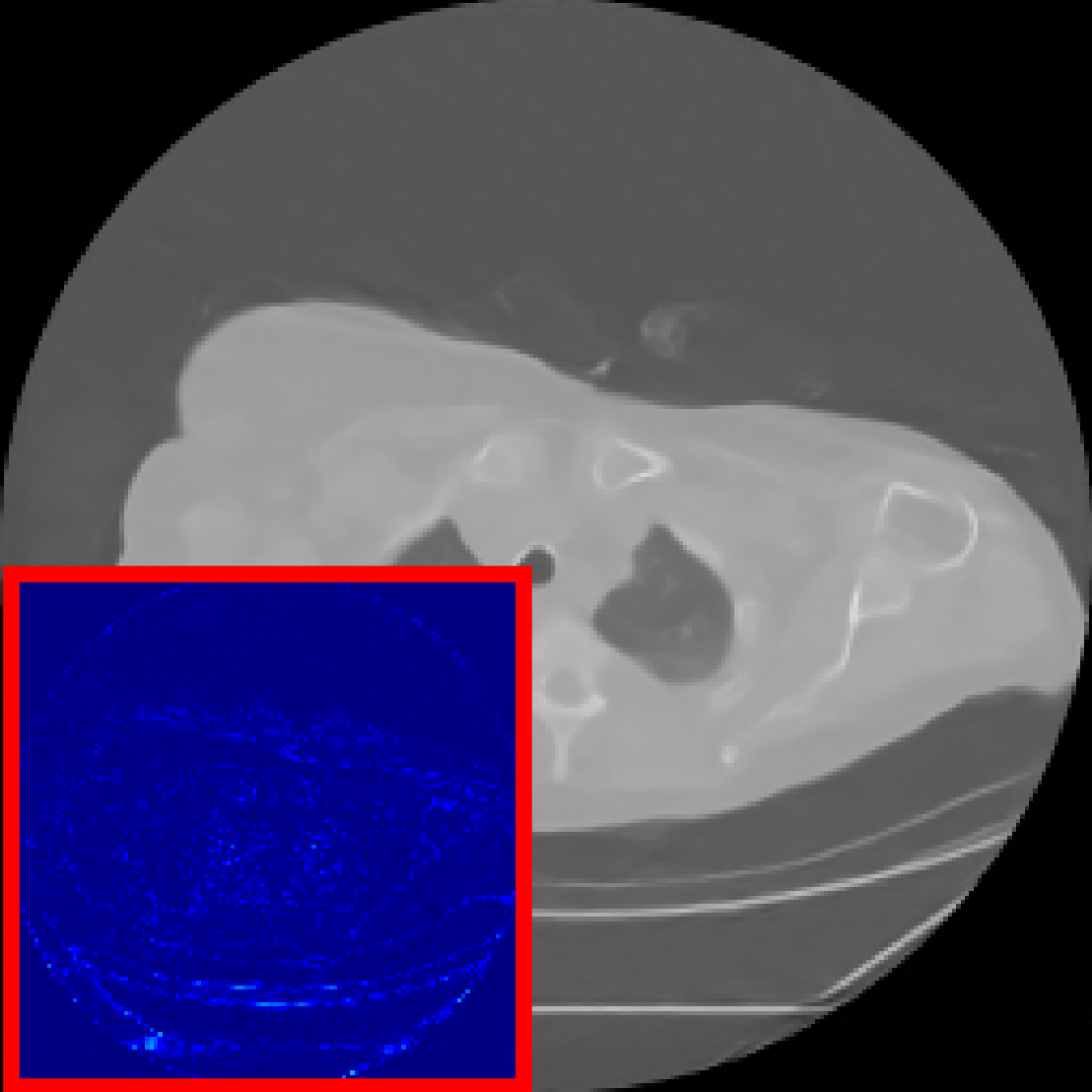} &    \includegraphics[width=0.19\linewidth]{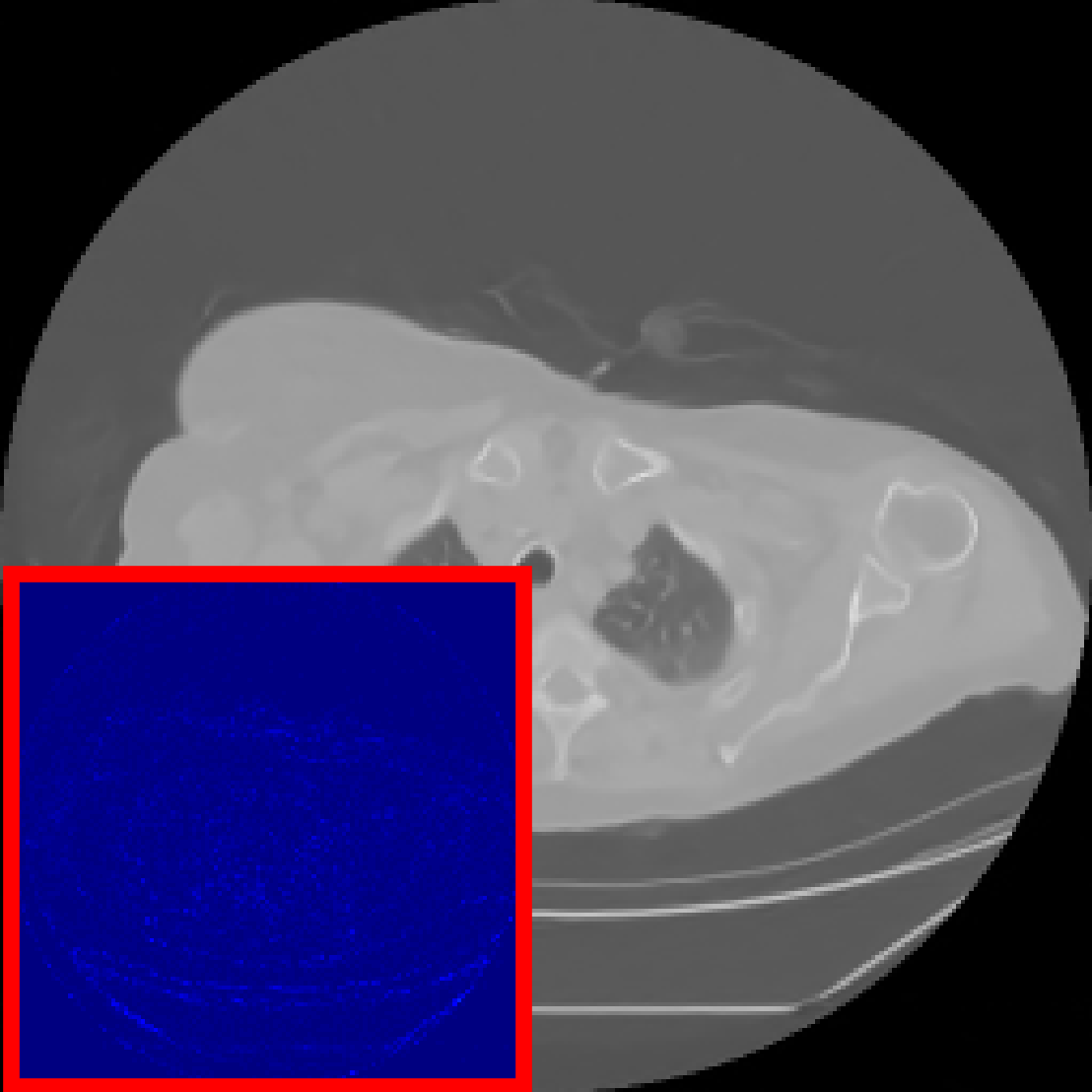} \\
        PSNR/SSIM & 29.44/0.6921 & \underline{\textcolor{blue}{41.57}}/\underline{\textcolor{blue}{0.9803}} & 38.41/0.9638 & \textbf{\textcolor{red}{44.81}}/\textbf{\textcolor{red}{0.9881}} \\ 

        \includegraphics[width=0.19\linewidth]{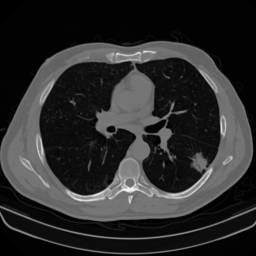} &         \includegraphics[width=0.19\linewidth]{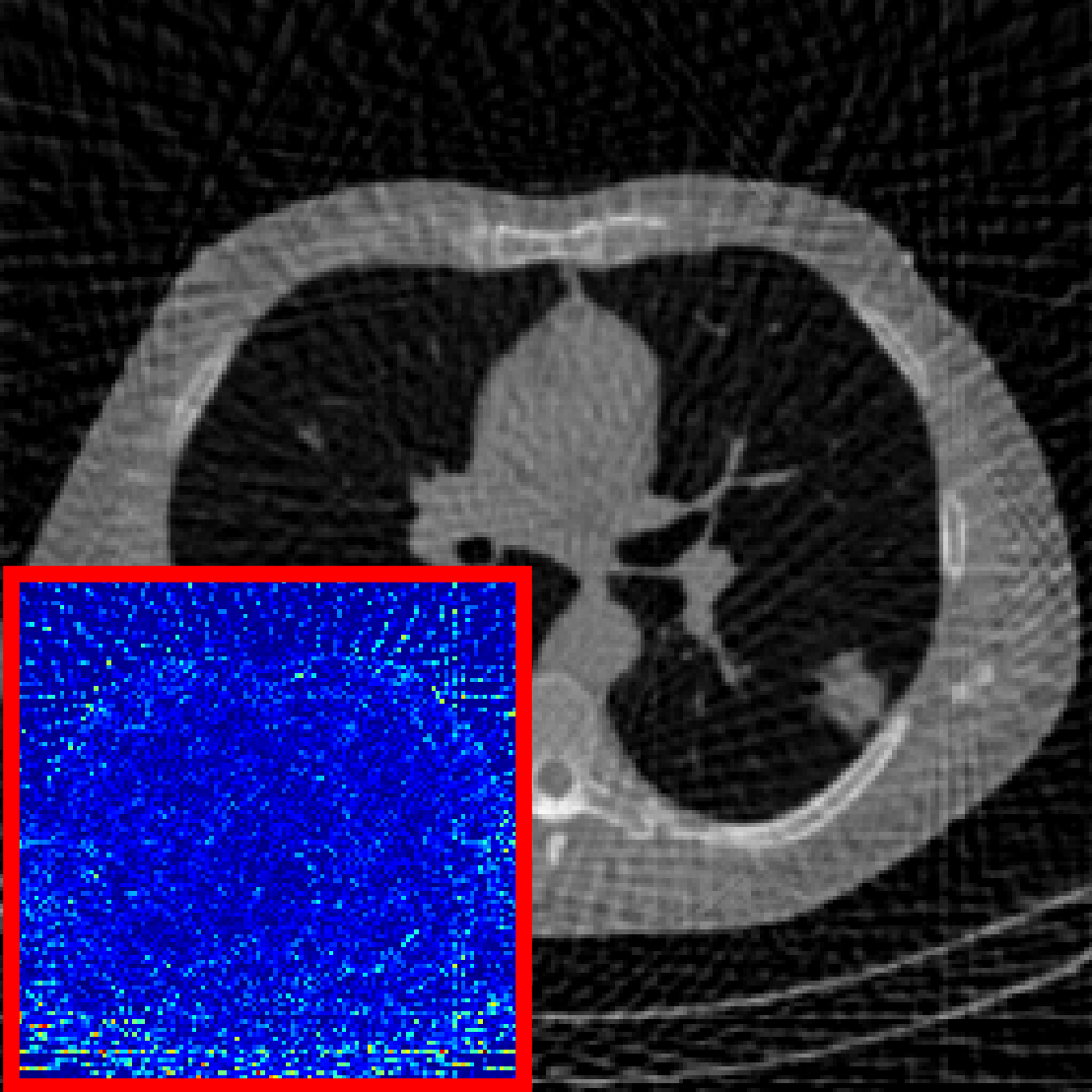}&     \includegraphics[width=0.19\linewidth]{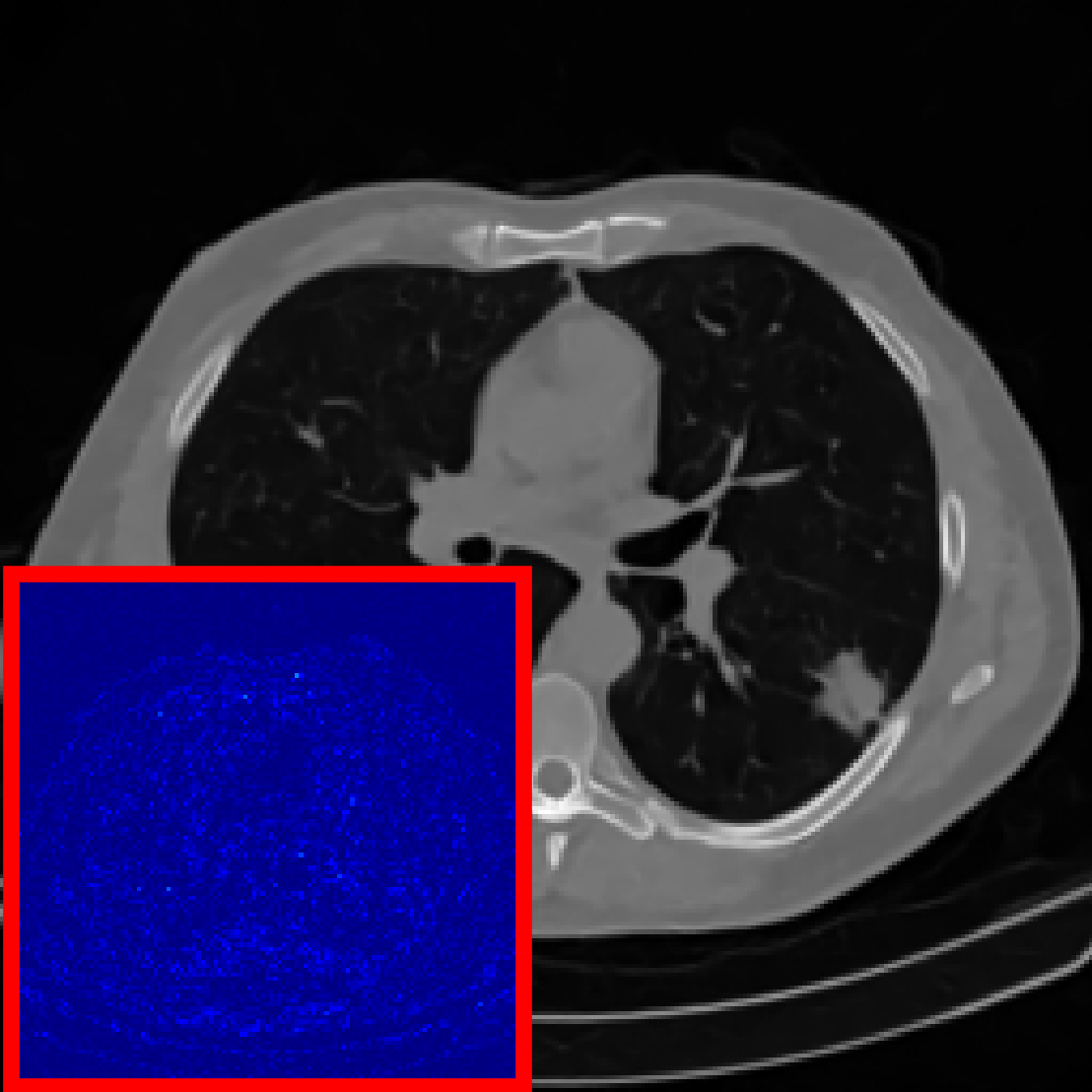}&     \includegraphics[width=0.19\linewidth]{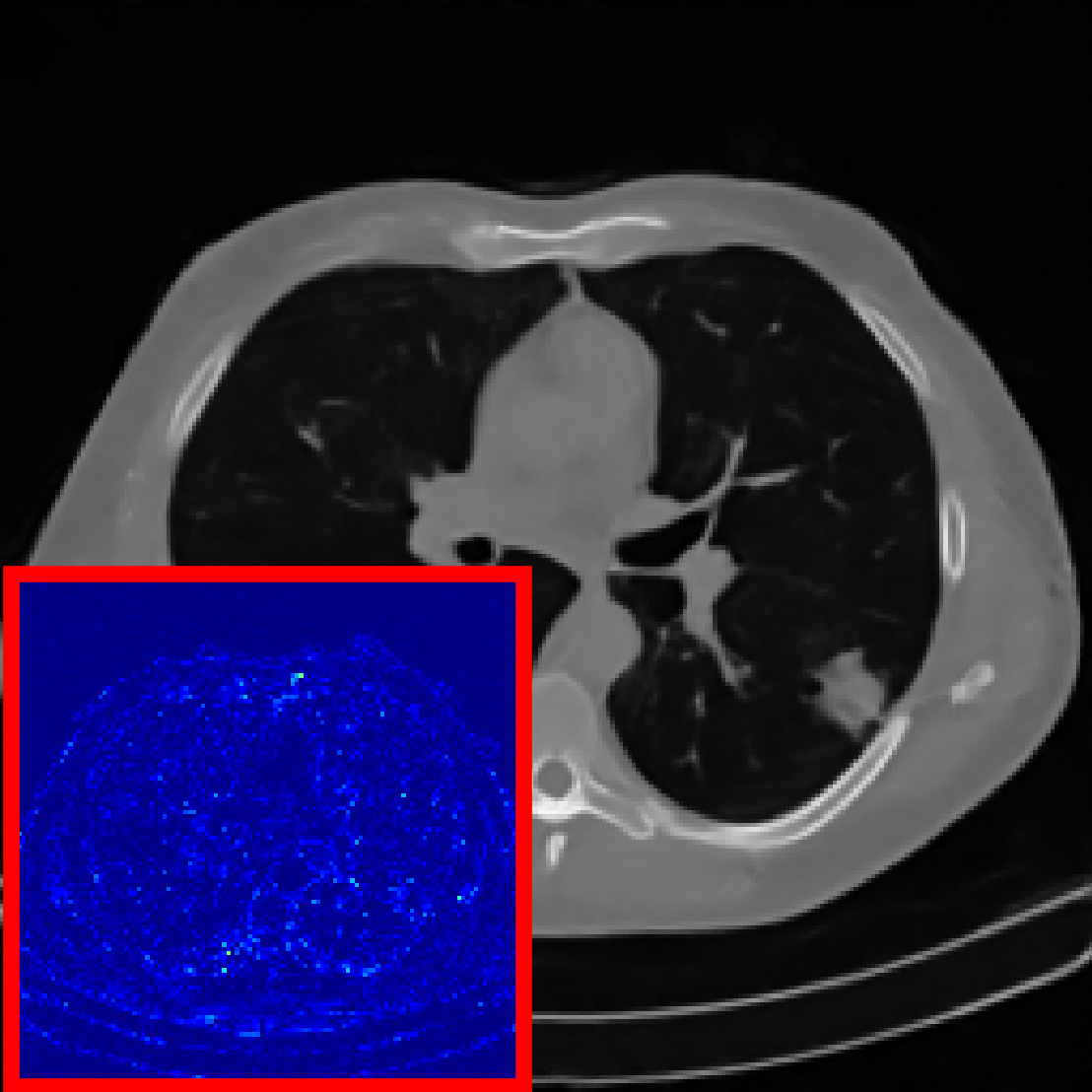} &    \includegraphics[width=0.19\linewidth]{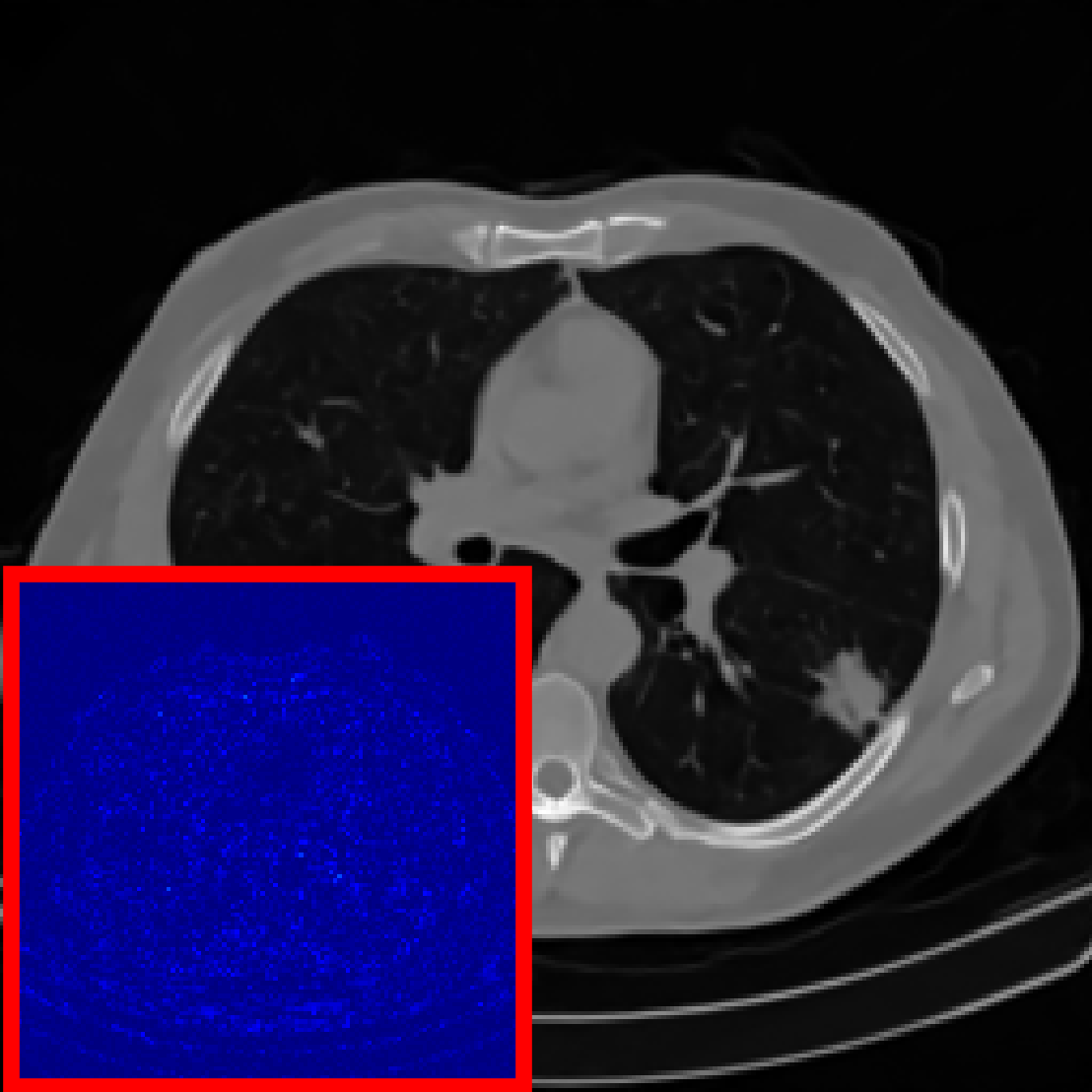} \\
        PSNR/SSIM & 26.29/0.5833 & \underline{\textcolor{blue}{40.13}}/\underline{\textcolor{blue}{0.9687}} & 35.69/0.9323 & \textbf{\textcolor{red}{40.77}}/\textbf{\textcolor{red}{0.9726}} \\ 
    \end{tabular}
    }
    \vspace{-5pt}
    \caption{Visual comparisons among four methods on inpainting (\textcolor{blue}{top}, 80\% masking ratio) and CT (\textcolor{blue}{bottom}, 50 views) tasks.}
    \label{fig:inpainting-CT}
    \vspace{-6pt}
\end{figure}

\begin{table}[!t]
    \centering
    \caption{Average PSNR/SSIM performance comparisons on image inpainting \textcolor{blue}{(top)} and sparse-view CT \textcolor{blue}{(bottom)} tasks.}
    \vspace{-5pt}
    \begin{minipage}[b]{1.0\linewidth}
    \resizebox{\linewidth}{!}{
    \begin{tabular}{c|cccc}
    \shline
         \rowcolor[HTML]{EFEFEF} Task & \multicolumn{4}{c}{Inpainting (Masking Ratio 80\%) on Urban100~\cite{huang2015single}}\\
         \hline \hline
         Method & $\mathbf{A}^{\dagger}\mathbf{y}$ & ID-Only & CCD-Only & \cellcolor[HTML]{FFFFA5}Dual-Domain\\
         \hline
         PSNR (dB) & 6.91 & 26.97~{\scriptsize(\textcolor{purple}{-0.70})} & 27.34~{\scriptsize(\textcolor{purple}{-0.33})} & \cellcolor[HTML]{FFFFA5}27.67\\
         SSIM & 0.0718 & 0.8708 & 0.8786 & \cellcolor[HTML]{FFFFA5}0.8850\\
         \shline
    \end{tabular}
    }
    \end{minipage}
    
    \vspace{5pt}
    
    \begin{minipage}[b]{1.0\linewidth}
    \resizebox{\linewidth}{!}{
    \begin{tabular}{c|cccc}
    \shline
         \rowcolor[HTML]{EFEFEF} Task & \multicolumn{4}{c}{Sparse-View CT (50 Views) on CT100~\cite{clark2013cancer}} \\
         \hline \hline
         Method & $\mathbf{A}^{\dagger}\mathbf{y}$ & ID-Only & CCD-Only & \cellcolor[HTML]{FFFFA5}Dual-Domain  \\
         \hline
         PSNR (dB) & 29.16 & 40.42~{\scriptsize(\textcolor{purple}{-0.77})} & 36.80~{\scriptsize(\textcolor{purple}{-4.39})} & \cellcolor[HTML]{FFFFA5}41.19 \\
         SSIM & 0.6513 & 0.9644 & 0.9349 & \cellcolor[HTML]{FFFFA5}0.9684 \\
         \shline
    \end{tabular}
    }
    \end{minipage}

    \label{tab:inpainting_CT}
    \vspace{-3pt}
\end{table}

\textbf{Application to Image Inpainting:} We further extend the capabilities of our $\DC$-Net to the challenging task of image inpainting, which involves reconstructing missing regions or pixels within a target image. In this task, we employ a binary diagonal matrix of size $N \times N$ as the sampling matrix, resulting in the random removal of a portion of pixels in the original image (80\% in our experiments) and replacing missing areas with zeros. We use the same training set as the one in image CS experiments and evaluate different methods on Urban100 \cite{huang2015single}. Quantitative/qualitative results are presented in Tab.~\ref{tab:inpainting_CT} (top) and Fig.~\ref{fig:inpainting-CT} (top), respectively. Compared to ID-only and CCD-only variants, $\DC$-Net exhibits superior performance in reconstructing fine details with learned dual-domain priors.

\textbf{Application to Sparse-View CT:} We extend $\DC$-Net to the task of X-ray computed tomography (CT), where the imaging physics model is based on the discrete \texttt{radon} transform. In this task, the sampling matrix $\bPhi$ is implemented by \texttt{radon} transformation with 50 views (angles) uniformly subsampled to generate sparse-view measurement $\y$. We utilize the filtered back projection (FBP) function \texttt{iradon} to approximate the pseudo-inverse of $\bPhi$. We follow \cite{chen2021equivariant} to utilize the first 90 images from the CT100 dataset \cite{clark2013cancer} for training while reserving the remaining 10 images for testing. Quantitative/qualitative results are exhibited in Tab.~\ref{tab:inpainting_CT} (bottom) and Fig.~\ref{fig:inpainting-CT} (bottom), respectively. We observe that both ID- and CCD-only variants exhibit visible line artifacts, whereas $\DC$-Net significantly mitigates these issues by yielding high-quality results.

\textbf{Application to Fluorescence Microscopy with Real SPI Optics System:} To demonstrate the effectiveness of our $\DC$-Net in real-world CS imaging applications, we develop an SPI system tailored for fluorescence microscopy. Details on the experimental setup and results, including the SPI optics system layout, are provided by Sec.~\textcolor{red}{B} in the \textit{\textbf{\textcolor{blue}{supplemental material}}}.

\vspace{-5pt}
\section{Conclusion}\label{sec:conclusion}
Motivated by the algorithm-unfolding paradigm and convolutional coding methods, we introduce a dual-domain optimization framework that integrates both image-domain (ID) and convolutional-coding-domain (CCD) priors to effectively constrain the final solution space. Our framework is able to stand out by leveraging implicitly learned deep priors instead of traditional hand-crafted regularizers \cite{fu2019jpeg, xu2020limited, sreter2018learned, gao2022multi}, enabling the utilization of the learning capacity of NNs. Moreover, unlike existing deep convolutional coding methods \cite{wang2020model, zheng2021deep} that are limited to tasks where the degradation matrix $\bPhi$ is the identity $\mathbf{I}_N$, our framework demonstrates flexibility and generalizability across image image CS and other inverse imaging problems. Building upon this effective framework, we propose a novel \textbf{D}ual-\textbf{D}omain \textbf{D}eep \textbf{C}onvolutional \textbf{C}oding \textbf{Net}work, referred to as $\DC$-Net, for CS imaging. In contrast to existing CS DUNs \cite{zhang2018ista, zhang2020optimization, you2021coast, you2021ista}, our $\DC$-Net transmits high-capacity feature-level representations across all unfolded stages and adaptively captures informative features. Extensive experiments across five tasks, covering natural, medical, and scientific signals, demonstrate the superiority of $\DC$-Net over other CS methods in terms of accuracy and efficiency. Future work is to extend our optimization framework and $\DC$-Net to other inverse imaging tasks and video/multi-modal applications~\cite{deng2020deep}.

\rev{\textbf{Discussions about $\DC$-Net and diffusion models.} These two kinds of models have different fields of application. $\DC$-Net is a kind of regression-based DUN, which brings less distortion through a single fast forward propagation and, thus is suitable for fidelity-oriented tasks due to its higher reconstruction accuracy. In contrast, diffusion model-based methods ~\cite{han2024image,ho2022classifier,epstein2023diffusion} utilize generative diffusion priors, requiring thousands of iterations to produce results of high perceptual quality, which is generally preferred by humans.}

\vspace{-5pt}
\bibliography{ref}
\bibliographystyle{IEEEtran}


 

\vspace{-30pt}
\begin{IEEEbiography}[\vspace{-0.8cm}{\includegraphics[width=1in,height=1.25in,clip,keepaspectratio]{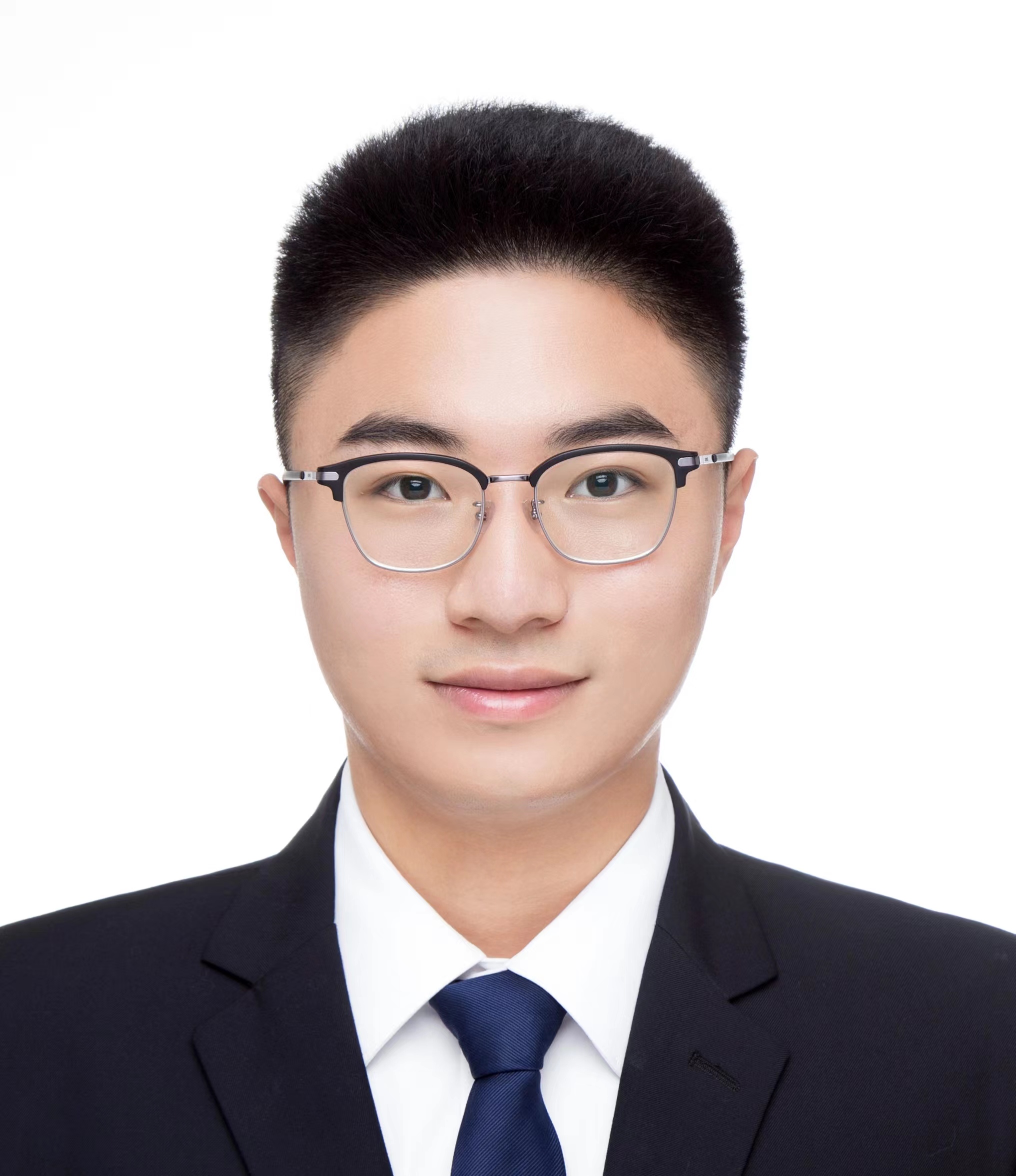}}]{Weiqi Li} received the B.E. degree from the school of Software Engineering, Tongji University, Shanghai, China, in 2022. He is currently working toward the master's degree in computer applications technology at the School of Electronic and Computer Engineering, Peking University, Shenzhen, China. His research interests include image super-resolution, compressive sensing, and computer vision. 
\end{IEEEbiography}

\vspace{-50pt}
\begin{IEEEbiography}
[\vspace{-0.8cm}{\includegraphics[width=1in,height=1.25in,clip,keepaspectratio]{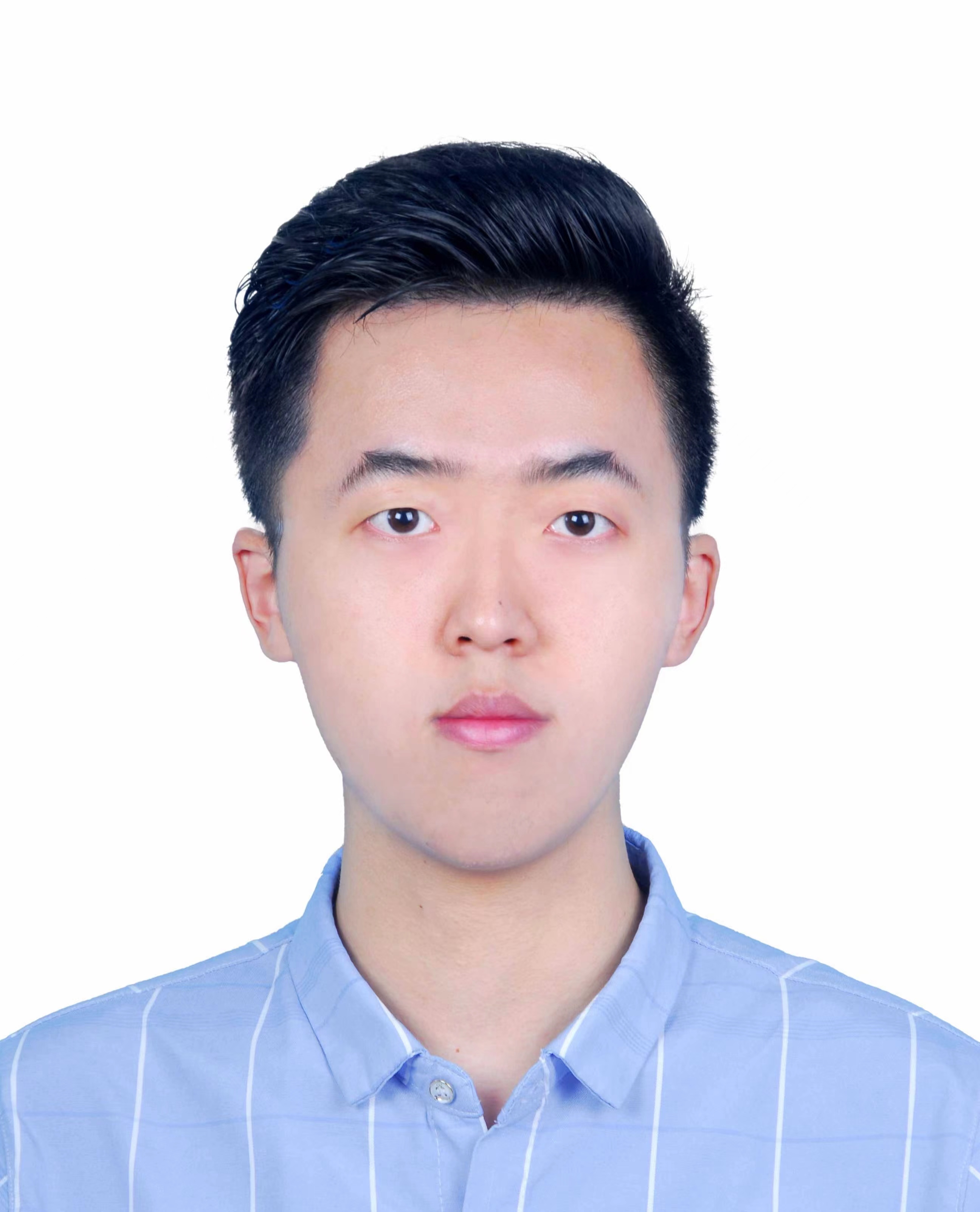}}]{Bin Chen} received the B.S. degree in the School of Computer Science, Beijing University of Posts and Telecommunications, Beijing, China, in 2021. He is currently working toward a Ph.D. degree in computer applications technology at the School of Electronic and Computer Engineering, Peking University, Shenzhen, China. His research interests include compressive sensing, image restoration, and computer vision.
\end{IEEEbiography}

\vspace{-40pt}
\begin{IEEEbiography}[\vspace{-0.8cm}{\includegraphics[width=1in,height=1.25in,clip,keepaspectratio]{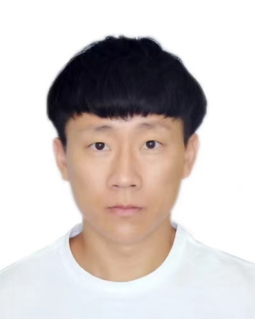}}]{Shuai Liu} received the B.S. degree from the School of Physical Science and Technology, Inner Mongolia University, China, in 2017, and the M.S. degree from the School of Physics, Beijing Institute of Technology, China, in 2020. He is currently working toward a Ph.D. degree in Control Science and Engineering at Tsinghua University, China. His research interests include compressive sensing, image restoration, and Computational Imaging.
\end{IEEEbiography}

\vspace{-30pt}
\begin{IEEEbiography}[{\includegraphics[width=1in,height=1.25in,clip,keepaspectratio]{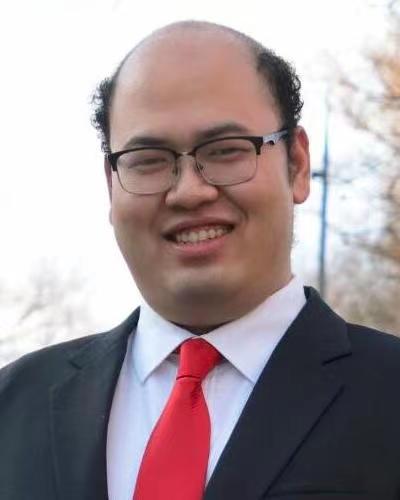}}]{Bowen Du} received his Ph.D. degree in computer science from the University of Warwick in 2022. He is currently an assistant professor at the School of Software Engineering, Tongji University, Shanghai, China. His research interests include cyber-physical systems, artificial intelligence, and software engineering.
\end{IEEEbiography}

\vspace{-30pt}
\begin{IEEEbiography}[\vspace{-0.4cm}{\includegraphics[width=1in,height=1.25in,clip,keepaspectratio]{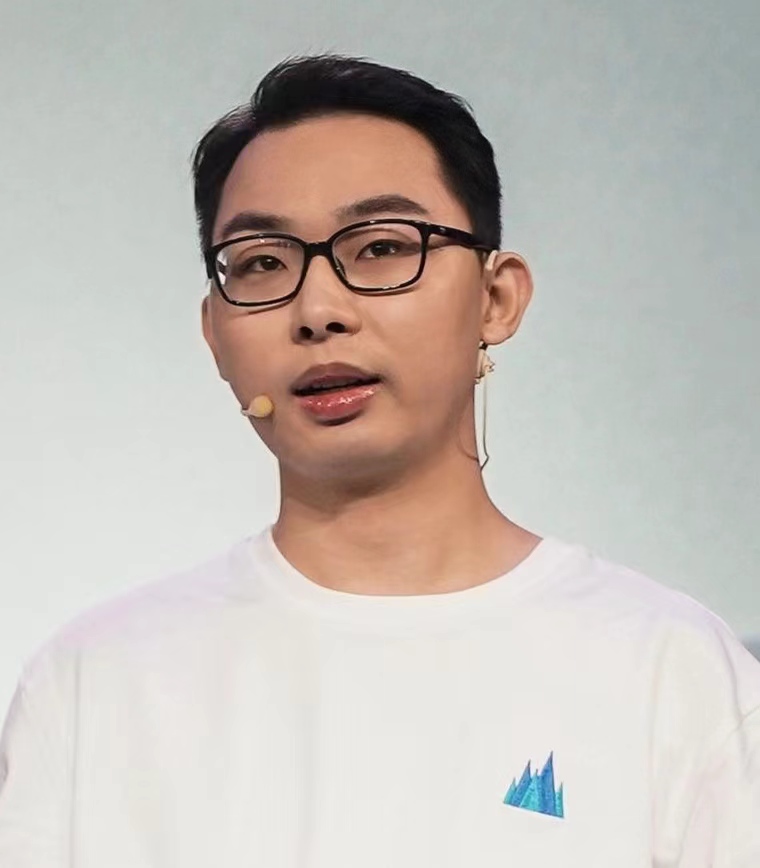}}]{Shijie Zhao} received his Bachelor's and Doctoral degrees from the School of Mathematics at Zhejiang University, in addition to a Master's degree from Imperial College London. Currently, he leads the Video Processing and Enhancement team at ByteDance's Multimedia Lab. His main areas of research include video enhancement, low-level vision, and video compression.
\end{IEEEbiography}

\vspace{-30pt}
\begin{IEEEbiography}[{\includegraphics[width=1in,height=1.25in,clip,keepaspectratio]{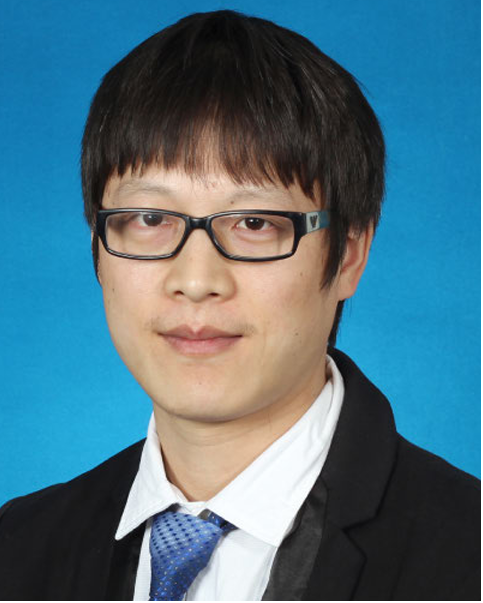}}]{Yongbing Zhang} received his Ph.D. degree in computer science from the Harbin Institute of Technology, Harbin, China, in 2010. He is currently a professor of computer science and technology at the Harbin Institute of Technology (Shenzhen), Shenzhen, 
China. His research interests include computational imaging, especially exploring new image acquisition and intelligent processing methods and equipment through the deep intersection of machine learning, signal processing, and Fourier optics.
\end{IEEEbiography}

\vspace{-30pt}
\begin{IEEEbiography}[{\includegraphics[width=1in,height=1.25in,keepaspectratio,clip]{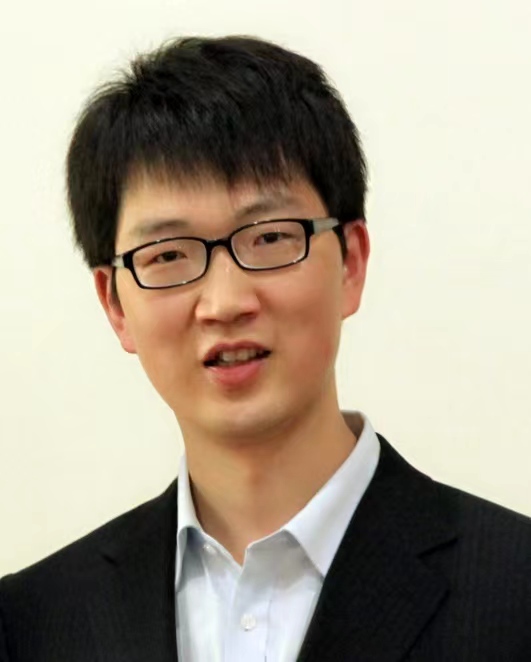}}]{Jian Zhang} received his Ph.D. degree in computer science from the Harbin Institute of Technology in 2014. Currently, he is an Assistant Professor and leads the Visual-Information Intelligent Learning LAB (VILLA) at the School of Electronic and Computer Engineering, Peking University, Shenzhen, China. His research interest focuses on intelligent multimedia processing, including low-level vision, AI-generated content (AIGC) and security. He has published over 100 technical articles in refereed international journals and proceedings. He received the Best Paper Award at the 2011 IEEE Visual Communications and Image Processing (VCIP) and was a co-recipient of the Best Paper Award of 2018 IEEE MultiMedia.

\end{IEEEbiography}



\vfill

\end{document}